\documentclass[Afour,sageh,times]{sagej}

\usepackage{moreverb,url}
\usepackage{anyfontsize}
\usepackage[colorlinks,bookmarksopen,citecolor=red,urlcolor=red]{hyperref}

\usepackage{pifont}
\usepackage{caption}
\usepackage{amssymb}
\usepackage{parskip}
\usepackage{amsmath}
\usepackage{bbm}
\usepackage{bm}
\usepackage{float}
\usepackage{graphicx}
\usepackage{booktabs}
\usepackage{multirow}
\usepackage{makecell}
\usepackage{titlesec}
\usepackage{pifont}
\usepackage[para,online,flushleft]{threeparttable}
\newcommand{\cmark}{\ding{51}}%
\newcommand{\xmark}{\ding{55}}%
\newcommand{\sstar}{\ding{73}}%
\usepackage{soul}
\usepackage[dvipsnames,table]{xcolor}
\usepackage{tikz}
\usetikzlibrary{decorations.text}
\usetikzlibrary{calc}
\usetikzlibrary{fit}
\usetikzlibrary{bayesnet}
\usetikzlibrary{arrows}
\usetikzlibrary {arrows.meta}
\usetikzlibrary{positioning}
\usetikzlibrary{shapes}
\usetikzlibrary{shadows}
\usetikzlibrary{shadings}
\usetikzlibrary{mindmap}
\usepackage{graphicx}
\usepackage{pgfplots}
\pgfplotsset{compat=1.18}
\usepackage{subcaption}
\usepackage{adjustbox}
\usepackage{enumitem,setspace,kantlipsum}
\definecolor{darkred}{RGB}{192,0,0}
\definecolor{lightgray}{HTML}{D2D1CE}
\definecolor{light-green}{RGB}{129, 184, 113}
\usepackage[utf8]{inputenc}
\usepackage{natbib}

\newcommand\BibTeX{{\rmfamily B\kern-.05em \textsc{i\kern-.025em b}\kern-.08em
T\kern-.1667em\lower.7ex\hbox{E}\kern-.125emX}}

\def\volumeyear{2024}

\newcommand{\citetitle}[2]{%
  \href{\#cite.#2}{\textcolor{red}{#1}~(\citeyear{#2})}%
}

\begin{document}

\runninghead{Bridging Language and Action: A Survey of Language-Conditioned Robot Manipulation}

\title{Bridging Language and Action:\\A Survey of Language-Conditioned Robot Manipulation}

\author{Xiangtong Yao\affilnum{1}*, Hongkuan Zhou\affilnum{2,12}*, Oier Mees\affilnum{3,4}*, Yuan Meng\affilnum{1}*, Ted Xiao\affilnum{5}, Yonatan Bisk\affilnum{6}, Jean Oh\affilnum{6}, Edward Johns\affilnum{7}, 
Mohit Shridhar\affilnum{5},
Dhruv Shah\affilnum{5,8},
Jesse Thomason\affilnum{9},
Kai Huang\affilnum{10}, Joyce Chai\affilnum{11}, Zhenshan Bing\affilnum{1,13}, Alois Knoll\affilnum{1}}

\affiliation{\affilnum{1} Technical University of Munich, Munich, Germany\\
\affilnum{2} Corporate Research, Robert Bosch GmbH, Renningen, Germany\\
\affilnum{3} University of California Berkeley, USA\\
\affilnum{4} Microsoft, Switzerland\\
\affilnum{5} Google DeepMind, USA \& UK\\
\affilnum{6} Carnegie Mellon University, USA\\
\affilnum{7} Imperial College London, UK\\
\affilnum{8} Princeton University, USA\\
\affilnum{9} University of Southern California, USA\\
\affilnum{10} Sun Yat-sen University, Guangzhou, China\\
\affilnum{11} University of Michigan, USA\\
\affilnum{12} Institute for Artificial Intelligence, University of Stuttgart, Stuttgart, Germany \\
\affilnum{13} The State Key Laboratory for Novel Software Technology, Nanjing University, Suzhou, China \\
* Equal contribution \\
\\
This version is accepted by \textit{The International Journal of Robotics Research} (IJRR).
}

\corrauth{Zhenshan Bing}
\email{zhenshan.bing@tum.de, bing@nju.edu.cn}

\begin{abstract}
Language-conditioned robot manipulation is an emerging field aimed at enabling seamless communication and cooperation between humans and robotic agents by teaching robots to comprehend and execute instructions conveyed in natural language. This interdisciplinary area integrates scene understanding, language processing, and policy learning to bridge the gap between human instructions and robot actions. In this comprehensive survey, we systematically explore recent advancements in language-conditioned robot manipulation. We categorize existing methods based on the primary ways language is integrated into the robot system, namely language for state evaluation, language as a policy condition, language for cognitive planning and reasoning, and \textcolor{black}{language in unified vision-language-action models}. Specifically, we further analyze state-of-the-art techniques from five axes of action granularity, data and supervision regimes, system cost and latency, environments and evaluations, and \textcolor{black}{task specification}. Additionally, we highlight the key debates in the field. Finally, we discuss open challenges and future research directions, focusing on potentially enhancing generalization capabilities and addressing safety issues in language-conditioned robot manipulators.
\end{abstract}


\maketitle
\section{Introduction}

\textcolor{black}{Robot manipulation, the ability of robots to physically interact with and manipulate objects in their environment~\citep{billard2019trends}, is an important component of autonomous systems. It has been widely adopted in structured environments, such as factory floors~\citep{sanchez2018robotic}, where robots perform repetitive tasks with high precision and efficiency. One vision of robotics is to integrate intelligent systems into the fabric of everyday life, moving them from the environment with controlled settings to unstructured scenarios in homes, hospitals, and warehouses, where high levels of interaction between humans and robots are required~\citep{silvera2024robotics, zhang2025new}. A fundamental barrier, however, has long hindered this vision: the complexity of instructing robots in unstructured environments, particularly for non-expert users. Traditional methods for instructing robots are non-generalizable, such as specialized programming of control rules~\citep{takeshita2025robot}, teleoperation~\citep{wu2019teleoperation}, or reward functions engineering~\citep{ibarz2021train}. These methods demand expert efforts, which are inaccessible to the general public without extensive training. This limitation has impeded the broader deployment of general-purpose robots.}

Language-conditioned robotic manipulation is emerging as a transformative solution to this challenge. It leverages natural language as an intuitive interface for humans to specify tasks, enabling robots to translate these high-level commands into physical actions. Enabling control via simple text or voice lowers the barrier for non-expert users and unlocks the potential for robots to function as general-purpose assistants, making robotics more accessible to a broader audience and increasing its importance. \textcolor{black}{This motivation is shared by a broader family of intuitive task-specification paradigms. Image- and video-conditioned manipulation methods, for instance, aim to reduce the burden of robot instruction by enabling users to specify tasks through target visual states, demonstrations, or perceptual references instead of low-level programming. However, different conditioning modalities provide different forms of task information.}

\textcolor{black}{Image-conditioned methods are well-suited for specifying desired object configurations or goal states~\citep{10161178,wang2023manipulate}, while video-conditioned methods convey motion patterns, temporal evolution, and demonstration trajectories~\citep{10064325,11215739}. Compared with these visual modalities, language provides a more compact and editable interface for expressing abstract task intent, relational constraints, safety requirements, preferences, and corrections. Beyond external task specification, language can also serve as an internal cognitive medium for robot reasoning, enabling systems to decompose long-horizon tasks into subgoals, reason about affordances and causality, and generate plans or executable code. Therefore, while language-, image-, and video-conditioned manipulation are complementary paradigms for intuitive robot instruction. The broader functional role of language motivates a dedicated survey on how it is grounded in perception, planning, and control. The advantages of language conditioning can be summarized as follows.}
\textcolor{black}{\begin{itemize}[nosep,noitemsep]
    \item \textbf{Accessibility and usability}: Language conditioning allows people who are not experts to use robots. As noted by \citet{tellex2020robots}: ``\textit{most future robot users will not be programmers}''. Natural language provides a ``zero-learning'' interface that allows anyone to specify complex and high-level goals without needing to understand the underlying code or control logic. Instead of navigating complex graphical user interfaces (GUIs) or writing scripts, a user can simply state their intent, such as ``\textit{bring me the red cup from the kitchen counter}'', lowering the barrier to entry for a wide range of applications.
    \item \textbf{Trust through bidirectional communication:} Trust is essential for human-robot collaboration, especially in sensitive environments, such as assistive care~\citep{jenamani2025feast}. Language-conditioned systems offer a natural pathway to building that trust. The same channel used to command a robot can be used for on-the-fly correction and explanation. A user asks, ``\textit{I have a trach and have to eat slowly}''~\citep{jenamani2025feast}, and the robot can provide a rationale for its actions. These transparent feedback loops enable more intuitive and safer interactions, a key theme in human-robot interaction research focused on closing the loop between robot learning and communication~\citep{habibian2025survey}.
    \item \textbf{Transferring textual knowledge to robotics:} Real-world environments like homes and factories are unpredictable, with many unique situations that pre-programmed robots cannot handle. Language provides a compact interface for importing the world's common sense knowledge, such as properties, procedures, and safety rules, into the control system. By grounding natural-language commands to perception and action primitives, a system can parse goals like ``\textit{stack the two lightest boxes}'' into perceptual queries, relational reasoning, and sequences of controllable skills. Leveraging broad linguistic priors reduces per-task engineering and enables zero/few-shot generalization to unseen tasks~\citep{brohan2023rt,zhang:rewind}, improving robustness when operating in dynamic environments.
\end{itemize}}

\input{tikz/summery}

Driven by these advantages, recent research goals in the emerging field of language-conditioned robot manipulation aim to enable natural language commands, instructions, and queries to robot systems that are translated into motor actions and behaviors conditioned on visual observations of the physical environment.
Remarkable success has been seen in robotic arm control~\citep{620146, zhang1999interactive, lynch2020language, silva2021lancon, mees2022matters,  open_x_embodiment_rt_x_2023,bing2023meta,octomodelteam2024octo} ,cross-embodiment learning~\citep{yang2024pushing, Doshi24-crossformer},
and robot navigation tasks~\citep{hermann2017grounded,fu2019language, hirose24lelan} such as autonomous driving~\citep{sriram2019talk, roh2020conditional}. Figure \ref{fig:summery} demonstrates fundamental disciplines and research in language-conditioned robot manipulation. \textcolor{black}{Success in this field requires tackling challenges across three key axes: language understanding, visual perception, and action generation. From a system perspective, it is useful to factor a language-conditioned manipulation pipeline into three interacting modules: Language, Perception, and Control. This decomposition is architectural rather than conventional robotics taxonomy: here, the language module includes instruction understanding, task representation, and often high-level semantic planning/reasoning; the perception module grounds language in observations and estimates the environment state; and the control module converts the resulting task specification into executable robot actions through learned policies or classical planners/controllers, as shown in Figure \ref{fig:architecture}.}

\input{tikz/architeture}

To extract semantics from natural language, early research works~\citep{kress2009temporal, raman2012temporal} leverage formal specification languages such as temporal logic, which support formal verification of provided commands. Nevertheless, specifying instructions in these languages can be challenging, complex, and time-consuming. Large-scale pretraining approaches in natural language processing have led to text embedding models, like GloVe~\citep{pennington2014glove}, RoBERTa~\citep{liu2019roberta},~and BERT~\citep{devlin2018bert}. In particular, large language models (LLMs) have shown impressive abilities in extracting semantic information and performing high-level planning and reasoning tasks.

To ground language commands in the physical scene, robots must perceive their environments, a challenge addressed by advances in computer vision. Foundational technologies include convolutional neural networks (e.g., ResNet~\citep{he2016deep}) for feature extraction, object detectors (e.g., Faster-RCNN~\citep{girshick2015fast}, YOLO~\citep{redmon2016you}), and more recent transformer-based architectures like Vision Transformers (ViT)~\citep{dosovitskiy2020image}. To connect visual perception with linguistic instructions, vision-language models (VLMs) such as CLIP~\citep{radford2021learning} and Flamingo~\citep{alayrac2022flamingo} have been developed. These models learn joint embeddings that align visual and textual data, enabling robots to perform tasks like language-conditioned object detection~\citep{DBLP:journals/corr/abs-2305-18279} and segmentation~\citep{DBLP:conf/iclr/LiWBKR22}, which are critical for identifying and localizing objects mentioned in a command.

\textcolor{black}{To translate high-level linguistic goals into low-level robot actions, researchers explore several paradigms. One major approach is to learn a {language-conditioned policy} using methods like reinforcement learning (RL)~\citep{sutton1998introduction} or imitation learning (IL)~\citep{Imitation_Learning}. In this paradigm, the language command is provided as an input to the policy, which then directly outputs low-level actions (e.g., joint torques or end-effector velocities). For instance, in language-conditioned IL, the model learns to map from images and text commands to expert actions from a dataset of language-annotated demonstrations. A second popular approach decouples high-level reasoning from low-level control. In these methods, a large language model or vision-language model is used to parse the instruction and predict an intermediate goal, such as a target end-effector pose or a grasp point. This high-level prediction is then passed to a classical motion planner that uses inverse kinematics to compute the required joint movements~\citep{habekost2024inverse,yang2025guiding}. This modular design leverages the \textcolor{black}{strong open-vocabulary and instruction-following priors} of foundation models (FMs) for goal specification while relying on robust non-learning-based controllers for precise manipulation~\citep{yang2025guiding}.} 

\textcolor{black}{While this ``Language-Perception-Control'' framework provides a useful high-level overview, a deeper analysis reveals that the central research questions are not just about these components, but about the \textit{specific functional role language plays in bridging them}. Different approaches leverage language in fundamentally distinct ways to solve the manipulation problem. This survey will focus on this perspective, categorizing recent methods based on how language enters the manipulation control loop. }

\subsection{Contributions}

\textbf{A New Taxonomy and Comprehensive Review:} We introduce a new taxonomy that organizes the field by the functional role language plays in the manipulation control loop. This taxonomy categorizes methods into \textcolor{black}{four} primary themes: language for state evaluation (\textcolor{black}{using language to define goals and assess task-relevant states for high-level planning, such as task decomposition and subgoal sequencing, or for learning, such as reward design, value estimation, and policy optimization}), language as a direct policy condition (using language to specify desired behavior), language for cognitive planning and reasoning, and language-driven end-to-end policy (vision-language-action models). Based on this framework, we provide a comprehensive review of state-of-the-art methods, from traditional language-conditioned policies to the latest approaches driven by FMs, including LLMs, VLMs, VLAs, and neuro-symbolic integrations. This structured review highlights how different methods leverage language to bridge \textcolor{black}{perception (grounding language in sensory observations and extracting task-relevant state information), planning/reasoning (inferring task structure, constraints, and subgoals), and control (generating executable actions or policy outputs). Moreover, to provide a multifaceted understanding, we conduct a systematic comparative analysis from an orthogonal perspective. We evaluate the surveyed methods across several key dimensions: action granularity, data and supervision requirements, system cost and latency, the environments and benchmarks used for validation, \textcolor{black}{and cross-modal task specification}. This analysis offers practical insights into the trade-offs and applicability of different approaches.} 

\textbf{Discussion and Future Directions:}~\textcolor{black}{Additionally, we delve into the key debates, such as whether scaling up VLA models is the most effective path forward compared to incorporating structured world models or other hybrid approaches. We also discuss open challenges currently shaping the field. Building on this discussion, we outline the primary limitations and future directions, focusing on two critical areas: enhancing generalization capability and ensuring real-world safety. To improve generalization, we propose focusing on the development of large-scale, diverse datasets, integrating lifelong learning frameworks to enable continuous adaptation, and establishing methods for cross-embodiment alignment. To address safety, we highlight the need for mechanisms to handle language ambiguity, improve failure recovery, and guarantee the real-time performance required for unstructured environments.}

\subsection{Relation to existing surveys}

Before the emergence of LLMs, \citet{tellex2020robots} provided a foundational review of language grounding in robotics, categorizing approaches by their technical underpinnings (e.g., ``lexically grounded'' vs. ``learning methods''). Recent surveys, prompted by the rise of FMs, have offered broad perspectives. Surveys by \citet{DBLP:journals/corr/abs-2312-08782,li2024foundation,DBLP:journals/corr/abs-2311-14379} discuss the application of FMs in robotics, organizing their findings by model type and the specific robotic module they enhance, such as perception and planning. Similarly, \citet{firoozi2025foundation} structures its analysis around general robotics capabilities like ``Perception'', ``Decision-making'', and ``Control''.

\textcolor{black}{While these surveys provide essential overviews, our work offers a distinct and orthogonal perspective. Rather than categorizing by model type or the robotic module it replaces, our survey organizes the field by the functional role language plays within the manipulation control loop. This taxonomy allows for a finer-grained analysis that spans multiple models and algorithms. We categorize approaches into \textcolor{black}{four} types: language for state evaluation, language as a policy condition, language for cognitive planning and reasoning, \textcolor{black}{and language in unified vision-language-action models (VLAs)}. This taxonomy allows us to systematically cover a wide range of techniques, including language-conditioned policy learning/planning, neuro-symbolic methods, and emerging paradigms based on LLMs, VLMs, and VLAs. By focusing on the role of language, our survey provides a new perspective for understanding the diverse ways we can bridge language and action in robotic manipulation.}

\subsection{Organization}
The rest of this paper is organized as follows. Section~\ref{sec:background} presents foundational concepts relevant to language-conditioned robot manipulation. In Section~\ref{sec:Taxonomy}, we elaborate on the taxonomy of the recent approaches that they can be categorized into \emph{language for state evaluation} (Section~\ref{sec:language-for-state-evaluation}), \emph{language as policy condition} (Section~\ref{sec:language-as-policy-condition}), \emph{language for cognitive planning and reasoning} (Section~\ref{sec:language-cognitive-planning-reasoning}), and \emph{Language in unified
vision-language-action models} (Section~\ref{sec:vla}). Additionally, in Section~\ref{sec:comparative_analysis}, \textcolor{black}{we conduct a comprehensive comparative analysis of various approaches from a different perspective, focusing on the dimensions of \emph{action granularity}, \emph{data and supervision regime}, \emph{system cost and latency}, \emph{environment and evaluation}, as well as \emph{manipulation task spectrum}. Finally, we present the key debates in this field in Section~\ref{sec:discussion}, outline the challenges and future directions in Section~\ref{sec:future_directions}, and provide conclusions in Section~\ref{sec:conclusion}. }

\section{Background}
\label{sec:background}

We present fundamental terms and concepts in this section. An understanding of these principles is crucial, as they serve as the cornerstone for the more advanced methods discussed throughout this article. Table \ref{tab:abbreviations} provides an overview of important abbreviations used in this article.

\begin{table}[htbp]
    \centering
    \caption{Abbreviations}

    \begin{tabular}{ l l }
        \toprule
        Abbreviations & Meaning \\
        \midrule
        MDP & Markov Decision Process \\
        IL & Imitation Learning \\
        RL & Reinforcement Learning \\
        BC & Behavior Cloning \\
        IRL & Inverse Reinforcement Learning \\
        GCIL & Goal-conditioned Imitation Learning \\
        KB & Knowledge Base \\
        KG & Knowledge Graph \\
        KGE & Knowledge Graph Embedding \\
        PDDL & Planning Domain Definition Language\\
        NLMs & Neural Language Models \\
        PLMs & Pre-trained Language Models \\
        LLMs & Large Language Models \\
        VLMs & Vision-Language Models \\
        VLAs & Vision-Language-Action Models \\
        FMs & Foundation Models \\
        \bottomrule
    \end{tabular}
    \label{tab:abbreviations}
\end{table}

\subsection{Markov decision process} 
A Markov decision process (MDP) is a discrete-time stochastic control model for sequential decision making under uncertainty~\citep{sutton1998reinforcement}. 
Formally, an MDP is a tuple $\mathcal{M}=(\mathcal{S},\mathcal{A},\mathcal{P},\mathcal{R},\gamma)$, where $\mathcal{S}$ is the state space and $\mathcal{A}$ is the action space. 
$\mathcal{P}:\mathcal{S}\times\mathcal{A}\times\mathcal{S}\to[0,1]$ is the transition probability, where $\mathcal{P}(s' \mid s,a)=\Pr\{S_{t+1}=s' \mid S_t=s, A_t=a\}$.
The reward function $\mathcal{R}:\mathcal{S}\times\mathcal{A}\to\mathbb{R}$ gives the expected immediate reward $\mathcal{R}(s,a)=\mathbb{E}[R_{t+1}\mid S_t=s,A_t=a]$.
\textcolor{black}{The discount factor $\gamma\in[0,1]$ encodes the trade-off between near-term and long-term rewards and, in continuing tasks, ensures that the (potentially infinite-horizon) return $\sum_{t=0}^{\infty}\gamma^t R_{t+1}$ is finite.}

\subsection{Reinforcement learning}
\textcolor{black}{
RL~\citep{sutton1998reinforcement} studies how an agent interacts with an environment modeled as an MDP to learn a policy that maximizes expected return \emph{when the environment's dynamics are not available to the agent}. 
Concretely, the transition probability $\mathcal{P}$ and reward function $\mathcal{R}$ are typically \emph{unknown} in RL and the agent must improve its behavior from trial-and-error experience.
A (stochastic) policy is a mapping $\pi:\mathcal{S}\to\mathcal{A}$ with $\pi(a\mid s)$ denoting the probability of taking action $a$ in state $s$. 
}

The goal of RL is to select a policy $\pi^*$ which maximizes expected return $J(\pi)$ when the agent acts according to it. The expected return can be written as 
\begin{equation}
    \begin{aligned}
        J(\pi) = \mathbb{E}_{a_t \sim \pi(\cdot|s_t)
        \atop s_{t+1} \sim P(\cdot|s_t, a_t)}\Big[\sum_t \gamma^t \mathcal{R}(s_t, a_t)\Big] \text{.}
    \end{aligned}
\end{equation}
The central optimization problem in RL can be expressed by 
\begin{equation}
    \begin{aligned}
    \pi^* = \arg \max_\pi J(\pi) \text{,}
    \end{aligned}
\end{equation}
with $\pi^*$ being the optimal policy. RL algorithms typically define value functions
\begin{equation}
    V^{\pi}(s) = \mathbb{E}_{a_t \sim \pi(\cdot|s_t) \atop s_{i+1} \sim \mathcal{P}(\cdot | s_t,a_t)}\Big[\sum_{t}^{\infty} \gamma^t \mathcal{R}(s_t, a_t) | s_0 = s\Big] \text{,}
    \label{eq: state_value_function}
\end{equation}
and action-value functions 
\begin{equation}
    Q^{\pi}(s,a) = \mathbb{E}_{s' \sim \mathcal{P}(\cdot|s,a)}\Big[\mathcal{R}(s,a) + \gamma V^{\pi}(s')\Big] 
\end{equation}
to guide the search for an optimal policy. 

\subsection{Imitation learning}

Imitation Learning (IL) aims to learn a policy $\pi$ by mimicking expert demonstrations, which are sequences of state-action pairs $\tau = \{s_0, a_0, s_1, a_1, \dots\}$, without relying on explicit reward signals. The main IL methodologies include behavioral cloning (BC), goal-conditioned imitation learning (GCIL), and inverse reinforcement learning (IRL).
\subsubsection{Behavioral cloning}
In BC~\citep{bain1995framework}, the trajectory executed by expert agents is treated as the reference or ground truth trajectory. Through supervised learning, an imitation policy is acquired by minimizing the disparity between the anticipated actions and the actual actions observed in the ground-truth trajectory. Considering a set of trajectories are collected from experts $\tau \in \mathcal{T}$, the optimization problem can be defined as:
\begin{equation}
    \hat{\pi}^{*} = \arg \min_{\pi} \sum_{\tau \in \mathcal{T}}\sum_{s \in \tau} L(\pi(s), \pi^*(s))\text{.}
\end{equation}
where $L$ is the cost function, $\pi^*(s)$ and $\pi(s)$ are the expert's and predicted actions at the state $s$, respectively.  

\textcolor{black}{\subsubsection{Goal-conditioned imitation learning}} 
\textcolor{black}{
GCIL extends standard imitation learning by conditioning the policy not only on the state but also on a desired goal, enabling agents to generalize across multiple tasks and achieve diverse outcomes from demonstrations~\citep{ding2020goalconditioned}.
This extension is essential in robotics, where demonstrations often target different configurations, and a goal-conditioned policy $\pi_\theta(a|s,g)$ can leverage these demonstrations to reach any goal $g \in \mathcal{S}$. 
In the goal-conditioned setting, the reward function is defined as an indicator of goal achievement, $r(s_t,a_t,s_{t+1},g)=\mathbbm{1}[s_{t+1}==g]$, meaning that the agent succeeds if its next state matches the goal. 
To learn from demonstrations without explicitly engineering rewards, the most direct approach is goal-conditioned behavioral cloning~\citep{ding2020goalconditioned}, which minimizes the error between the expert’s action $a_t^j$ and the agent’s predicted action $\pi_\theta(s_t^j, g^j)$ given state–goal pairs. 
Here, We assume access to $D$ expert demonstration trajectories $\{(s_0^j, a_0^j, s_1^j, \dots)\}_{j=1}^{D} \sim \tau_{\text{expert}}$, each produced by an expert policy pursuing a goal $g^j$, with $\{g^j\}_{j=1}^{D}$ uniformly sampled from the feasible goal space~\citep{ding2020goalconditioned}.
The supervised loss is:
\begin{equation}
L_{\text{BC}}(\theta,D) = \mathbb{E}_{(s_t^j, a_t^j, g^j) \sim D}\Big[ \, \big\|\pi_\theta(s_t^j,g^j) - a_t^j\big\|_2^2 \Big],
\end{equation}
This formulation directly adapts standard BC to the goal-conditioned case by embedding goals into the input space of the policy.} \textcolor{black}{
Beyond simple cloning, GCIL leverages the insight that trajectories labeled with a particular goal $g^j$ can also serve as valid demonstrations for any intermediate state along the trajectory, effectively enabling goal relabeling. 
For example, a transition tuple $(s_t^j, a_t^j, s_{t+1}^j, g^j)$ can equivalently be relabeled as $(s_t^j, a_t^j, s_{t+1}^j, g' = s_{t+k}^j)$ since reaching $s_{t+k}^j$ is also a valid goal achieved by the expert. 
This relabeling principle, analogous to hindsight experience replay (HER)~\citep{NIPS2017_453fadbd} in RL, substantially improves sample efficiency and generalization in sparse reward settings.
}

\subsubsection{Inverse reinforcement learning}
\textcolor{black}{
An alternative to direct BC in IL is to reason about and recover the hidden reward function that drives expert behavior. This approach, known as IRL~\citep{arora2021survey}, seeks to infer a reward function $R(s,a)$ from demonstrations rather than directly copying actions, thereby capturing the intent behind behavior and enabling agents to generalize or even surpass the expert. 
Formally, given demonstrations represented as trajectories $\tau = \{(s_0,a_0), (s_1,a_1), \dots \}$, the objective of IRL is to find a reward function under which the expert’s policy is (approximately) optimal:
\begin{equation}
\pi_E = \arg\max_\pi ; \mathbb{E}\Big[\sum_{t=0}^\infty \gamma^t R(s_t,a_t),\Big|\pi\Big].
\end{equation}
To address the ill-posed nature of IRL, two representative techniques are widely used. 
Apprenticeship Learning~\citep{abbeel2004apprenticeship} avoids recovering a unique reward and instead matches the expert’s and learner’s feature expectations. 
Assuming a linear reward form $R(s,a) = w^\top \phi(s,a)$, where $w$ is a weight vector and $\phi(s,a)$ is a feature representation of state–action pairs, it seeks a policy $\pi$ such that
\begin{equation}
\mu_\phi(\pi) \approx \mu_\phi(\pi_E),\quad
\mu_\phi(\pi) = \mathbb{E}_\pi\Big[\sum_{t=0}^\infty \gamma^t \phi(s_t,a_t)\Big].
\end{equation}
In contrast, Maximum Entropy IRL~\citep{ziebart2008maximum} resolves ambiguity by modeling expert demonstrations as samples from a maximum entropy trajectory distribution
\begin{equation}
P(\tau) = \frac{1}{Z} \exp\!\Big( \sum_{t} R(s_t,a_t) \Big),
\end{equation}
where $Z$ is a normalization term.
These two approaches capture the main routes in IRL: feature-matching to approximate expert performance and probabilistic modeling to handle reward uncertainty, both of which extend imitation learning beyond BC.
}

\textcolor{black}{\subsection{Diffusion model-based policy learning}}
\textcolor{black}{
Denoising Diffusion Probabilistic Models (DDPMs)~\citep{ho2020denoising} are a class of generative models that define a latent-variable Markov chain to progressively denoise a sample drawn from Gaussian noise into a data sample. The model learns to reverse a fixed forward process that gradually adds Gaussian noise to data according to a variance schedule. Formally, the reverse (generative) process is defined as
\begin{align}
p_\theta(x_{0:T}) &:= p(x_T)\prod_{t=1}^T p_\theta(x_{t-1}\,|\,x_t), \\
p_\theta(x_{t-1}|x_t) &= \mathcal{N}\!\left(x_{t-1}; \mu_\theta(x_t,t), \Sigma_\theta(x_t,t)\right).
\end{align}
Here, $\mu_\theta(x_t,t)$ and $\Sigma_\theta(x_t,t)$ are the mean and variance predicted by a neural network parameterized by $\theta$, which attempt to approximate the true posterior of the forward process.
The forward diffusion process is defined as
\begin{align}
q(x_{1:T}|x_0) &:= \prod_{t=1}^T q(x_t|x_{t-1}), \\
q(x_t|x_{t-1}) &= \mathcal{N}\!\left(x_t;\sqrt{1-\beta_t}\,x_{t-1}, \beta_t I\right),
\end{align}
where $\beta_t \in (0,1)$ is a variance schedule that determines the amount of Gaussian noise injected at step $t$. 
This process admits a closed-form expression for sampling at step $t$:
\begin{align}
q(x_t|x_0) &= \mathcal{N}\!\left(x_t;\sqrt{\bar{\alpha}_t}\,x_0,\,(1-\bar{\alpha}_t)I\right), \\
\bar{\alpha}_t &= \prod_{s=1}^t (1-\beta_s).
\end{align}
Here, $\alpha_t = 1-\beta_t$ represents the retained signal at each step, and the cumulative product $\bar{\alpha}_t$ measures how much of the original data signal $x_0$ remains after $t$ steps of noise addition.
}

\textcolor{black}{
Building on this foundation, recent works in robotics propose Diffusion Policy (DP)~\citep{chi2023diffusion}, which represents visuomotor control as a conditional denoising diffusion process in action space. 
Instead of directly regressing robot actions, the policy refines Gaussian noise into an action sequence by learning the score function of the conditional distribution $p(\bm{A}_t|\bm{O}_t)$, where $\bm{O}_t$ are observations and $\bm{A}_t$ are action trajectories. 
This formulation leverages the advantages of diffusion, such as naturally modeling multimodal distributions, scaling to high-dimensional action sequences, and exhibiting stable training. 
These properties make it highly effective for robot manipulation tasks.
The action generation in diffusion policy follows an iterative denoising process akin to Langevin dynamics~\citep{welling2011bayesian}. 
At inference, starting from a noisy action sequence $\bm{A}_t^K$, the denoising step is given by
\begin{equation}
\bm{A}_t^{k-1} = \alpha \big(\bm{A}_t^k - \gamma\varepsilon_\theta(\bm{O}_t, \bm{A}_t^k, k)\big) + \mathcal{N}(0, \sigma^2 I),
\end{equation}
where $\varepsilon_\theta$ is a neural network that predicts noise conditioned on $\bm{O}_t$, $\gamma$ is the learning rate, $k$ is the iteration step, and $\alpha$ and $\sigma$ follow a noise schedule. 
Training minimizes the mean-squared error between predicted and true noise:
\begin{equation}
L \;=\; \mathbb{E}_{k,\varepsilon}\left[ \,\big\| \varepsilon \;-\; \varepsilon_\theta\!\big(\bm{O}_t,\; \bm{A}_t^{0} + \varepsilon,\; k\big) \big\|_2^{2} \right].
\end{equation}
This process allows the model to iteratively refine noise into structured action sequences that are both temporally consistent and reactive. 
As a result, diffusion-based approaches have emerged as a powerful paradigm for trajectory generation and visuomotor policy learning in robotics, bridging generative modeling and control.
}

\subsection{Knowledge base \& Knowledge graph}
A knowledge base (KB) serves as a foundational repository of structured knowledge, encapsulating facts, rules, and relationships pertinent to a given domain. Information is organized in a structured format in a knowledge base, facilitating efficient retrieval and inference. 
Formally, knowledge bases often employ languages such as Resource Description Framework (RDF) or Web Ontology Language (OWL) to encode knowledge in a machine-readable form.

A Knowledge Graph (KG) can be considered as a specialized form of KB with a graph structure. Formally, a Knowledge Graph (KG) is defined as $\mathcal{G}\subseteq \mathcal{E}\times \mathcal{R}\times \mathcal{E}$ over a set $\mathcal{E}$ of entities and a set $\mathcal{R}$ of predicates. 
In the field of robot manipulation, entities often contain real-world objects, actions, skills, and other abstract concepts; relations represent the relations between these entities, such as \emph{isComponentOf}, \emph{withForce}, and \emph{hasPose}. Several well-known knowledge graphs (KGs) have been developed in this field, as highlighted in various works~\citep{RoboEarth,diab2019pmk,beetz2018know,kwak2022semantic}.

Knowledge Graph Embedding (KGE) methods represent the information from a KG as dense embeddings. Entities and predicates in the KG are mapped into a $d$-dimensional vector $M_\theta: \mathcal{E}\cup \mathcal{R}\rightarrow \mathbb{R}^d$. 
These methods can be either score function based~\citep{bordes2013translating,wang2014knowledge,lin2015learning} or graph neural network based~\citep{schlichtkrull2018modeling, nathani2019learning}.

\textcolor{black}{
\subsection{Language and Multimodal Fusion}
Language-conditioned robot manipulation inherently requires integrating linguistic instructions with perceptual observations and robot states \citep{rth2024rss}. 
This process relies on multimodal fusion, aiming to learn joint representations that align and combine information from heterogeneous modalities such as vision and language. 
In general multimodal learning, fusion strategies are often categorized according to how interactions between modalities are modeled during representation learning~\citep{baltruvsaitis2018multimodal}. 
Three representative paradigms have emerged in recent vision-language multimodal frameworks. 
The first paradigm adopts \emph{dual-stream architectures} in which visual and textual features are encoded by separate networks and interact through cross-modal attention mechanisms, enabling fine-grained alignment between modalities, as demonstrated in models such as ViLBERT~\citep{lu2019vilbert}. 
The second paradigm employs \emph{single-stream architectures}, where tokens from different modalities are projected into a shared embedding space and processed by a unified transformer encoder, allowing deeper cross-modal interactions and joint reasoning~\citep{chen2020uniter}. 
The third paradigm focuses on \emph{contrastive alignment}, which learns a shared embedding space by maximizing the similarity between paired vision and language representations, a strategy popularized by models such as CLIP~\citep{radford2021learning}. 
These multimodal fusion mechanisms provide the foundation for modern vision--language models and vision--language--action models, which ground language instructions in visual observations and enable robots to translate high-level semantic commands into executable manipulation behaviors.
}

\subsection{Language model}

Language models are designed to estimate the likelihood of sequences of words in human language. They learn patterns, structures, and semantics from vast amounts of textual data, enabling them to understand and predict language usage. In recent years, neural language models have advanced significantly, allowing for generating coherent and contextually relevant text. We provide a brief history of neural language models.

\noindent \textbf{Neural Language/Lexical Models (NLMs):} NLMs~\citep{bengio2000neural,mikolov2010recurrent,kombrink2011recurrent} leverage neural networks to estimate the probability of word sequences, e.g., recurrent neural networks (RNNs), offering a more powerful alternative to traditional statistical methods.
\citet{collobert2011natural} make a remarkable contribution by developing a unified neural network approach capable of handling various Natural Language Processing (NLP) tasks within a single framework, demonstrating the versatility of neural models in language understanding. 
Furthermore,  word2vec  \citep{mikolov2013distributed,mikolov2013efficient} revolutionizes word representations employing a simple yet efficient shallow neural network to learn distributed word embeddings. 
These embeddings have proven to be highly effective across a wide range of NLP tasks and have been instrumental in advancing applications like language-conditioned robot manipulation by enhancing semantic understanding.
    
\noindent \textbf{Pre-trained Language Models (PLMs):} PLMs are an early attempt to extract semantic meaning from natural language. ELMo~\citep{peters-etal-2018-deep} aims to capture context-aware word representations by first pre-training a bidirectional LSTM (biLSTM) network and then fine-tuning the biLSTM network according to specific downstream tasks. 
Moreover, drawing inspiration from the Transformer architecture~\citep{vaswani2017attention} and incorporating self-attention mechanisms, BERT~\citep{devlin2018bert} takes language model pre-training a step further. 
It accomplishes this by conducting bidirectional pre-training exercises on extensive unlabeled text corpora. 
These specially crafted pre-training tasks imbue BERT with contextual understanding.
    
\noindent \textbf{Large Language Models:} LLMs become popular as scaling of PLMs leads to improved performance on downstream tasks. 
Many researchers~\citep{hoffmann2022training} attempt to study the performance limit of PLMs by scaling the size of models and datasets, e.g., the comparatively small 1.5B parameter GPT-2~\citep{radford2019language} versus larger 175B GPT-3~\citep{brown2020language} and 540B PaLM~\citep{chowdhery2022palm}. 
Although these models share similar architectures, larger models exhibit enhanced capabilities such as few-shot learning, in-context learning, and improved performance on language understanding and generation benchmarks. 
Additionally, models like ChatGPT have adapted the GPT family for dialogue by incorporating techniques such as instruction tuning and reinforcement learning from human feedback~\citep{ouyang2022training}. 
This results in more coherent and contextually appropriate conversational abilities, which are crucial for interactive robot manipulation tasks where ongoing dialogue between the human and robot can enhance task execution and adaptability. 
By integrating LLMs into robotic systems, researchers~\citep{ding2022robot, kant2022housekeep} aim to leverage their advanced language understanding to enable robots to handle challenging tasks by reasoning and inferring missing information. 
This contributes to developing more robust and flexible manipulators that can operate effectively in unstructured real-world environments~\citep{lin2023text2motion,ren2023robots,wu2023tidybot}.
    
\noindent \textbf{Vision-Language Models:} VLMs play a key role in extracting and combining visual and textual content from large-scale web data. This paradigm equips the agent with an expansive understanding of the world. 
Seminal models in this domain, such as CLIP~\citep{radford2021learning}, Flamingo~\citep{alayrac2022flamingo},  PaLM-E~\citep{driess2023palme}, and PaLI-X~\citep{chen2023pali} underscore the potential of VLMs. 
Integrating these pre-trained VLMs into robotic systems makes it feasible for robots to tackle diverse tasks in real-world scenarios.

\noindent \textbf{Vision-Language-Action Models}: \textcolor{black}{In this work, we define vision-language-action (VLA) models as unified embodied policy models that integrate the three modalities of language, visual observations, and robot actions within a shared or tightly coupled architecture. In such models, language, images, and actions are represented in a common modeling framework, for example, as text tokens, visual tokens, and action tokens (Gato~\citep{reed2022a}, RT-1~\citep{brohan2022rt}, and RT-2~\citep{brohan2023rt}). 
More recent VLAs generalize this idea from discrete action tokens to continuous action-token embeddings or diffusion-/flow-based action representations, enabling trajectory-level or low-level control generation~\citep{kim2024openvlaopensourcevisionlanguageactionmodel, black2024pi_0,wen2025diffusionvla}.}

\section{Taxonomy of language-conditioned methods for robotic manipulation}
\input{tikz/overview}
\label{sec:Taxonomy}

Over the past few years, there has been a growing interest in leveraging natural language to enhance robotic manipulation tasks, particularly in making these systems more intuitive and accessible during interactions. To facilitate a clear review of these approaches, this section outlines the taxonomy employed in the narration. \textcolor{black}{While robotic learning in bridging language instructions and robot actions is often categorized by the algorithmic paradigm, such as RL, IL, or planning, this can obscure the specific and varied roles that language plays. For instance, an RL agent might use language to shape its reward function or, in a completely different manner, to directly condition its policy learning. Although both involve RL methods, the function of language is fundamentally different. To provide a clearer, more orthogonal taxonomy, we structure this survey around the primary ways language is integrated into robot systems.}

\textcolor{black}{As illustrated in the fine-grained taxonomy tree in Figure \ref{fig:overview}, our classification logic progresses from macro-level functional roles down to specific algorithmic implementations and evolutionary process. The four primary categories and their underlying classification logic are detailed below:}

\begin{itemize}[nosep,noitemsep]
    \item \textcolor{black}{\textbf{Language for state evaluation (Section \ref{sec:language-for-state-evaluation}):} Language is used to define goals and quantify task progress, by converting text into \textcolor{black}{numerical feedback signals (reward or cost)}. The policy is then optimized against these signals. The core question is: How can language specify whether a state or outcome is desirable (feedback \textit{how the progress})?} \textcolor{black}{This category is subdivided based on the type of algorithm receiving the signal and the technological evolution of the signal generation:}
        \begin{itemize}[nosep,noitemsep]
            \item \textcolor{black}{\textbf{Reward Functions}: For learning-based agents (RL), language is translated into rewards. We trace the logical evolution of this process from manual Reward Designing (sparse vs. dense), to data-driven Reward Learning (like Inverse RL), and finally to modern Foundation Model (FM)-driven automated reward generation.}
            \item \textcolor{black}{\textbf{Cost Functions}: For optimization-based motion planners, language is translated into cost maps. The narrative trajectory moves from specific linguistic-to-cost mappings to automated 3D cost map generation empowered by FMs.}
        \end{itemize}
    \item \textbf{Language as a policy condition (Section \ref{sec:language-as-policy-condition}):} Language is used as \textcolor{black}{an explicit conditioning signal for} a policy that maps observations to actions. \textcolor{black}{The primary learning target is a control policy, and language specifies the desired behavior for the current episode or timestep. Although such methods may use multimodal encoders, language remains an external task condition rather than the organizing principle of the whole architecture.} The core question is: \textcolor{black}{How can language condition a policy to produce the correct behavior?} \textcolor{black}{This category is subdivided by the underlying algorithmic paradigms used to learn this language-conditioned mapping:}
        \begin{itemize}[nosep,noitemsep]
            \item \textcolor{black}{\textbf{Reinforcement Learning}: Integrates language to solve trial-and-error exploration, evolving from simple goal-conditioning to lifelong multi-task learning.}
            \item \textcolor{black}{\textbf{Behavioral Cloning}: Learns directly from expert demonstrations to bypass the sample inefficiency and complex reward engineering inherent in RL.}
            \item \textcolor{black}{\textbf{Diffusion-based Policy}: Represents the latest generative evolution to address BC's limitation in handling multi-modal action distributions (traditional BC averages out expert behaviors).}
        \end{itemize}
    \item \textcolor{black}{\textbf{Language for cognitive planning and reasoning (Section \ref{sec:language-cognitive-planning-reasoning}):} Language serves as an internal reasoning medium for decomposing complex tasks and forming strategies. Core question: How a robot ``\textit{thinks}'' in the language space to structure its own behavior, i.e., utilizing language to reason about its goals and plan its actions.} \textcolor{black}{This branch is subdivided based on the type of cognitive system used to bridge abstract reasoning and perceptual reality:}
        \begin{itemize}[nosep,noitemsep]
            \item \textcolor{black}{\textbf{Classic Neuro-symbolic approaches}: The foundational methods that use language to bridge explicit symbolic logic (like Knowledge Graphs) with neural perception.}
            \item \textcolor{black}{\textbf{Empowered by Large Language Models}: Approaches that replace rigid symbolic systems with LLMs for open-vocabulary text reasoning and code generation, often without task-specific retraining for the planning module.}
            \item \textcolor{black}{\textbf{Empowered by Vision-Language Models}: \\Methods that resolve the ``blindness'' of pure LLMs by directly grounding textual reasoning in visual observations.}
        \end{itemize}
    \item \textcolor{black}{\textbf{Language in unified vision-language-action models (Section \ref{sec:vla}):} 
    Language is not used merely as an external condition; instead, it is jointly modeled with visual observations, language and robot actions within \textcolor{black}{an embodied policy architecture which unifies the modalities of images, text, and actions}. In such a system, perception, semantic grounding, and action generation are increasingly learned within one embodied foundation model, often by extending pretrained VLMs/LLMs with robot action representations. 
    The core question is: How can vision (perception), language (semantic grounding), and action be unified into a single scalable model for embodied decision-making? 
    This branch is subdivided based on different optimization directions for bridging language and action:}
        \begin{itemize}
            \item \textcolor{black}{\textbf{Perception}: Methods that focus on optimizing how VLAs perceive and understand their environment.}
            \item \textcolor{black}{\textbf{Reasoning}: Approaches that enhance the model's internal ``thought process'', that is, how it forms plans, leverages prior knowledge, and predicts outcomes to solve complex tasks.}
            \item \textcolor{black}{\textbf{Action}: Methods that focus on the optimization regarding the ``output'' stage of the policy, concerning the form and mechanism of the robot's actions. This bridges the gap between the model's internal plan and its physical embodiment.}
            \item \textcolor{black}{\textbf{Adaptation}: Techniques that adapt pretrained VLAs to new tasks, objects, or scenes while preserving or improving the grounding between language instructions and executable actions.}
        \end{itemize}
\end{itemize}

\textcolor{black}{Each subsequent section explores the advanced techniques developed within these sub-branches, highlighting their unique contributions, chronological evolution, and the specific challenges they aim to resolve.} 

\hypertarget{language-for-state-evaluation}{}

\section{\textcolor{black}{Language for state evaluation}}
\label{sec:language-for-state-evaluation}

\textcolor{black}{Traditionally, specifying a robot's objective has required significant expert engineering, such as hand-crafting dense reward functions~\citep{eschmann2021reward} or defining precise goal coordinates in the state space~\citep{la2011motion}. This process is not only labor-intensive but also rigid, and the robot cannot easily generalize to new goals without being reprogrammed. Natural language provides a powerful solution to this limitation. It offers a flexible, intuitive, and generalizable interface, enabling non-experts to communicate a vast range of complex, abstract, or compositional goals in multi-task scenarios without programming efforts~\citep{tellex2020robots}, such as from ``\textit{pick up the red block}'' to ``\textit{tidy up the table}''. This shifts the paradigm from low-level programming to high-level human-centric goal specification.}
\begin{figure}[tb!]
    \centering
      \begin{adjustbox}{width=0.49\textwidth}
        \begin{tikzpicture}
              \def\cellwidth{0.75}
              \def\longcellwidth{1.5}
              \def\cellheight{0.75}
            
              \node (cell11) [draw=JungleGreen, minimum width=\cellwidth cm, minimum height=\cellheight cm] at (0, 0) {\sstar};
            
              \node (cell12) [draw=JungleGreen, minimum width=\cellwidth cm, minimum height=\cellheight cm, right = 0cm of cell11, fill=JungleGreen!20] {+1};
              \node (cell13) [draw=JungleGreen, minimum width=\cellwidth cm, minimum height=\cellheight cm, right=0cm of cell12, fill=JungleGreen!30] {+2};
              \node (cell14) [draw=JungleGreen, minimum width=\cellwidth cm, minimum height=\cellheight cm, right=0cm of cell13, fill=JungleGreen!40] {...};
              
              \node (cell21) [draw=JungleGreen, minimum width=\cellwidth cm, minimum height=\cellheight cm, below=0cm of cell11, fill=JungleGreen!20] {+1};
              \node (cell22) [draw=JungleGreen, minimum width=\cellwidth cm, minimum height=\cellheight cm, right=0cm of cell21, fill=JungleGreen!30] {+2};
              \node (cell23) [draw=JungleGreen, minimum width=\cellwidth cm, minimum height=\cellheight cm, right=0cm of cell22, fill=JungleGreen!40] {...};
              \node (cell24) [draw=JungleGreen, minimum width=\cellwidth cm, minimum height=\cellheight cm, right=0cm of cell23, fill=JungleGreen!50] {};
              
              \node (cell31) [draw=JungleGreen, minimum width=\cellwidth cm, minimum height=\cellheight cm, below=0cm of cell21, fill=JungleGreen!30] {+2};
              \node (cell32) [draw=JungleGreen, minimum width=\cellwidth cm, minimum height=\cellheight cm, right=0cm of cell31, fill=JungleGreen!40] {...};
              \node (cell33) [draw=JungleGreen, minimum width=\cellwidth cm, minimum height=\cellheight cm, right=0cm of cell32, fill=JungleGreen!50] {};
              \node (cell34) [draw=JungleGreen, minimum width=\cellwidth cm, minimum height=\cellheight cm, right=0cm of cell33, fill=JungleGreen!60] {};
              
              \node (cell41) [draw=JungleGreen, minimum width=\cellwidth cm, minimum height=\cellheight cm, below=0cm of cell31, fill=JungleGreen!40] {...};
              \node (cell42) [draw=JungleGreen, minimum width=\cellwidth cm, minimum height=\cellheight cm, right=0cm of cell41, fill=JungleGreen!50] {};
              \node (cell43) [draw=JungleGreen, minimum width=\cellwidth cm, minimum height=\cellheight cm, right=0cm of cell42, fill=JungleGreen!60] {};
              \node (cell44) [draw=JungleGreen, minimum width=\cellwidth cm, minimum height=\cellheight cm, right=0cm of cell43, fill=JungleGreen!70] {\small +1k};

            \node(label1)[below=0.3cm of cell42, xshift=0.375cm]{(a)};
              \node (cell11) [draw=JungleGreen, minimum width=\cellwidth cm, minimum height=\cellheight cm] at (3.5, 0) {\sstar};
            
              \node (cell12) [draw=JungleGreen, minimum width=\cellwidth cm, minimum height=\cellheight cm, right = 0cm of cell11] {0};
              \node (cell13) [draw=JungleGreen, minimum width=\cellwidth cm, minimum height=\cellheight cm, right=0cm of cell12] {0};
              \node (cell14) [draw=JungleGreen, minimum width=\cellwidth cm, minimum height=\cellheight cm, right=0cm of cell13] {0};
              
              \node (cell21) [draw=JungleGreen, minimum width=\cellwidth cm, minimum height=\cellheight cm, below=0cm of cell11] {0};
              \node (cell22) [draw=JungleGreen, minimum width=\cellwidth cm, minimum height=\cellheight cm, right=0cm of cell21] {0};
              \node (cell23) [draw=JungleGreen, minimum width=\cellwidth cm, minimum height=\cellheight cm, right=0cm of cell22] {0};
              \node (cell24) [draw=JungleGreen, minimum width=\cellwidth cm, minimum height=\cellheight cm, right=0cm of cell23] {0};
              
              \node (cell31) [draw=JungleGreen, minimum width=\cellwidth cm, minimum height=\cellheight cm, below=0cm of cell21] {0};
              \node (cell32) [draw=JungleGreen, minimum width=\cellwidth cm, minimum height=\cellheight cm, right=0cm of cell31] {0};
              \node (cell33) [draw=JungleGreen, minimum width=\cellwidth cm, minimum height=\cellheight cm, right=0cm of cell32] {0};
              \node (cell34) [draw=JungleGreen, minimum width=\cellwidth cm, minimum height=\cellheight cm, right=0cm of cell33] {0};
              
              \node (cell41) [draw=JungleGreen, minimum width=\cellwidth cm, minimum height=\cellheight cm, below=0cm of cell31] {0};
              \node (cell42) [draw=JungleGreen, minimum width=\cellwidth cm, minimum height=\cellheight cm, right=0cm of cell41] {0};
              \node (cell43) [draw=JungleGreen, minimum width=\cellwidth cm, minimum height=\cellheight cm, right=0cm of cell42] {0};
              \node (cell44) [draw=JungleGreen, minimum width=\cellwidth cm, minimum height=\cellheight cm, right=0cm of cell43, fill=JungleGreen!70] {\small +1k};
            \node(label1)[below=0.3cm of cell42, xshift=0.375cm]{(b)};
            
              \node (cell11) [draw=JungleGreen, minimum width=\longcellwidth cm, minimum height=\cellheight cm] at (7.4, 0) {\sstar};
            
              \node (cell12) [draw=JungleGreen, minimum width=\longcellwidth cm, minimum height=\cellheight cm, right=0cm of cell11, fill=JungleGreen!10] {\small $f(s_{01},g)$};
              \node (cell13) [draw=JungleGreen, minimum width=\longcellwidth cm, minimum height=\cellheight cm, right=0cm of cell12, fill=JungleGreen!10] {\small $f(s_{02},g)$};
              \node (cell14) [draw=JungleGreen, minimum width=\longcellwidth cm, minimum height=\cellheight cm, right=0cm of cell13, fill=JungleGreen!15] {\small $f(s_{03},g)$};
              
              \node (cell21) [draw=JungleGreen, minimum width=\longcellwidth cm, minimum height=\cellheight cm, below=0cm of cell11, fill=JungleGreen!10] {\small $f(s_{10},g)$};
              \node (cell22) [draw=JungleGreen, minimum width=\longcellwidth cm, minimum height=\cellheight cm, right=0cm of cell21, fill=JungleGreen!20] {\small $f(s_{11},g)$};
              \node (cell23) [draw=JungleGreen, minimum width=\longcellwidth cm, minimum height=\cellheight cm, right=0cm of cell22, fill=JungleGreen!30] {\small $f(s_{12},g)$};
              \node (cell24) [draw=JungleGreen, minimum width=\longcellwidth cm, minimum height=\cellheight cm, right=0cm of cell23, fill=JungleGreen!20] {\small $f(s_{13},g)$};
              
              \node (cell31) [draw=JungleGreen, minimum width=\longcellwidth cm, minimum height=\cellheight cm, below=0cm of cell21, fill=JungleGreen!25] {\small $f(s_{20},g)$};
              \node (cell32) [draw=JungleGreen, minimum width=\longcellwidth cm, minimum height=\cellheight cm, right=0cm of cell31, fill=JungleGreen!50] {\small $f(s_{21},g)$};
              \node (cell33) [draw=JungleGreen, minimum width=\longcellwidth cm, minimum height=\cellheight cm, right=0cm of cell32, fill=JungleGreen!60] {\small $f(s_{22},g)$};
              \node (cell34) [draw=JungleGreen, minimum width=\longcellwidth cm, minimum height=\cellheight cm, right=0cm of cell33, fill=JungleGreen!50] {\small $f(s_{23},g)$};
              
              \node (cell41) [draw=JungleGreen, minimum width=\longcellwidth cm, minimum height=\cellheight cm, below=0cm of cell31, fill=JungleGreen!40] {\small $f(s_{30},g)$};
              \node (cell42) [draw=JungleGreen, minimum width=\longcellwidth cm, minimum height=\cellheight cm, right=0cm of cell41, fill=JungleGreen!45] {\small $f(s_{31},g)$};
              \node (cell43) [draw=JungleGreen, minimum width=\longcellwidth cm, minimum height=\cellheight cm, right=0cm of cell42, fill=JungleGreen!60] {\small $f(s_{32},g)$};
              \node (cell44) [draw=JungleGreen, minimum width=\longcellwidth cm, minimum height=\cellheight cm, right=0cm of cell43, fill=JungleGreen!70] {\small $f(s_{33},g)$};
              \node(label1)[below=0.3cm of cell42, xshift=0.75cm]{(c)};
            \end{tikzpicture}
     \end{adjustbox}
    \caption{An illustration of different reward schemes. (a) Dense reward: \textcolor{black}{The agent receives a gradually increasing reward as it approaches the goal, providing continuous guidance.} (b) Sparse reward: \textcolor{black}{The agent only receives a large reward upon reaching the final goal state.} (c) Reward function learning: \textcolor{black}{A function is learned to map state-goal pairs to a continuous reward value, creating a smooth reward gradient.}  \sstar~the starting position.} 
    \label{fig:rewards}
    \vspace{-1em}
\end{figure}
\textcolor{black}{This brings us to the first key research question: How can we use language to quantify task progress? The core idea is to translate a language instruction into a quantitative scoring function that evaluates how well a robot's state or action aligns with the desired outcome, guiding the robot towards desired behaviors efficiently. This numerical signal, which can serve as a reward for RL agents or a cost for planners, provides the essential feedback for the robot to learn or plan effectively. Grounding language in state-space valuations is a central challenge and can be implemented in two primary ways.}
\begin{itemize}
    \item \textcolor{black}{Language-conditioned reward functions: In RL, the language-derived score serves as a reward function. It guides the agent's trial-and-error learning, reinforcing behaviors that bring it closer to fulfilling the instruction. These methods can be further categorized by the numerical properties of the reward signal, such as dense rewards, sparse rewards, or fully learned reward functions, as illustrated in Figure~\ref{fig:rewards}.}
    \item \textcolor{black}{Language-conditioned cost functions: In task and motion planning, the score serves as a cost function. It guides a search algorithm to find an optimal sequence of actions that minimizes the cost, thereby achieving the goal specified in the language command. For example, for command ``\textit{Pick up the apple, but stay away from the vase}'', the cost function would assign a high penalty to any trajectory that nears the vase. A motion planner would then search for a path that minimizes this total cost, resulting in a trajectory that safely navigates around the vase to reach the apple.}
\end{itemize}

\textcolor{black}{This section reviews the methods developed to address this problem, including Language-conditioned reward designing/learning in Sec. 4.1 and Language-conditioned cost functions in Sec. 4.2. The taxonomy illustrated in Figure~\ref{fig:taxonomy-sec4}.} \textcolor{black}{In addition, to provide a clearer overview of how language is utilized to quantify task progress, Table \ref{tab:language_state_evaluation} summarizes representative state-of-the-art methods discussed in this section. These approaches are categorized by the type of algorithmic signal they generate, such as rewards for reinforcement learning or cost functions for motion planning. Furthermore, the table traces the technological evolution of these signals, moving from manual reward design and data-driven learning to the recent integration of foundation models that automate signal generation. This comparison explicitly highlights the core mechanisms, key advantages, and inherent limitations of each state-evaluation paradigm.}

\begin{figure}[ht!]
    \centering
    \resizebox{.49\textwidth}{!}{
    \begin{tikzpicture}
    \selectcolormodel{cmyk}
    \hypersetup{linkcolor=black}    
    \node[circle, draw=RoyalBlue!50, fill=RoyalBlue!50, text width=3cm, align=center, minimum size=3cm]
        (title) at (0,0) {\Large\textbf{\textsf{4. Language for State Evaluation}}};
    
    \node[circle, draw=YellowOrange!50, fill=YellowOrange!50, text width=2.5cm, align=center, minimum size=2.5cm, left]
        (title-41) at ([xshift=-2cm]title.west) {\Large\textbf{\textsf{4.1 Reward Functions}}};
    \node[circle, fill=YellowOrange!30, draw=YellowOrange!30, text width=2.5cm, align=center, minimum size=2.2cm, above]
        (title-41-1) at ([xshift=3cm, yshift=.5cm]title-41.north) {\hyperlink{lang-rew-sig-design}{\Large\textbf{\textsf{4.1.1\\Reward Design}}}};
    \node[circle, fill=YellowOrange!30, draw=YellowOrange!30, text width=2.5cm, align=center, minimum size=2.2cm, left]
        (title-41-2) at ([xshift=-2cm]title-41.west) {\hyperlink{reward-learning}{\Large\textbf{\textsf{4.1.2\\Reward Learning}}}};
    \node[circle, fill=YellowOrange!30, draw=YellowOrange!30, text width=2.5cm, align=center, minimum size=2.2cm, below]
        (title-41-3) at ([xshift=3cm, yshift=-.5cm]title-41.south) {\hyperlink{FM-driven-reward-learning}{\Large\textbf{\textsf{4.1.3\\FM-driven Learning}}}};
    
    \node[circle, fill=YellowOrange!15, draw=YellowOrange!15, text width=2.5cm, align=center, minimum size=2.0cm, above right]
        (title-41-1-1) at ([xshift=2cm, yshift=-.5cm]title-41-1.north east) {\Large\textbf{\textsf{Dense Reward}}};
    \node[circle, fill=YellowOrange!15, draw=YellowOrange!15, text width=2.5cm, align=center, minimum size=2.0cm, above left]
        (title-41-1-2) at ([xshift=-2cm, yshift=-.5cm]title-41-1.north west) {\Large\textbf{\textsf{Sparse Reward}}};
    
    \node[circle, fill=YellowOrange!15, draw=YellowOrange!15, text width=2.5cm, align=center, minimum size=2.0cm, above]
        (title-41-2-1) at ([xshift=-1.5cm, yshift=1cm]title-41-2.north) {\Large\textbf{\textsf{Direct}}};
    \node[circle, fill=YellowOrange!15, draw=YellowOrange!15, text width=2.5cm, align=center, minimum size=2.0cm, below]
        (title-41-2-2) at ([xshift=-1.5cm, yshift=-1cm]title-41-2.south) {\Large\textbf{\textsf{IRL}}};
    
    \node[circle, fill=YellowOrange!15, draw=YellowOrange!15, text width=2.5cm, align=center, minimum size=2.0cm, below right]
        (title-41-3-1) at ([xshift=2cm, yshift=.5cm]title-41-3.south east) {\Large\textbf{\textsf{LLM Code Gen.}}};
    \node[circle, fill=YellowOrange!15, draw=YellowOrange!15, text width=2.5cm, align=center, minimum size=2.0cm, below left]
        (title-41-3-2) at ([xshift=-2cm, yshift=.5cm]title-41-3.south west) {\Large\textbf{\textsf{VLM-dirven Learning}}};

    \node[circle, draw=JungleGreen!50, fill=JungleGreen!50, text width=2.5cm, align=center, minimum size=2.5cm, right]
        (title-42) at ([xshift=2cm]title.east) {\hyperlink{cost-function-mapping}{\Large\textbf{\textsf{4.2 Cost Functions}}}};
    \node[circle, draw=JungleGreen!30, fill=JungleGreen!30, text width=2.5cm, align=center, minimum size=2.0cm, above]
        (title-42-1) at ([xshift=1.5cm, yshift=1cm]title-42.north) {\Large\textbf{\textsf{Specific Text2value Mapping}}};
    \node[circle, draw=JungleGreen!30, fill=JungleGreen!30, text width=2.5cm, align=center, minimum size=2.0cm, below]
        (title-42-2) at ([xshift=1.5cm, yshift=-1cm]title-42.south) {\Large\textbf{\textsf{FM-driven Cost Mapping}}};

    \path (title) to[circle connection bar switch color=from (RoyalBlue!50) to (YellowOrange!50)] (title-41);
    \path (title-41) to[circle connection bar switch color=from (YellowOrange!50) to (YellowOrange!30)] (title-41-1);
    \path (title-41) to[circle connection bar switch color=from (YellowOrange!50) to (YellowOrange!30)] (title-41-2);
    \path (title-41) to[circle connection bar switch color=from (YellowOrange!50) to (YellowOrange!30)] (title-41-3);
    \path (title-41-1) to[circle connection bar switch color=from (YellowOrange!30) to (YellowOrange!15)] (title-41-1-1);
    \path (title-41-1) to[circle connection bar switch color=from (YellowOrange!30) to (YellowOrange!15)] (title-41-1-2);
    \path (title-41-2) to[circle connection bar switch color=from (YellowOrange!30) to (YellowOrange!15)] (title-41-2-1);
    \path (title-41-2) to[circle connection bar switch color=from (YellowOrange!30) to (YellowOrange!15)] (title-41-2-2);
    \path (title-41-3) to[circle connection bar switch color=from (YellowOrange!30) to (YellowOrange!15)] (title-41-3-1);
    \path (title-41-3) to[circle connection bar switch color=from (YellowOrange!30) to (YellowOrange!15)] (title-41-3-2);

    \path (title) to[circle connection bar switch color=from (RoyalBlue!50) to (JungleGreen!50)] (title-42);
    \path (title-42) to[circle connection bar switch color=from (JungleGreen!50) to (JungleGreen!30)] (title-42-1);
    \path (title-42) to[circle connection bar switch color=from (JungleGreen!50) to (JungleGreen!30)] (title-42-2);

    \node[below](idea)at([yshift=-2cm]title-41-3.south){\Large \textbf{\textsf{Core idea:}} \textsf{Language to ``quantify task progress''.}};
    \node[below](role)at([yshift=-.1cm]idea.south){\Large \textbf{\textsf{Method:}} \textsf{Translate language into quantifiable functions.}};
        
    \end{tikzpicture}
    }
    \caption{\textcolor{black}{Taxonomy of Sec. \ref{sec:language-for-state-evaluation} Language for state evaluation. 
    }}
    \label{fig:taxonomy-sec4}
    \vspace{-1em}
\end{figure}

\subsection{\textcolor{black}{Language-conditioned reward designing/learning}} 

\textcolor{black}{In many RL scenarios, particularly for complex manipulation tasks, agents often learn from sparse rewards, which provide a positive signal only upon task completion~\citep{NIPS2017_453fadbd, riedmiller2018learning, 9772990}. This is because defining a continuous measure of progress for abstract or contact-rich tasks like ``folding a shirt''~\citep{jangir2020dynamic} or ``inserting a key''~\citep{ocana2023overview} is difficult, as their success is often binary and depends on a complex combination of factors that are hard to quantify. This approach is sample-inefficient, as the agent may struggle to discover the goal through random exploration, resulting in long learning time. On the contrary, dense rewards, such as the distance to a target, provide stronger learning signals but are difficult to specify, and often require intensive manual engineering and domain expertise for each new task~\citep{sutton1998reinforcement}. A common technique to accelerate learning is reward shaping~\citep{ng1999policy}, which provides the agent with additional, intermediate rewards to guide it toward the goal. However, designing these shaping functions can also be a challenging and time-consuming process in multitask scenarios~\citep{yu2020meta}. Using natural language instructions offers a more intuitive solution, instead of requiring an expert to engineer a complex function, anyone can provide simple instructions, like ``\textit{Jump over the skull while going to the left}'', to specify desired behaviors~\citep{goyal2019using}. These instructions can then be translated into intermediate language-based rewards by a pre-trained language-action matching model, guiding the agent's exploration and accelerate learning. Language-conditioned reward designing/learning approaches make it convenient for non-experts to teach new skills for RL agents.}

\begin{table*}[htbp]
    \centering
    \caption{\textcolor{black}{Comparison of representative state-of-the-art methods for \textbf{Section \ref{sec:language-for-state-evaluation} Language for state evaluation}. The table highlights how different algorithms translate language into quantitative signals (rewards for reinforcement learning or costs for motion planning) , tracing the evolution from manual design and data-driven learning to automated foundation model-driven generation.}}
    \label{tab:language_state_evaluation}
    \small
    \begin{tabular}{p{0.12\linewidth}| p{0.1\linewidth}| p{0.21\linewidth}| p{0.21\linewidth}| p{0.21\linewidth}}
        \toprule[1pt]
        \textbf{Method} & \textbf{Signal\quad Category} & \textbf{Core Mechanism} & \textbf{Key Advantages} & \textbf{Key Disadvantages} \\
        \midrule[0.2pt]
        ZSRM\quad\citep{pmlr-v162-mahmoudieh22a} & Sparse Reward Design & Design reward by calculating similarity between a camera image and goal text using CLIP. & Enables reward specification for some new goals without additional reward-model training. & Performance is fundamentally limited by the CLIP model's spatial reasoning abilities. \\
        \midrule[0.2pt]
        PixL2R\quad\citep{goyal2021pixl2r} & Dense Reward Design & Learns a relatedness model from paired data (language and trajectory) to generate a dense shaping reward. & Significantly improves policy learning efficiency by providing continuous guidance. & Relies on absolute scores as ground truth and requires diverse language datasets. \\
        \midrule[0.2pt]
        LOREL\qquad\citep{nair2022learning} & Reward Learning & Implements a binary classifier trained on offline expert demonstrations. & Automates the translation of visual observations and instructions into reward signals. & Struggles with complex manipulation tasks involving fine-grained behaviors or multiple objects. \\
        \midrule[0.2pt]
        Text2Reward\quad\citep{xie2024textreward} & FM-driven Reward Generation & Uses LLMs to write dense Pythonic reward functions directly from natural language goals. & Enables autonomous reward generation and matches expert rewards on simulated tasks. & Relies heavily on the correctness of generated code and requires a privileged simulator state. \\
        \midrule[0.2pt]
        ReWiND\quad\citep{zhang:rewind} & VLM-driven Reward Learning & Learns a progress-predicting reward from language-annotated demonstrations and generated failures. & Improves success rates significantly in both simulation and real bi-manual setups. & Requires a handful of demonstrations to learn a general reward.\\        
        \bottomrule[1pt]
    \end{tabular}
\end{table*}

\hypertarget{sparse-reward}{}
\hypertarget{dense-reward}{}

\subsubsection{\textcolor{black}{Language-based reward signal designing}}
\hypertarget{lang-rew-sig-design}{}

\textcolor{black}{Language provides a natural and expressive medium for designing reward signals that guide robot learning. Instead of manually engineering complex reward functions, language allows users to specify desired behaviors, goals, or preferences intuitively. These signals can be broadly categorized into two types: \emph{sparse reward} and \emph{dense reward}, as illustrated in Figure~\ref{fig:rewards}. Each type presents a distinct trade-off between design simplicity and learning efficiency. A \emph{sparse reward} provides a meaningful signal only upon task completion (e.g., a single positive reward for success), which is simple to define but often leads to sample-inefficient learning. In contrast, a \emph{dense reward} offers continuous feedback at each step, guiding the agent more effectively but typically requiring significant manual engineering.} 

\textcolor{black}{For the former one, language can be used to generate a \textit{sparse} reward signal to indicate whether the agent's current state successfully fulfills the language instruction, providing a flexible way to specify complex goals or human preferences. For example, ZSRM~\citep{pmlr-v162-mahmoudieh22a} derives the entire reinforcement learning reward from a natural language goal description. At each step, it generates a reward by calculating the similarity between a camera image and the goal text using CLIP encoders. This text-vision matching method \textcolor{black}{enables reward specification for some newly described goals without retraining the reward model, although transfer degrades on tasks that require stronger spatial reasoning from the CLIP model.} An alternative approach is to design a reward function by interpreting human preferences expressed through comparative language. \citet{pmlr-v270-yang25e} designs a reward learning scheme with comparative language feedback, e.g., ``\textit{move farther from the stove}''. The system aligns trajectory data with language feedback in a shared latent space, turning each piece of comparative language into a learning signal to train an episodic reward model that captures the user's underlying preferences. However, the model's ability to generalize is limited by the objects and concepts seen during its pre-training phase. Despite the convenience of language-based sparse reward design, such methods may still face challenges with low sample efficiency, requiring extended training periods or even failing to converge~\citep{DBLP:journals/inffus/LadoszWKO22}.}

\textcolor{black}{To address the sample inefficiency of sparse rewards, another line of work uses language to create \textit{dense}, \textit{intermediate} reward signals that provide more continuous guidance. These approaches typically learn a model that scores how well an ongoing trajectory aligns with a language command, using this score to shape the reward at every timestep. An example is PixL2R~\citep{goyal2021pixl2r}, which maps natural language and pixels directly to a continuous reward signal. It first learns a relatedness model from paired trajectory and language data. During policy training, this model evaluates the agent's trajectory and turns the language command into a potential function~\citep{ng1999policy}. This potential generates a dense shaping reward that is added to the environment's original reward at every step, significantly improving policy learning efficiency. This suggests a powerful paradigm where language can be used to refine and improve upon existing hand-designed reward functions. However, this method relies on classification and regression training objectives with absolute scores as ground truth, which can be rigid and may not fully capture the task progress. Furthermore, a limited language dataset may not showcase the diversity of language descriptions needed to guide robot learning effectively.}

\hypertarget{reward-learning}{}

\subsubsection{\textcolor{black}{Language-based reward function learning}}

\textcolor{black}{The above reward designing approaches highlight the importance of designing effective reward signals with language instructions, while requiring extensive manual effort and are often challenging in real-world scenarios, limiting real-world manipulation performance~\citep{deshpande2025advances}. Reward function learning aims to learn a reward function from data (e.g., demonstrations or human video~\citep{das2021model,pmlr-v164-zakka22a}), rather than manually designing it. This flexible and scalable manner enables the reward models can adapt to multi-task~\citep{dimitrakakis2011bayesian} or cross-embodiments~\citep{pmlr-v164-zakka22a,pmlr-v205-kumar23a} scenarios without requiring expert knowledge.} \textcolor{black}{In language-conditioned manipulation, reward learning methods typically learn a function that maps observations and a language instruction to a scalar reward value, which can then be used to guide a standard RL agent}. Some studies directly pre-train a reward mapping model using offline expert demonstrations. For example, LOReL~\citep{nair2022learning} implements a binary classifier $R(s_0, s, l)$ that receives the initial state $s_0$, current state $s$, and language instruction $l$, predicting whether a trajectory segment satisfies a given language instruction. This model translates visual observations and language instructions into reward signals. Besides direct learning, a classic alternative for learning the reward function from demonstrations is IRL~\citep{arora2021survey}. 

IRL is utilized to deduce the underlying reward structure, alleviating the challenge of manual reward design. It infers the rewards that could explain the observed actions of an expert. Rather than simply learning from experts, IRL has the potential to achieve better performance than the expert by optimizing the inferred reward function~\citep{finn2016guided}. \textcolor{black}{Integrating language instructions into IRL learn reward functions that are not only consistent with the expert's actions but also grounded in the specific textual command, showcasing promising performance in various tasks, such as navigation~\citep{zhou2021inverse} and mobile manipulation tasks~\citep{fu2019language}.} In mobile pick-and-place manipulation tasks, \citet{MacGlashan2015GroundingEC} present a system that grounds natural language commands to reward functions through IRL, using demonstrations of different natural language commands being carried out in the environment. \citet{fu2019language} propose language-conditioned reward learning based on MaxEnt IRL, which grounds language commands as a reward function represented by a deep neural network, incorporating generic, differentiable function approximators that can handle arbitrary observations (such as raw images). Learning the reward functions, combining language instructions and IRL, empowers the robot to solve novel tasks and transfer to realistic environments~\citep{fu2019language}, or the learned reward functions can be transferred to new robots~\citep{MacGlashan2015GroundingEC}. \textcolor{black}{However, these methods struggle with complex manipulation tasks that involve fine-grained behaviors or interactions of multiple objects~\citep{das2021model, zhang:rewind}. A primary reason is that the ambiguity of the expert's intent grows exponentially with the complexity of the task~\citep{adams2022survey}.}

\textcolor{black}{In the simple pick-and-place tasks, most demonstrations follow a simple and optimal path, making it easy for traditional IRL agents to infer the reward with the language instruction like ``\textit{move X object to Y position}''~\citep{fu2019language}. However, for complex manipulation tasks like ``\textit{sort out the kitchen table and wiping a stain}'', expert demonstrations can vary significantly in terms of manipulating trajectory for different types of objects. This variability makes it difficult for traditional IRL methods to learn the powerful reward functions that can distinguish the true underlying goal from spurious correlations in the limited demonstration data. Consequently, these methods often require a large number of demonstrations to learn a robust reward function~\citep{hwang2025masked} and struggle to generalize beyond the specific conditions they were trained on. This highlights a bottleneck, that is, the need for a reward model that possesses a general, common-sense understanding of the world, rather than one learned from scratch for each new task~\citep{glazer2024multi}.}

\subsubsection{\textcolor{black}{Foundation model-driven reward designing and learning}}
\hypertarget{FM-driven-reward-learning}{}
\textcolor{black}{The limitations of aforementioned traditional reward models, particularly their reliance on task-specific datasets and rigid training objectives, have motivated a shift towards more flexible and generalizable solutions. \textcolor{black}{Foundation models (such as LLMs and VLMs), with their broad pre-trained knowledge and powerful open-vocabulary inference abilities}~\citep{hurst2024gpt,yang2025qwen3}, provide a powerful alternative. Instead of learning a reward function from scratch on a limited dataset, these models can leverage their inherent understanding of language and vision in foundation models to directly infer task progress, thereby automating the reward design process and overcoming the need for extensive manual engineering or data collection.}

\noindent\textbf{\small LLM-driven reward code generation}\\[4pt]
\textcolor{black}{Early works leveraging FMs for reward generation explored several distinct strategies. One initial approach repurposed LLMs into reward functions. For example, \citet{kwon2023reward} reframe GPT-3~\citep{brown2020language} itself as a proxy reward function, prompting it with a trajectory transcript and asking whether the behavior satisfied a textual objective. This \textcolor{black}{prompting-based `language reward' can align agents with minimal or no task-specific demonstrations, but the setting remains confined to symbolic domains and incurs high query costs}. To overcome the latency and domain-shift limits of such on-the-fly calls of LLMs, LAMP~\citep{adeniji2023language} proposes a two-stage strategy: first, use a frozen vision-language model combination (R3M~\citep{nair2022r3m} + DistilBERT~\citep{sanh2019distilbert}) to score image-language alignment and treat that score as a dense learning signal during pre-training, then fine-tune with conventional task rewards. The approach warm-starts manipulation policies, yet its intrinsic rewards are noisy, and VLM inference during pre-training remains expensive.}

\textcolor{black}{Language2Reward~\citep{yu2023language} pushes further by asking an LLM to write executable reward code from natural-language instructions, enabling real-time user corrections and covering both quadruped locomotion and dexterous manipulation tasks. However, the method relies on the correctness of the generated code and presumes an accurate simulator. Recognizing that even task design and scene layout constrain scalability, RoboGen~\citep{wang2023robogen} wraps foundation models in a propose-generate-learn loop: the system autonomously proposes tasks, builds 3-D scenes, selects between RL, motion planning, or trajectory optimisation, and writes its own reward functions, producing an endless stream of diverse skills. Its current limitation is that everything still happens in simulation, with no guarantee of real-world transfer. Parallel work tackles reward generalization itself: Vision-Language Success Detectors~\citep{du2023visionlanguage} fine-tune Flamingo on human-labelled ``success/failure'' clips and cast success detection as a VQA problem, achieving stronger out-of-distribution robustness than traditional detectors. However, they still require labelled videos, and their binary output leaves temporal credit assignment open.}

\noindent\textbf{\small Reward code generation with self-improving}\\[4pt]
\textcolor{black}{Building on the above insights, some research focused on creating fully autonomous and sensor-grounded reward systems. The initial problem was that ``\emph{LLM-generated code was often fragile and required human oversight}''~\citep{kwon2023reward,guo2024utilizing}. Driven by the need for self-improving signals, Text2Reward~\citep{xie2024textreward} lets GPT-4 and Codex~\citep{chen2021evaluating} write dense reward functions directly from a natural-language goal with a Pythonic environment sketch, even refining the code through self-execution and optional human feedback. Policies trained on the generated code match or beat expert rewards on simulating manipulation tasks. To improve the robustness of LLM-driven reward code generation, EUREKA~\citep{ma2024eureka} generates a high-performance reward function in an evolutionary manner. It uses RL policy performance as a fitness score to guide the LLM in iteratively improving a population of reward programs. Such a loop is iterated until it meets the termination condition.} 

\textcolor{black}{A key limitation remained: the generated code used static numeric weights for different reward components (like ``Reward=0.8$\times$grasp\_success-0.2$\times$dist\_to\_cube''), which may be suboptimal. If the first draft of the generated reward code selects fragile features, the RL agent can not learn effectively. This motivated a new focus on learning these parameters automatically. Methods like Reward-Self-Align~\citep{pmlr-v235-zeng24d} and R*~\citep{li2025r} move beyond just code generation to parameter optimization. Reward-Self-Align~\citep{pmlr-v235-zeng24d} first lets an LLM write feature templates, then iteratively aligns their parameters with the LLM's pair-wise ranking of real roll-outs. R*~\citep{li2025r} learns dense reward weights from LLM preferences, but it disentangles the two challenges altogether: evolving reward structure while a critic ensemble aligned the reward function weights. The learned rewards outperform both human and EUREKA~\citep{ma2024eureka}.}

\textcolor{black}{In addition to developing methods for self-aligning critics and searching for both reward structure and weights, some methods turn to the twin bottlenecks of sample efficiency and latent-preference mis-specification. RLingua~\citep{chen2024rlingua} provides a solution by addressing the millions of interactions required by RL agents. The core idea is to prompt an LLM to generate an imperfect, rule-based controller that seeds the replay buffer and guides policy exploration, thereby reducing sample complexity. However, because this controller is bound to hand-coded state variables, it underperforms on more complex tasks. To mitigate this limitation, ELEMENTAL~\citep{chen2025elemental} fuses VLM reasoning with IRL. In this interactive pipeline, a VLM drafts executable reward features from a text prompt and a key-frame demonstration, while MaxEnt-IRL learns the weights that best match the demonstration, iteratively refining the features through a self-reflection loop. This improves generalization and success rates by narrowing the gap between a generated reward and human preferences.}

\noindent\textbf{\small VLM-driven reward learning}\\[4pt]
\textcolor{black}{However, a drawback of these LLM-driven code-centric methods is their dependence on privileged simulator state information. To operate in the real world using visual input, the focus shifted to VLMs. Video-Language Critic~\citep{alakuijala2025videolanguage} trains a temporal contrastive VLM with ranking loss on large ``Open X-Embodiment'' videos~\citep{open_x_embodiment_rt_x_2023}. The critic assigns a dense and monotonically increasing reward based on pixel-instruction pairs only and transfers across robots, thereby doubling sample efficiency over sparse rewards on unseen manipulation tasks. While this method needs thousands of good trajectories to learn a general reward. RealBEF~\citep{wang2025guiding} fine-tunes the smaller VLM ALBEF~\citep{li2021align} backbone with pair-wise image comparisons, so only a few task videos are needed. Rewards guide RL on Meta-World benchmarks~\citep{yu2020meta} better than previous image-based methods and alleviate data hunger. To further improve data efficiency, ReWiND~\citep{zhang:rewind} learns a progress-predicting reward from a handful of language-annotated demonstrations combined with ``video-rewind'' generated failures, then fine-tunes policies on unseen tasks with that reward. Its success rate beats baselines 2$\times$ in the Meta-World simulation and 5$\times$ on a real bi-manual setup. Systems like ARCHIE~\citep{TurcatoARCHIE} demonstrate a path toward real-world deployment by using an LLM to generate both the reward code and a success classifier, enabling autonomous training in simulation and successful transfer to a physical robot. However, this method relies on an assumption that the generated success detector is correct.}

\textcolor{black}{In summary, the integration of foundation models into reward designing and learning represents an advancement in language-conditioned robot manipulation. Early FM-driven methods (e.g., LAMP~\citep{adeniji2023language}, Text2Reward~\citep{xie2024textreward}, and EUREKA~\citep{ma2024eureka}) showed that LLMs can replace manual reward engineering, but exposed the weight-tuning fragility of the generated reward code. Reflection and preference-alignment methods (like Reward-Self-Align~\citep{pmlr-v235-zeng24d}, R*~\citep{li2025r}) improve robustness by learning parameters, whereas evolutionary search overcame local optima. Methods like Rlingua~\citep{chen2024rlingua} and ELEMENTAL~\citep{chen2025elemental} enhance sample efficiency and preference alignment. But these code-centric methods still required privileged simulator state information. The VLM-based methods (Video-Language Critic~\citep{alakuijala2025videolanguage}, RealBEF~\citep{wang2025guiding}) bypass this limitation by learning dense rewards directly from pixels and language, but they introduce data-scale issues. Hybrid approaches, such as ReWiND~\citep{zhang:rewind} and ARCHIE~\citep{TurcatoARCHIE}, decrease demonstration cost and mitigate the sim-to-real gap, thereby showcasing a fully autonomous reward design.}

\hypertarget{cost-function-mapping}{}
\subsection{\textcolor{black}{Language-conditioned cost functions}}
\textcolor{black}{While reward functions guide learning-based agents, cost functions are essential for optimization-based motion planners. Translating language into a cost function enables a robot to understand not only a goal, but also the constraints and preferences that indicate how to achieve it, turning natural language into a formal optimization objective.}

\subsubsection{Specific linguistic-text cost mapping}
\textcolor{black}{Early motion-planning pipelines couldn't directly interpret a human's verbal goals, requiring hand-coded cost terms or exact goal configurations~\citep{yamashita2003motion,cambon2009hybrid}. This rigidity made them unsuitable for open-ended environments like households. The initial problem was ``\emph{how to extract a task-relevant cost map from text}''. Some research focuses on creating a tighter loop between specific linguistic text and the planner's cost function. \citet{park2019efficient} introduce dynamic constraint mapping, where a Conditional Random Field~\citep{sutton2012introduction} grounds each \textit{noun phrase} or \textit{adverb} (``upright'' and ``slowly'') into the continuous parameters of a trajectory optimizer. This allows the same natural-language template to generate different cost functions on the fly. However, the limitation is that it relies on attribute-level annotations and requires retraining to handle new linguistic structures. ~\citet{sharma2022correcting} shift the supervision burden to the user. Their planner executes a trajectory and then incorporates spoken \textit{corrections} like ``\textit{stay away from the yellow bottle}'' by learning a residual cost function that modifies the optimizer's objective. They sidestep the need for a fully annotated dataset but are limited to generating 2D cost maps, which struggle in 3D scenes.}

\subsubsection{FM-driven cost mapping}
\textcolor{black}{To tackle 3D complexity, VoxPoser~\citep{huang2023voxposer} harnesses foundation models. It uses GPT-4 to write Python code that queries a VLM (OWL-ViT~\citep{minderer2022simple}) to compose dense 3D value maps. These maps assign high values (rewards) to affordances and low values (costs) near constraints (like ``\textit{watch out for the vase}''), \textcolor{black}{enabling a standard motion planner to synthesize trajectories from novel language instructions without planner retraining.} The primary limitation is its heavy reliance on the perception module, which integrates OWL-ViT~\citep{minderer2022simple}, Segment Anything~\citep{kirillov2023segment}, and XMEM~\citep{cheng2022xmem}. If the object detector fails, the resulting cost map is incorrect. To capture complex spatio-temporal relationships, ReKep~\citep{huang2024rekep} alternatively uses a VLM to write Python code defining symbolic constraints instead of creating a value map~\citep{huang2023voxposer}. ReKep first uses a vision model (DINOv2~\citep{Oquab2023DINOv2LR}) to automatically propose semantically meaningful 3D keypoints on objects. An image with these keypoints and user language is then fed to a VLM, which generates a sequence of cost functions defining arithmetic relationships between the keypoints (like \textit{minimizing the distance between a teapot spout and a cup}). These code-based constraints are passed to a real-time optimization solver that plans the robot's trajectory. A key advantage is its ability to implicitly specify complex 6-DoF motions by having the VLM reason only about simple 3D point relationships, offloading the difficult geometric calculations to the numerical solver.} 

\textcolor{black}{Instead of composing value maps, LACO~\citep{xie2023language}  learns a collision function directly. It uses a transformer to fuse a single RGB image, the robot's joint state, and a language prompt (such as ``\textit{can collide with toy}'') to predict a language-conditioned collision score. This score is then used as a cost by the planner. By avoiding the need for depth data or 3D meshes, it generalizes well. However, its reliance on a single camera view means it ignores risks posed by occluded objects. IMPACT~\citep{ling2025impact} leverages the advanced semantic and spatial reasoning capabilities of VLMs (GPT-4o~\citep{hurst2024gpt}) to automate the inference of ``acceptable contact''. It feeds multi-view RGBD images to GPT-4o, rating each object's ``contact tolerance'' on a 0-10 scale and is then used to create a 3D cost map that integrates directly with standard planners ( e.g., RRT*~\citep{karaman2011sampling}). The robot can make incidental contact with a soft object while strictly avoiding a fragile one, without explicit instructions to do so. IMPACT unifies high-level semantic reasoning with continuous optimization, but is limited by its assumption of a static scene and perfect object segmentation, opening up a challenge: online cost refinement as the environment changes.}

\subsection{Summary}
\textcolor{black}{In its role as a tool for {state evaluation}, language provides a flexible and intuitive interface for specifying a robot's objectives, translating high-level instructions into quantitative scoring functions that guide robot behavior. This paradigm addresses our first core question by grounding language into reward functions for reinforcement learning or cost functions for motion planning.}

\textcolor{black}{Initial methods focused on manually designing language-based rewards, which could be sparse~\citep{pmlr-v162-mahmoudieh22a} or dense~\citep{goyal2021pixl2r}. While more intuitive than traditional reward engineering, these approaches often struggled with sample inefficiency or required significant expert effort. Reward function learning, particularly through IRL, offered a data-driven alternative by inferring rewards from expert demonstrations. However, these methods face challenges in complex multi-object manipulation tasks where expert intent is ambiguous, often requiring an impractically large number of demonstrations to generalize effectively. The development of FMs accelerates this shift, automating reward design by leveraging their vast pre-trained knowledge. Early approaches used LLMs to generate reward code~\citep{yu2023language} or act as proxy reward functions~\citep{kwon2023reward}. More advanced methods have created fully autonomous systems that can write, refine, and optimize reward functions through self-reflection, evolutionary algorithms~\citep{ma2024eureka}, or alignment with human preferences ~\citep{pmlr-v235-zeng24d,li2025r}. VLM-based methods further bridge the sim-to-real gap by learning dense rewards directly from pixels and language~\citep{alakuijala2025videolanguage,zhang:rewind}. Similarly, in motion planning, FMs are used to generate dense 3D cost maps from language, \textcolor{black}{enabling planners to handle previously unseen language constraints and affordances without task-specific retraining of the planner, although success remains conditional on perception quality and scene coverage}~\citep{huang2023voxposer,huang2024rekep,ling2025impact}. The overarching insight is that the role of language in state evaluation has evolved from a tool for manual reward/cost specification to a medium for automated, knowledge-driven reward and cost generation, powered by FMs' reasoning capabilities. This progression has enhanced the scalability, flexibility, and data efficiency of teaching robots complex manipulation tasks.}

\section{Language as a policy condition}
\hypertarget{language-as-a-policy-condition}{}
\label{sec:language-as-policy-condition}

\begin{figure}[ht!]
    \centering
    \resizebox{.49\textwidth}{!}{
    \begin{tikzpicture}
    \selectcolormodel{cmyk}
    \hypersetup{linkcolor=black}    
        \node[circle, draw=RoyalBlue!50, fill=RoyalBlue!50, text width=3cm, align=center, minimum size=3cm]
            (title) at (0,0) {\Large\textbf{\textsf{5. Language as Policy Condition}}};
    
        \node[circle, draw=Salmon!50, fill=Salmon!50, text width=2.5cm, align=center, minimum size=2.5cm, left]
            (title-51) at ([xshift=-1.5cm]title.west) {\large\textbf{\textsf{5.1 Language in RL}}};
        \node[circle, draw=YellowOrange!50, fill=YellowOrange!50, text width=2.5cm, align=center, minimum size=2.5cm, above]
            (title-52) at ([yshift=1.5cm]title.north) {\large\textbf{\textsf{5.2 Language in BC}}};
        \node[circle, draw=JungleGreen!50, fill=JungleGreen!50, text width=3cm, align=center, minimum size=2.5cm, right]
            (title-53) at ([xshift=1.5cm]title.east) {\large\textbf{\textsf{5.3 Language in Diffuison-based Policy Learning}}};

        \node[circle, draw=Salmon!30, fill=Salmon!30, text width=2.5cm, align=center, minimum size=2.5cm, left]
            (title-51-1) at ([yshift=2cm]title-51.north west) {\large\textbf{\textsf{Modular Skills}}};
        \node[circle, draw=Salmon!30, fill=Salmon!30, text width=2.5cm, align=center, minimum size=2.5cm, left]
            (title-51-2) at ([xshift=-1.5cm]title-51.west) {\large\textbf{\textsf{Multi-task Learning}}};
            \node[circle, draw=Salmon!30, fill=Salmon!30, text width=2.5cm, align=center, minimum size=2.5cm, left]
            (title-51-3) at ([yshift=-2cm]title-51.south west) {\Large\textbf{\textsf{Lifelong Learning}}};

        \node[circle, draw=YellowOrange!30, fill=YellowOrange!30, text width=2.5cm, align=center, minimum size=2.5cm, left]
            (title-52-1) at ([xshift=-2cm, yshift=1cm]title-52.north west) {\large\textbf{\textsf{Data-efficient Learning}}};
        \node[circle, draw=YellowOrange!30, fill=YellowOrange!30, text width=2.5cm, align=center, minimum size=2.5cm, above]
            (title-52-2) at ([yshift=1cm]title-52.north) {\large\textbf{\textsf{Transfromer-based Learning}}};
        \node[circle, draw=YellowOrange!30, fill=YellowOrange!30, text width=2.5cm, align=center, minimum size=2.5cm, right]
            (title-52-3) at ([xshift=2cm, yshift=1cm]title-52.north east) {\large\textbf{\textsf{Long-horizon Task Learning}}};
            
        \node[circle, draw=JungleGreen!30, fill=JungleGreen!30, text width=2.5cm, align=center, minimum size=2.5cm, right]
            (title-53-1) at ([yshift=2cm]title-53.north east) {\large\textbf{\textsf{Skill Discover}}};
        \node[circle, draw=JungleGreen!30, fill=JungleGreen!30, text width=2.5cm, align=center, minimum size=2.5cm, right]
            (title-53-2) at ([xshift=1.5cm]title-53.east) {\large\textbf{\textsf{Real-world Tasks}}};
        \node[circle, draw=JungleGreen!30, fill=JungleGreen!30, text width=2.5cm, align=center, minimum size=2.5cm, right]
            (title-53-3) at ([yshift=-2cm]title-53.south east) {\large\textbf{\textsf{Cross-embodiment Transfer}}};

        \path (title) to[circle connection bar switch color=from (RoyalBlue!50) to (Salmon!50)] (title-51);
        \path (title) to[circle connection bar switch color=from (RoyalBlue!50) to (YellowOrange!50)] (title-52);
        \path (title) to[circle connection bar switch color=from (RoyalBlue!50) to (JungleGreen!50)] (title-53);
        \path (title-51) to[circle connection bar switch color=from (Salmon!50) to (Salmon!30)] (title-51-1);
        \path (title-51) to[circle connection bar switch color=from (Salmon!50) to (Salmon!30)] (title-51-2);
        \path (title-51) to[circle connection bar switch color=from (Salmon!50) to (Salmon!30)] (title-51-3);
        \path (title-52) to[circle connection bar switch color=from (YellowOrange!50) to (YellowOrange!30)] (title-52-1);
        \path (title-52) to[circle connection bar switch color=from (YellowOrange!50) to (YellowOrange!30)] (title-52-2);
        \path (title-52) to[circle connection bar switch color=from (YellowOrange!50) to (YellowOrange!30)] (title-52-3);
        \path (title-53) to[circle connection bar switch color=from (JungleGreen!50) to (JungleGreen!30)] (title-53-1);
        \path (title-53) to[circle connection bar switch color=from (JungleGreen!50) to (JungleGreen!30)] (title-53-2);
        \path (title-53) to[circle connection bar switch color=from (JungleGreen!50) to (JungleGreen!30)] (title-53-3);
        \draw [-{Triangle Cap []. Fast Triangle[]}, draw=Salmon!50, line width=10pt] (title-51) to[out=90, in=-170] node[midway, above, sloped, yshift=2pt]{\large\textbf{\textsf{\textcolor{YellowOrange}{addresses limitations of}}}} (title-52);
        \draw [-{Triangle Cap []. Fast Triangle[]}, draw=YellowOrange!50, line width=10pt] (title-52) to[out=-10, in=100] node[midway, above, sloped, yshift=2pt]{\large\textbf{\textsf{\textcolor{JungleGreen}{addresses limitations of}}}} (title-53);

        \node[text width=3.1cm, align=left]at(-2.4cm, 2.5cm)
        {\textcolor{BrickRed}{Cons of RL:\\ \textsf{Reward engineering,}\\ \textsf{Inefficient learning,}\\ \textsf{limited grounding}}};
        \node[text width=3.1cm, align=left]at(2.9cm, 3cm)
        {\textcolor{BrickRed}{Cons of BC:\\ \textsf{Learn suboptimal behavior}}};
        \node[text width=3.1cm, align=left]at([yshift=-.5cm]title-53.south)
        {\textcolor{BrickRed}{Cons of DP:\\ \textsf{Inference latency}}};
        
        \node[align=center, below](idea)at([yshift=-2cm]title.south){\Large\textbf{\textsf{Core idea:}} \textsf{language instructs ``how to do''.}};
        \node[align=center, below](role)at([yshift=-.1cm]idea.south){\Large\textbf{\textsf{Role:}} \textsf{Language as behavior-specifier.}};
    
    \end{tikzpicture}
    }
    \caption{\textcolor{black}{Taxonomy of Sec. \ref{sec:language-as-policy-condition} Language as policy condition.} \textcolor{black}{The arrow (``addresses limitations of'') head is commonly used to introduce a family that alleviates key practical limitations of the preceding family (not a strict hierarchy or a complete replacement). For instance, language-conditioned behavioral cloning can reduce reliance on reward engineering and exploration in RL by learning directly from demonstrations, while diffusion-based policies can further alleviate BC's tendency to imitate suboptimal or averaged behaviors, albeit sometimes at the cost of higher inference latency.}}
    \label{fig:taxonomy-sec5}
\end{figure}

\textcolor{black}{While the previous section focused on using language to quantify the task progress for implicitly directing the robot's behaviors. This section shifts to an alternative paradigm for clarifying the role of the language in robotic manipulation: using language as an explicit condition for policy learning to specify \textit{how} a robot should act. Instead of translating language into a reward or cost function that indirectly guides a learning or planning algorithm, the methods discussed here integrate language into the policy itself. This approach addresses our second key research question: How can language condition a policy to produce the correct behavior? Here, the policy $\pi_\theta$, parameterized by the neural network $\theta$, learns a direct mapping from the current observation $s_t$ and the language instruction $l$ to an action $a_t$, i.e., $\pi_\theta(a_t | s_t, l)$. This changes the role of language from a goal specifier to a behavior specifier.} We will explore how this concept is realized through different families of algorithms, including reinforcement learning, imitation learning, and emerging diffusion-based policy learning. Figure \ref{fig:taxonomy-sec5} presents the taxonomy of this section. 

\textcolor{black}{Table \ref{tab:language_as_policy_condition} presents a comprehensive comparison of the state-of-the-art methods that shift the role of language from a goal specifier to a behavior specifier, mapping observations and instructions directly to actions. The table categorizes these approaches by their underlying algorithmic paradigms: reinforcement learning, behavioral cloning, and diffusion-based policy learning. By analyzing these methods side-by-side, we can observe the progression of techniques designed to address distinct learning bottlenecks, such as addressing sample inefficiency and exploration challenges in RL, or overcoming the averaging of multi-modal behaviors in standard BC through generative diffusion models.}
\begin{table*}[htbp]
    \centering
    \caption{\textcolor{black}{Comparison of representative state-of-the-art methods of \textbf{Section \ref{sec:language-as-policy-condition} Language as a policy condition}. These selected approaches illustrate the progression across reinforcement learning, behavioral cloning, and diffusion-based paradigms , demonstrating how language acts as a behavior specifier to solve distinct learning bottlenecks like sample inefficiency and multi-modal behavior averaging.}
    \label{tab:language_as_policy_condition}}
    \small
    \begin{tabular}{p{0.12\linewidth}| p{0.1\linewidth}| p{0.21\linewidth}| p{0.21\linewidth}| p{0.21\linewidth}}
        \toprule[1pt]
        \textbf{Method} & \textbf{Learning Paradigm} & \textbf{Addressed Bottleneck} & \textbf{Key Advantages} & \textbf{Key Disadvantages} \\
        \midrule[0.2pt]
        MILLION\quad\citep{bing2023meta} & Reinforcement Learning & Sample inefficiency during exploration. & Achieves rapid adaptation on unseen tasks via memory-based meta-learning. & Requires decoupling the reading phase from the acting phase. \\
        \midrule[0.2pt]
        FLaRe\qquad\citep{hu2025flare} & Reinforcement Learning & Inefficient learning of multi-task behaviors from scratch. & Achieves 15x faster learning than dense-reward baselines via fine-tuning. & Inherits the limitations of the pre-trained BC model's training data distribution. \\
        \midrule[0.2pt]
        PerAct\qquad\citep{shridhar2023perceiver} & Behavioral Cloning & Learning 3D spatial relationships from 2D projections. & Provides a strong 3D structural prior by voxelizing both observations and action spaces. & Introduces high computational overhead due to dense, high-resolution voxelization requirements. \\
        \midrule[0.2pt]
        HULC\qquad\citep{mees2022matters} & Behavioral Cloning & Long-horizon task learning and generalization. & Creates robust language-conditioned representations using a transformer-based architecture. & Standard BC methods inherently suffer from compounding execution errors over long horizons. \\
        \midrule[0.2pt]
        StructDiffusion\quad\citep{liu2022structdiffusion} & Diffusion-based Policy & Grounding language in physically plausible scene arrangements. & Samples diverse, collision-free goal poses for multiple unseen objects. & Performance is heavily limited by the capabilities of subsequent low-level policies.\\
        \midrule[0.2pt]
        ChainedDiffuser\quad\citep{xian2023chaineddiffuser} & Diffusion-based Policy & Long-horizon continuous trajectory generation. & Unifies discrete keypose prediction with local trajectory diffusion for smooth motion. & Inherits the high inference latency typical of iterative denoising processes.\\
        \bottomrule[1pt]
    \end{tabular}
\end{table*}

\hypertarget{reinforcement-learning}{}
\subsection{Language in reinforcement learning}

RL algorithms can be applied to tasks based on language-conditioned rewards. Early attempts at language-conditioned RL concentrated on games~\citep{fu2019language, MacGlashan2015GroundingEC, bahdanau2018learning, kaplan2017beating, goyal2019using}, since games often have well-defined rules and objectives, and also easy to reproduce experiments and compare the performance. These studies train an agent capable of comprehending natural language instructions given by humans. In these games, human languages are given to control the agent to solve navigation tasks~\citep{chaplot2018gated, misra2017mapping, andreas2017modular, MacGlashan2015GroundingEC, janner2018representation}, scoring games~\citep{kaplan2017beating, goyal2019using}, and object manipulations~\citep{bahdanau2018learning}. However, this does not explain how language can guide robots to perform complex manipulation tasks with high-dimensional actions. 

\textcolor{black}{An early solution was to decouple language from control. LCGG~\citep{colas2020language} uses language to condition a goal generator, which then provides a language-agnostic goal to a pre-trained goal-conditioned policy. This modularity allows any off-the-shelf RL controller to be used, enabling a single instruction to generate a diverse range of behaviors. However, because language never directly touches the policy network, the robot cannot adapt its low-level behavior to linguistic intent and is limited to simple pick-and-place tasks involving different colors. To address this limitation, subsequent research aims to integrate language more tightly into the policy loop. For example, LanCon-Learn~\citep{silva2021lancon} achieves this by directly feeding a language embedding into an attention router that gates multiple skill modules inside a shared actor-critic network. By allowing language to re-weight reusable skills at every timestep, \textcolor{black}{a single network could master dozens of tasks and recombine them for limited task recomposition on held-out instructions or task combinations.} The drawback is that learning the router and control policies simultaneously proved to be fragile and sample-hungry. To improve sample efficiency,  MILLION~\citep{bing2023meta} introduces a memory-based meta-learning approach using a Gated Transformer-XL~\citep{parisotto2020stabilizing}. It separates the process into a brief instruction phase, where the model reads the language command into its memory, and a trial phase, where it acts based on that stored context. This decoupling of ``reading'' from ``acting'' accelerates exploration and enables rapid adaptation to unseen manipulation tasks. Building on this, \citet{yao2023learning} observes that many manipulation tasks come in symmetric pairs (``open left drawer'' vs. ``open right drawer''). It automatically generates symmetric language instructions using antonym rules and co-trains the MILLION backbone on both the original and symmetric tasks, leading to faster convergence and improved performance. Yet hand-crafted symmetry rules are not always applicable. }

\textcolor{black}{Free-form language instructions exhibit expression diversity, where the same task can be described with varied expressions. Thus, directly learning from language instructions can be sample-inefficient. Some works explore whether a more structured language representation could improve efficiency. For example, TALAR~\citep{pang2023natural} proposes translating free-form text into a compact and discrete set of Task-Language (TL) predicates using a VAE-based translator. The RL policy is then trained only on this simplified TL, making the learning process more data-efficient and the resulting task codes more interpretable. Similarly, LOVM~\citep{ye2024task} grounds language at the pixel level by using FiLM layers to modulate image features based on the instruction that is encoded by BiGRU or DistilBERT, which in turn predicts object masks for training a downstream DQN. Both approaches demonstrate high success rates in pick-place tasks but come with their own trade-offs: TALAR~\citep{pang2023natural} requires a curated dataset to pre-train its translator, while LOVM~\citep{ye2024task} needs pixel-level mask supervision. InstructRobot~\citep{cleveston2025instructrobot} shows that a small transformer can directly map raw language combined with RGB-D observations to a 26-DoF robot's joint commands without requiring curated language-action pairs or handcrafted task predicates. The approach is modular and lightweight, but it handles only short-horizon simple tabletop tasks and relies on customized sparse rewards. }

\textcolor{black}{Learning multiple complex manipulation tasks with a single RL policy remains an open challenge, which may bring the issue of catastrophic forgetting. LEGION~\citep{meng2025preserving} clusters language embeddings online and uses the resulting cluster ID to gate reusable skills, demonstrating that language helps disambiguate tasks during lifelong learning and enables recombining mastered subtasks to solve unseen long-horizon goals. Furthermore, FLaRe~\citep{hu2025flare} mitigates this limitation by fine-tuning a large pre-trained behavioral cloning policy (vision-language transformer) with PPO under sparse linguistic rewards, achieving 15 times faster learning than dense-reward baselines and supporting rapid embodiment transfer. However, the reliance on a pre-trained BC model means it inherits the limitations of its training data and may struggle with tasks outside that distribution. To avoid fine-tuning or even accessing the weights of the pre-trained policy, V-GPS~\citep{nakamoto2024steering} learns a language-conditioned value function via offline RL, and the learned value function can re-rank the actions of the pre-trained policy, steering the agent's behavior at deployment time. Recent work pushes further by injecting additional structure into the multi-task learner: LIMT~\citep{aljalbout2025limt} combines language with world-model imagination. A Sentence-BERT vector conditions both a VQ-VAE tokenizer and a transformer dynamics model inside a Dreamer pipeline, and the latent actor-critic agent (model-based RL) then plans in imagination before acting. Therefore, when the agent predicts the future during training and execution, the language itself can provide information to it, thereby improving sampling efficiency and generalization.}

\textcolor{black}{In summary, using language as a direct policy condition in RL transforms it from a goal specifier to a behavior specifier, but this shift introduces challenges, primarily sample inefficiency and inability to generalize in multi-task settings. Early approaches that decoupled language from control were modular but could not capture fine behaviors~\citep{colas2020language}. This leads to end-to-end methods that integrate language directly into the policy loop, which is more expressive, while suffering from poor sample efficiency~\citep{silva2021lancon}. To address this, researchers develop techniques such as memory-based meta-learning (MILLION~\citep{bing2023meta}), exploiting task symmetries for data augmentation~\citep{yao2023learning}, and translating free-form text into structured predicates (TALAR~\citep{pang2023natural}). To achieve richer conditioning, recent works have moved beyond simple feature concatenation, instead using FiLM layers to modulate visual features based on the instruction (LOVM~\citep{ye2024task}), clustering language embeddings to gate reusable skills for lifelong learning (LEGION~\citep{meng2025preserving}), and even conditioning the latent dynamics of a world model to shape an agent's ``imagined'' futures (LIMT~\citep{aljalbout2025limt}). For scaling to hundreds of complex manipulation tasks, a powerful paradigm has emerged: fine-tuning large pre-trained behavioral cloning policies with sparse linguistic rewards (FLaRe~\citep{hu2025flare}), and steering the agent's behavior through a learned language-conditioned value function (V-GPS~\citep{nakamoto2024steering}). Although these advances show a clear path toward deeply integrating language into different levels of the RL decision-making process, open challenges remain. These include the reliance on curated reward functions, the high computational and data cost of large-scale pre-training, the inefficient learning of multi-task, and the limited ability to ground nuanced or stylistic language beyond direct commands.}

\hypertarget{behavioral-cloning}{}

\subsection{Language in behavioral cloning}
\label{sec:language-IL}

\textcolor{black}{Given the challenges of sample inefficiency and complex reward engineering in RL, an alternative paradigm is to learn policies directly from expert demonstrations. This approach, known as imitation learning, circumvents the difficult exploration problem by training an agent to mimic expert behavior. The most direct form of IL is BC, which frames policy learning as a supervised learning problem: given a dataset of expert trajectories, the goal is to learn a policy $\pi_\theta(a_t | s_t)$ that maps an observation $s_t$ to the expert's action $a_t$. To address our initial question of \emph{how language can specify robot behavior} in this field, a direct approach is to treat language instruction $l$ as a conditional input to the imitation learning policy, such that $\pi_\theta(a_t | s_t, l)$. A mainstream strategy for implementing this is to adapt goal-conditioned imitation learning frameworks, where language instructions serve as the goal.}

\begin{figure}
    \centering
    \begin{tikzpicture}
        \node[draw=none, inner sep=0pt](fig) at (0,0) {\includegraphics[width=0.95\linewidth]{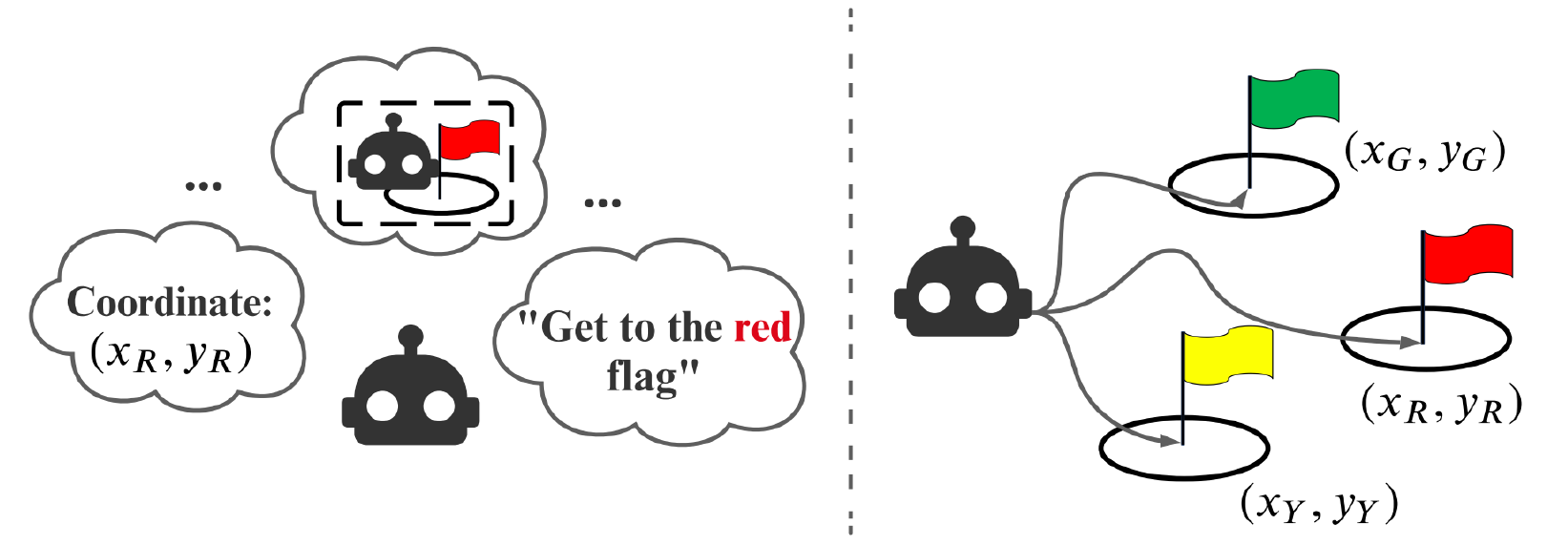}};
        \node[draw=none, anchor=north west, inner sep=1pt](atext) at ($(fig.south west)+(1, 0)$) {(a) Specify goals};
        \node[draw=none, anchor=north east, inner sep=1pt](btext) at ($(fig.south east|-atext.north)+(-0.5, 0)$) {(b) Execute actions};
    \end{tikzpicture}

  \caption{Typical representations of goals in goal-conditioned learning: vectors, images, and languages (illustration adapted from \cite{DBLP:journals/corr/abs-2201-08299}).}
  \label{fig:goals}
  \vspace{-1.5em}
\end{figure}

\noindent \textbf{\small Language as goals in goal-conditioned IL}\\[4pt] 
\textcolor{black}{Goal-conditioned IL (GCIL) provides a natural foundation for language-conditioned IL. GCIL provides various types of goal conditions for policy learning, including visual and spatial goals, or linguistic goals, as shown in Figure~\ref{fig:goals}. In language-conditioned IL, by replacing abstract goal representations (e.g., state vectors or images) with language instructions, the policy learns to map observations and text to actions. This paradigm shift empowers robots to understand and execute a wide range of commands specified in natural language. A key challenge in this domain is effectively grounding language instructions in visual observations to guide manipulation. Early approaches tackled this by using task-focused visual attention (TFA) mechanisms~\citep{abolghasemi2019pay,stepputtis2020language}, which directed the vision system to task-relevant regions of an image based on the command, thereby improving manipulation precision. CLIPORT~\citep{shridhar2022cliport} combines the broad semantic understanding of CLIP with the spatial precision of Transporter~\citep{zeng2021transporter}, solving a variety of language-specified tabletop tasks from packing unseen objects to folding cloths. However, these methods are often constrained by the limited availability of paired action-language data and architectural bottlenecks, which hinder their ability to learn complex multi-step tasks.} \textcolor{black}{To overcome the data scarcity problem, some research focus on data-efficient learning and leveraging diverse data sources, such as unstructured data (or play data). Multi-context imitation learning (MCIL)~\citep{lynch2020language} demonstrates that an agent could learn effectively even when less than 1\% of the demonstration data is language-annotated, by co-training on multiple goal modalities (task IDs, images, and language). Similarly, BC-Z~\citep{jang2022bc} improves data efficiency by learning from a mix of expert demonstrations and cheaper, imperfect interventions. Other works explore alternative data sources entirely, such as MimicPlay~\citep{wang2023mimicplay}, which enables robots to learn from videos of humans freely interacting with objects, bypassing the need for costly teleoperated demonstrations.} 

\textcolor{black}{Motivated by the architectural bottlenecks of convolutional and recurrent backbones, a second wave of work replaces task-specific backbones with fully-attentional transformers~\citep{vaswani2017attention}. These works share the intuition that: \emph{if language is inherently sequential and relational, then the action policy should process tokens of words, images and robot proprioception with the same permutation-invariant mechanism}. HiveFormer~\citep{guhur2023instruction} asks \emph{``How can a robot remember what it has already done?''}. To overcome the challenge of \emph{partial observability in multi-step tasks}, HiveFormer concatenates language tokens with all past visual-proprioceptive tokens into a single sequence and processes them with a multimodal Transformer. The resulting history-aware policy improves long-horizon success on 74 RLBench~\citep{james2020rlbench} tasks, demonstrating that transformers can scale beyond single images. While in multi-task scenarios, learning 3D spatial relationships and 6-DoF actions directly from 2D image projections is challenging. PerAct~\citep{shridhar2023perceiver} mitigates this by framing the problem as ``\textit{detecting the next best voxel action}''. It contrastes with earlier image-to-action methods~\citep{shridhar2022cliport,jang2022bc} by voxelizing both the RGB-D observations and the 6-DoF action space, providing a strong 3D structural prior that proved more data-efficient for complex language-conditioned manipulation. PerAct enables the agent to learn dozens of diverse tasks from only a few demonstrations each. While voxelization provides a powerful 3D inductive bias, it introduces computational challenges, particularly for high-resolution grids needed for precision manipulation tasks.}

\textcolor{black}{To address the computational overhead of dense voxelization, subsequent research explored two main directions. The first path aimed to improve the efficiency and semantic richness of 3D representations. Act3D~\citep{gervet2023act3d} tackles the computational cost by representing the workspace as a 3D feature field, using a coarse-to-fine attention mechanism to adaptively sample and focus on relevant 3D points rather than processing a dense grid. This allows for high-resolution action prediction with a fraction of the compute. In parallel, GNFactor~\citep{ze2023gnfactor} argues that existing 3D fields~\citep{driess2022reinforcement} lack semantics. It distilled pre-trained 2D vision-language models' features into a generalizable neural feature field, thereby grounding the 3D representation with a deeper understanding of object semantics and functionality. GNFactor transferred to language-conditioned real kitchen tasks with only 100 demonstrations, whereas geometry-only fields struggled with novel objects. }

\textcolor{black}{Concurrent to the development of 3D policies, other research questioned the necessity of explicit voxelization, seeking to inherit the scalability and efficiency of view-based methods while achieving high performance in 3D tasks. Instead of voxelizing space, RVT~\citep{goyal2023rvt} re-renders the point cloud from 8-16 virtual cameras and applies a multi-view ViT~\citep{dosovitskiy2020image}  with cross-view attention. This approach trained 36x faster than PerAct~\citep{shridhar2023perceiver} for achieving the same performance in language-conditioned RLBench tasks, showing that clever view selection can replace heavy 3-D convolutions. Building on this multi-view paradigm, $\Sigma$-agent~\citep{ma2025contrastive} revisits data efficiency by introducing Contrastive Imitation Learning, adding an auxiliary contrastive loss to the standard BC objective. It pulls together embeddings of vision-language pairs from the same task and pushs apart others in the feature space, outperforming RVT~\citep{goyal2023rvt} in multi-task settings. While these methods advanced the state-of-the-art for rigid object manipulation, they still struggled with deformable objects. To address this, ~\citet{deng2024learning} propose a model that encodes cloth as a graph and uses a transformer to reason jointly over pixels, graph nodes, and language instructions, improving performance on complex cloth manipulation tasks.}

\textcolor{black}{Pushing the boundaries of what language-conditioned IL could achieve, some work explores long-horizon manipulations. In a long-horizon task setting, the robot needs to solve complex manipulation tasks by understanding a series of unconstrained language expressions in a row, for example, ``\textit{open the drawer} $\rightarrow$ \textit{pick up the blue block} $\rightarrow$ \textit{push the block into the drawer} $\rightarrow$ \textit{open the sliding door}''~\citep{9788026}. Integrating the advantages of data-source improvements and architectural innovations. Some work empowers the language-conditioned policy to solve long-horizon manipulations. For example, HULC~\citep{mees2022matters} introduces a transformer-based architecture with contrastive learning to create more robust language-conditioned representations. This is later extended by HULC++~\citep{mees23hulc2}, which integrates visual affordance models and leverages LLMs for long-horizon planning. Moreover, to enhance generalization to novel scenarios, SPIL~\citep{zhou2023languageconditioned} builds upon the HULC framework by incorporating pre-trained skill priors, allowing the agent to better adapt to unfamiliar environments.} 

\textcolor{black}{However, an inherent issue in BC is compounding error, hindering learning language-conditioned policies to solve long-horizon manipulation tasks. To mitigate the issue of compounding error, ACT~\citep{Zhao-RSS-23} introduces \textit{action chunking}, which predicts sequences of actions at a time, and uses temporal ensembling to improve robustness. Building on this idea, MT-ACT~\citep{bharadhwaj2024roboagent} trains a language-conditioned policy using a multi-task action-chunking transformer architecture, leveraging efficient action representations for ingesting multi-modal multitask data into a single policy. Some researchers believe that for long-horizon high-precision tasks like autonomous surgery, a simple end-to-end policy is insufficient. SRT-H~\citep{kim2025srt} introduces a hierarchical framework where a high-level Transformer policy generates language instructions (e.g., corrective ones) to guide a low-level policy, enabling robust multi-step execution and linguistic error correction in realistic surgical settings.}

\textcolor{black}{In summary, using language as a direct policy condition in BC has evolved, addressing challenges like data scarcity and long-horizon task execution. Early methods focused on grounding language in visual observations using task-focused attention mechanisms~\citep{abolghasemi2019pay,stepputtis2020language} and combining pre-trained vision-language models with spatial reasoning~\citep{shridhar2022cliport}. To overcome data scarcity, approaches like MCIL~\citep{lynch2020language} and BC-Z~\citep{jang2022bc} demonstrate effective learning from limited language-annotated data and imperfect interventions, while MimicPlay~\citep{wang2023mimicplay} leverages unstructured human interaction videos. The adoption of transformer architectures~\citep{vaswani2017attention} enables models like HiveFormer~\citep{guhur2023instruction} to handle partial observability in multi-step tasks by attending over entire action histories, and PerAct~\citep{shridhar2023perceiver} to learn 6-DoF actions from voxelized 3D spaces. To address the computational challenges of dense voxelization, subsequent works pursued more efficient 3D representations (Act3D~\citep{gervet2023act3d}, GNFactor~\citep{ze2023gnfactor}) and multi-view approaches (RVT~\citep{goyal2023rvt}, $\Sigma$-agent~\citep{ma2025contrastive}). For long-horizon tasks, innovations like HULC~\citep{mees2022matters}, SPIL~\citep{zhou2023languageconditioned} introduce latent-representation and skill-prior mechanisms for maintaining long-term context. Hierarchical frameworks like SRT-H~\citep{kim2025srt} further enhance robustness by combining high-level language planning with low-level execution. These works demonstrate that deeply integrating language into BC policies can enable robots to perform complex multi-step tasks specified in natural language. However, challenges remain in scaling to more diverse tasks, improving data efficiency, and enhancing the robustness of learned policies.} 
\input{tikz/diffusion2}
\\
\hypertarget{diffusion-based-policy-learning}{}

\subsection{Language in diffusion-based policy learning}

\textcolor{black}{While traditional imitation learning methods like BC have been successful, they may struggle with tasks that exhibit multi-modal action distributions, where multiple valid action sequences can achieve the same goal. Methods that rely on direct regression, mixtures of Gaussian~\citep{mandlekar2021matters}, or discretization~\citep{shafiullah2022behavior} perform poorly on this problem, as illustrated in Figure~\ref{fig:dp}\textcolor{red}{(a)}. For example, the regression methods, such as using a Mean Squared Error (MSE) loss~\citep{torabi2018behavioral}, tend to average these diverse expert behaviors, resulting in conservative or even invalid actions~\citep{chi2023diffusion}. To overcome this limitation, generative models have been introduced into this field, which are adept at learning complex high-dimensional data distributions. Among these, diffusion models have emerged as a powerful and stable alternative to Generative Adversarial Networks~\citep{srivastava2017veegan} and energy-based models~\citep{florence2022implicit}, offering a robust framework for modeling expressive and multi-modal robot behavior~\citep{urain2024deep}.}

\textcolor{black}{The core of a diffusion-based policy is a two-stage process, as illustrated in Figure \ref{fig:dp}(b). In the forward (diffusion) phase, Gaussian noise is incrementally added to expert action data over a series of timesteps, gradually corrupting it into pure noise. The model, typically a noise-prediction network, is then trained to reverse this process. In the reverse (denoising) phase at inference time, the model starts with random noise and iteratively refines it over a sequence of steps, conditioned on the current state observation, to generate a clean, executable action sequence~\citep{chi2023diffusion,ho2020denoising}. This generative paradigm has proven highly effective and has been applied to a wide array of robotic manipulation applications, including visual-motor control~\citep{pearce2022imitating, chi2023diffusion,scheikl2024movement, octomodelteam2024octo,dipalo2024kat,dasari2024ditpi}, tasks in 3D scenes~\citep{xian2023chaineddiffuser,yan2024dnact,vosylius2024render,ze20243d,lu2024manicm}, human-robot interactions\citep{ng2023diffusion,wang2025inference}, cross-embodiment transfer~\citep{yao2025pick,yao2025inference}, contact-rich manipulation~\citep{xue2025reactive}, long-horizon planning~\citep{mishra2023generative}, and robust control~\citep{hu2026dreamingunseenworldmodelregularized}.}

\textcolor{black}{To direct the generation process toward specific goals, the denoising network can be conditioned on additional information not just observations (robot states and scene context). By incorporating language instructions as a condition, these models become language-conditioned diffusion policies. This allows a single policy to generate behaviors for a wide range of tasks specified by text. The language instruction, along with the visual observation, guides the denoising process, ensuring the generated action sequence is not only physically plausible but also semantically aligned with the command. This approach has been successfully used to learn goal-and-state-conditioned distributions, enabling policies to capture multiple valid solutions from demonstration data and solve complex tasks specified by image or language goals~\citep{xian2023chaineddiffuser,reuss2023goal,chen2023playfusion,ha2023scaling,reuss2024multimodal,hao2025language,ke20253d}, even for unseen objects~\citep{liu2022structdiffusion}. In the following, we will discuss the integration and role of language instructions in this field.}

\textcolor{black}{A challenge in this area was bridging high-level abstract language commands with low-level physically-valid robot actions. To overcome the challenge of grounding language in physically plausible scene arrangements, StructDiffusion~\citep{liu2022structdiffusion} introduced language as a high-level constraint, such as ``\emph{Set the table in the center left, relative to you}'', to guide an object-centric language-conditioned diffusion model to sample diverse and collision-free goal poses for multiple objects, even those unseen during training. The method's performance is limited by subsequent low-level goal-conditioned policies. This highlights the need for policies that could produce complete trajectories with numerous high-quality expert demonstrations, while bringing a new bottleneck: such demonstrations are expensive to collect, and unstructured data (``play'' data) is plentiful but noisy. A key line of research is the need for greater data efficiency and scalability. For example, Scaling\&Distilling~\citep{ha2023scaling} leverages an LLM to propose textual subtasks, executes them in simulation, and \emph{distils} the successful trajectories into a compact language-conditioned diffusion policy. ChainedDiffuser~\citep{xian2023chaineddiffuser} utilizes language guides a global transformer to predict a sequence of discrete keyposes, and a local trajectory diffuser then generates smooth motion segments to connect them, unifying keypose prediction with trajectory diffusion to solve long-horizon tasks.} 

\textcolor{black}{Alternatively, LCD~\citep{zhang2024language} tackles long-horizon tasks by employing a diffusion model as a high-level planner that operates in a learned low-dimensional latent space. Language instructions guide this planner to generate a sequence of abstract goal states for execution. \citet{chen2023playfusion,ju2024rethinking,wu2025discrete} explore to learn policies in a more generalizable and reusable manner, aiming to discover and disentangle fundamental skills. PlayFusion~\citep{chen2023playfusion} introduces a skill discovery scheme, which treats language-annotated play as weak supervision. A conditional diffusion model segments free-play trajectories into latent \emph{skills} that are managed via a discrete codebook. These skills can later be recombined when given a new language instruction. Similarly, \citet{ju2024rethinking} and \citet{wu2025discrete} propose leveraging vector quantization (VQ) to map continuous action sequences into a discrete latent skill space, guided by language instructions. Whereas StructDiffusion~\citep{liu2022structdiffusion} focuses on single-step goal sampling, sequential works expand language's role to \emph{self-supervised data curation}~\citep{ha2023scaling}, \emph{skill composition}~\citep{chen2023playfusion,ju2024rethinking,wu2025discrete}, and \emph{long-term plan decomposing}~\citep{xian2023chaineddiffuser,zhang2024language}, improving data and computational efficiency.}

\textcolor{black}{For language-conditioned diffusion policies to be practical in the real world, they must overcome several challenges related to data heterogeneity and deployment constraints. A primary issue is handling diverse data sources, such as simulations, different robot platforms, and human videos. To address this, PoCo~\citep{wang2024poco} formulates policy composition as a conditional diffusion problem, allowing it to combine multiple pre-trained policies with language guidance, thus avoiding the difficulty of training a single model on disparate data distributions. Another challenge is cross-embodiment transfer. RoLD~\citep{tan2024roldrobotlatentdiffusion} tackles this by first pre-training a task-agnostic autoencoder to create a unified latent action space. A language-conditioned diffusion policy is then trained within this compact space, enabling effective knowledge transfer across different robot platforms. Furthermore, real-world datasets often have sparse or missing language annotations. To handle this, methods like MDT~\citep{reuss2024multimodal} and GR-MG~\citep{li2025gr} are designed to accept multimodal goals (e.g., language or images). GR-MG, for instance, uses the language instruction to generate a corresponding goal image with a diffusion model, then conditions its policy on both modalities for increased robustness. While these methods improve data efficiency and generalization, a fundamental architectural bottleneck remains: the high inference latency of the iterative denoising process, which is often prohibitive for real-time control. Although not language-conditioned, other works offer potential solutions; for example, some methods~\citep{lu2024manicm,prasad2024consistency} impose consistency constraints on the diffusion process to enable low-latency decision-making, suggesting a promising direction for future language-conditioned models.}

\textcolor{black}{In summary, the integration of language into diffusion-based policy learning transforms the generative process from simply modeling multi-modal action distributions to directing it toward specific, semantically meaningful goals. This addresses a core limitation of regression-based methods, which tend to average out diverse expert behaviors. Early works using language as a high-level constraint for single-step goal sampling~\citep{liu2022structdiffusion}, while relying on separate low-level controllers for execution. Researchers leveraged language for more sophisticated, structured learning paradigms, including self-supervised data curation~\citep{ha2023scaling}, hierarchical plan decomposition~\citep{xian2023chaineddiffuser,zhang2024language}, and skill discovery and composition~\citep{chen2023playfusion,ju2024rethinking,wu2025discrete}. To enhance the practicality of diffusion model-based policies for real-world applications, recent work has used language to tackle challenges of data heterogeneity and sparse annotation. This includes using language to guide the composition of multiple pre-trained policies~\citep{wang2024poco}, enabling cross-embodiment transfer via shared latent spaces~\citep{tan2024roldrobotlatentdiffusion}, and designing multimodal goal representations~\citep{reuss2024multimodal,li2025gr} that can handle missing or imperfect language inputs~\citep{li2025gr}. Although these advancements demonstrate a clear trajectory toward using language for structured and efficient learning, an open challenge remains: the inherent inference latency of the iterative denoising process of the diffusion models, which is a bottleneck for real-time robot control.}

\subsection{Summary}
\textcolor{black}{When used as a policy condition, language shifts from quantifying ``\emph{task progress}'' to dictating ``\emph{how to do it}''. In this paradigm, language serves as a direct input to policy, mapping observations and instructions to actions. This approach has been explored across RL, BC, and DP, each offering distinct advantages and challenges. In RL, conditioning the policy on language helps address sample inefficiency and multi-task generalization by enabling more structured learning, such as through memory-based meta-learning~\citep{bing2023meta} or by shaping the latent dynamics of a world model~\citep{aljalbout2025limt}. In BC, language-conditioned policies learn directly from expert demonstrations, circumventing the need for reward engineering and enabling complex long-horizon manipulations~\citep{mees2022matters, zhou2023languageconditioned, kim2025srt}. However, standard regression-based BC methods struggle with multi-modal behaviors, often averaging diverse expert actions into a single, suboptimal output. Diffusion-based policies mitigate this by using generative models to capture the full distribution of expert behaviors, with language guiding the generation process toward the intended goal. Across all three learning paradigms, there is a clear progression from simple feature conditioning toward more deeply integrated roles for language. This includes using language to modulate visual features~\citep{ye2024task}, guide hierarchical planning~\citep{mees2022matters, zhou2023languageconditioned, kim2025srt}, and structure the learning process itself~\citep{zhang2024language,wang2024poco,tan2024roldrobotlatentdiffusion}, leading to more data-efficient, generalizable, and expressive robotic manipulation. A key remaining challenge, particularly for diffusion policies, is their inference latency, which can be a bottleneck for real-time control.}

\section{Language for cognitive planning and reasoning}
\hypertarget{language-cognitive-planning-reasoning}{}
\label{sec:language-cognitive-planning-reasoning}

\begin{figure}[ht!]
    \centering
    \resizebox{.49\textwidth}{!}{
    \begin{tikzpicture}
    \selectcolormodel{cmyk}
    \hypersetup{linkcolor=black}    
        \node[circle, draw=RoyalBlue!50, fill=RoyalBlue!50, text width=3cm, align=center, minimum size=3cm]
            (title) at (0,0) {\Large\textbf{\textsf{6. Language for Cognitive Planning and Reasoning}}};
        \node[circle, draw=Violet!50, fill=Violet!50, text width=3cm, align=center, minimum size=3.5cm, above left]
            (title-61) at ([xshift=-4cm, yshift=0cm]title.north west) {\large\textbf{\textsf{6.1 Classic Neural-symbolic}}};
        \node[circle, draw=Salmon!50, fill=Salmon!50, text width=3cm, align=center, minimum size=3.5cm, above right]
            (title-62) at ([xshift=4cm, yshift=0cm]title.north east) {\large\textbf{\textsf{6.2 Empowered by LLMs}}};            
        \node[circle, draw=YellowOrange!50, fill=YellowOrange!50, text width=3cm, align=center, minimum size=3cm, below]
            (title-63) at ([yshift=-1cm]title.south) {\large\textbf{\textsf{6.3 Empowered by VLMs}}};
        \node[circle, draw=Violet!30, fill=Violet!30, text width=2.5cm, align=center, minimum size=2.5cm, above left]
            (title-61-1) at ([xshift=-1cm, yshift=1cm]title-61.west) {\large\textbf{\textsf{Learning for Reasoning}}};
        \node[circle, draw=Violet!30, fill=Violet!30, text width=2.5cm, align=center, minimum size=2.5cm, below left]
            (title-61-2) at ([xshift=-1cm, yshift=-1cm]title-61.west) {\large\textbf{\textsf{Reasoning for Learning}}};
        \node[circle, draw=Violet!30, fill=Violet!30, text width=2.5cm, align=center, minimum size=2.5cm, above right]
            (title-61-3) at ([xshift=1cm, yshift=1cm]title-61.east) {\large\textbf{\textsf{Reasoning-Learning}}};

        \node[circle, draw=Salmon!30, fill=Salmon!30, text width=2.5cm, align=center, minimum size=2.5cm, above left]
            (title-62-1) at ([xshift=-1cm, yshift=1cm]title-62.west) {\large\textbf{\textsf{Structured Planning}}};
        \node[circle, draw=Salmon!30, fill=Salmon!30, text width=2.5cm, align=center, minimum size=2.5cm, above right]
            (title-62-2) at ([xshift=1cm, yshift=1cm]title-62.east) {\large\textbf{\textsf{Planning}}};
        \node[circle, draw=Salmon!30, fill=Salmon!30, text width=2.5cm, align=center, minimum size=2.5cm, below right]
            (title-62-3) at ([xshift=1cm, yshift=-1cm]title-62.east) {\large\textbf{\textsf{Reasoning}}};

        \node[circle, draw=YellowOrange!30, fill=YellowOrange!30, text width=2.5cm, align=center, minimum size=2.5cm, right]
            (title-63-1) at ([xshift=2cm, yshift=-2cm]title-63.east) {\large\textbf{\textsf{Contrastive}}};
        \node[circle, draw=YellowOrange!30, fill=YellowOrange!30, text width=3cm, align=center, minimum size=2.5cm, below]
            (title-63-2) at ([yshift=-1cm]title-63.south) {\large\textbf{\textsf{Autoregressive}}};
        \node[circle, draw=YellowOrange!30, fill=YellowOrange!30, text width=2.5cm, align=center, minimum size=2.5cm, left]
            (title-63-3) at ([xshift=-2cm, yshift=-2cm]title-63.west) {\large\textbf{\textsf{Generative}}};
        

        \path (title) to[circle connection bar switch color=from (RoyalBlue!50) to (Violet!50)] (title-61);
        \path (title) to[circle connection bar switch color=from (RoyalBlue!50) to (Salmon!50)] (title-62);
        \path (title) to[circle connection bar switch color=from (RoyalBlue!50) to (YellowOrange!50)] (title-63);
        \path (title-61) to[circle connection bar switch color=from (Violet!50) to (Violet!30)] (title-61-1);
        \path (title-61) to[circle connection bar switch color=from (Violet!50) to (Violet!30)] (title-61-2);
        \path (title-61) to[circle connection bar switch color=from (Violet!50) to (Violet!30)] (title-61-3);
        \path (title-62) to[circle connection bar switch color=from (Salmon!50) to (Salmon!30)] (title-62-1);
        \path (title-62) to[circle connection bar switch color=from (Salmon!50) to (Salmon!30)] (title-62-2);
        \path (title-62) to[circle connection bar switch color=from (Salmon!50) to (Salmon!30)] (title-62-3);
        \path (title-63) to[circle connection bar switch color=from (YellowOrange!50) to (YellowOrange!30)] (title-63-1);
        \path (title-63) to[circle connection bar switch color=from (YellowOrange!50) to (YellowOrange!30)] (title-63-2);
        \path (title-63) to[circle connection bar switch color=from (YellowOrange!50) to (YellowOrange!30)] (title-63-3);

        \node[below, align=center]at([yshift=-9cm]title.south){\Large \textbf{\textsf{Core idea:}} \textsf{``think'' in language to structure behavior}};
        \draw [-{Triangle Cap []. Fast Triangle[]}, draw=Violet!30, line width=10pt] (title-61-3) to[out=0, in=-180] node[midway, above, sloped, yshift=2pt]{\Large\textbf{\textsf{\textcolor{BrickRed}{enhance}}}} (title-62-1);

    \end{tikzpicture}
    }
    \caption{\textcolor{black}{Taxonomy of Sec. \ref{sec:language-cognitive-planning-reasoning} Language for cognitive planning and reasoning.}}
    \label{fig:taxonomy-sec6}
\end{figure}

\textcolor{black}{The previous sections explore how language can quantify \textit{task progress} (state evaluation in Sec. \ref{sec:language-for-state-evaluation}) or \textit{how} to do it (policy conditioning in Sec. \ref{sec:language-as-policy-condition}). In both paradigms, language serves as an external instruction that guides the robot's low-level behavior. This section explores a more cognitive role for language, utilizing it as an internal tool for reasoning and planning. We now turn to our third key research question: \textit{How can a robot ``think'' in language to structure its own behavior?} This involves leveraging language not just to follow commands, but to reason about the world, decompose complex problems reasonably, and formulate strategies, enabling more autonomous and intelligent manipulation in real-world environments. This perspective is inspired by cognitive science, which posits that human language comprehension is dually embodied and symbolic~\citep{louwerse2008language}. While embodied approaches connect language to human's perceptual and motor experiences~\citep{goldin2005hearing} (much like a robot policy grounds ``grasp the cup'' in sensorimotor control), symbolic approaches highlight that language also derives meaning from the abstract interdependencies between words~\citep{kintsch1998comprehension}. This symbolic structure allows humans to reason efficiently, providing a ``shortcut'' to meaning without needing to constantly simulate every physical detail~\citep{louwerse2008language}.}

\begin{table*}[htbp]
    \centering
    \caption{\textcolor{black}{Comparison of representative state-of-the-art methods for \textbf{Section \ref{sec:language-cognitive-planning-reasoning} Language for cognitive planning and reasoning}. By contrasting classic neuro-symbolic systems with modern LLM- and VLM-empowered approaches , this table summarizes the key mechanisms, advantages, and limitations of using language to internally structure robot behavior and bridge abstract reasoning with perceptual reality.}}
    \label{tab:language_as_cognitive_system}
    \small
    \begin{tabular}{p{0.12\linewidth}| p{0.1\linewidth}| p{0.21\linewidth}| p{0.21\linewidth}| p{0.21\linewidth}}
        \toprule[1pt]
        \textbf{Method} & \textbf{Cognitive System} & \textbf{Reasoning Mechanism} & \textbf{Key Advantages} & \textbf{Key Disadvantages} \\
        \midrule[0.2pt]
        DANLI\qquad\citep{zhang-etal-2022-danli} & Classic Neuro-symbolic & Learning for reasoning via dialogue history and symbolic sub-goals. & Allows the agent to reason about progress and dynamically recover from errors. & Relies on labor-intensive, task-specific Knowledge Graph construction. \\
        \midrule[0.2pt]
        SayCan\qquad\citep{ahn2022can} & LLM-Empowered & Open-loop planning integrated with affordance functions. & Translates natural language into semantically relevant and physically feasible action sequences. & Assumes faultless skill execution and lacks real-time feedback for replanning. \\
        \midrule[0.2pt]
        SayPlan\qquad\citep{rana2023sayplan} & LLM-Empowered & Closed-loop planning via 3D scene graphs and semantic simulators. & Operates successfully in large-scale environments by providing continuous textual feedback. & Performance remains capped by the underlying LLM's reasoning capabilities and potential latency. \\
        \midrule[0.2pt]
        Code as Policies\qquad\citep{liang2023code} & LLM-Empowered & Auto-regressive code generation for orchestrating policy logic. & Flexibly recomposes API calls to generalize better to unseen objects. & Code-generation LLMs struggle with complex commands and may call nonexistent functions. \\
        \midrule[0.2pt]
        PaLM-E\qquad\citep{driess2023palme} & VLM-Empowered & Text generation via a massive, single embodied reasoning model. & Translates high-level commands into robot-executable plans with little or no robot-task-specific finetuning in some evaluated settings. & Highly resource-intensive to deploy and run.\\
        \midrule[0.2pt]
        SuSIE\qquad\citep{black2023zero} & VLM-Empowered & Image generation for visualizing sequential subgoals. & Decouples high-level semantic understanding from low-level control optimization. & Generated images can contain physically implausible details, creating unreliable control signals.\\
        \midrule[0.2pt]
        LEMMo-Plan\qquad\citep{chen2025lemmo} & LLMs-driven structured planning (Symbolic) & Incorporates multi-modal demonstrations (tactile and force-torque data) into PDDL symbolic planning. & Allows the LLM to reason about ``invisible'' events like cable tension, defining robust force-based skill conditions. & Integrating LLMs with PDDL can introduce slow symbolic search speeds, delaying robot action. \\
        \midrule[0.2pt]
        BETR-XP-LLM\qquad\citep{styrud2025automatic} & LLMs-driven structured planning (Behavior Trees) & Casts the LLM as a ``repair agent'' to propose minimal preconditions and matching subtrees for execution failures & Permanently integrates verifiable fixes into the policy, making the robot more robust with every failure. & Relying on LLMs for runtime feedback and replanning increases token and latency costs. \\
        \bottomrule[1pt]
    \end{tabular}
\end{table*}

\textcolor{black}{This cognitive duality mirrors a fundamental challenge and a corresponding solution in artificial intelligence. On one hand, neural systems excel at perceptual intelligence, learning directly from unstructured data (e.g., images, videos, or texts). Their information processing units are typically vectors, inherently lacking explicit reasoning capabilities. On the other hand, symbolic systems offer powerful, interpretable reasoning but are brittle when faced with noisy, real-world sensory input~\citep{Yu2021ASO}. An ideal solution is a hybrid approach that combines the strengths of both: neural-symbolic learning systems~\citep{DBLP:series/faia/BesoldGBBDHKLLPPPZ21,Yu2021ASO}. Following the idea of these frameworks, in the field of language-conditioned robotic manipulation, language serves as the bridge that connects the neural and symbolic components, grounding abstract symbols in perceptual reality, enhancing both interpretability and robustness of the robot policies. This section will delve into approaches that leverage this paradigm, exploring both classic neural-symbolic methods and the recent emergence of Foundation Models (LLMs, VLMs, VLAs) as powerful engines for task planning and reasoning in robotics.}

\textcolor{black}{Figure \ref{fig:taxonomy-sec6} displays the taxonomy of this section. Moreover, to synthesize the diverse approaches that enable robots to use language as an internal tool to reason, decompose problems, and structure their behavior, Table \ref{tab:language_as_cognitive_system} provides a detailed comparison of the cognitive planning and reasoning methods reviewed in this section. These methods are categorized based on the type of cognitive system employed to bridge \emph{abstract textual reasoning} with \emph{perceptual reality}. The table spans classic neuro-symbolic frameworks to modern paradigms empowered by the broad open-vocabulary reasoning priors of Large Language Models and the grounded perception of Vision-Language Models. This side-by-side comparison emphasizes how each reasoning mechanism balances interpretability and structural rigor against scalability constraints and hallucination risks.}

\hypertarget{neuro-symbolic-approaches}{}
\subsection{Classic neuro-symbolic approaches}

Neuro-symbolic artificial intelligence~\citep{DBLP:series/faia/BesoldGBBDHKLLPPPZ21}, combining both neural and symbolic traditions to solve tasks, continually develops alongside popular data-driven machine-learning approaches. In this context, ``neuro'' refers to neural networks, while ``symbolic'' refers to high-level (human-readable) representations like logic or graphs. In the field of robot manipulation, integrating neural and symbolic approaches can enhance the robot's reasoning capabilities and the interpretability of its decisions by grounding abstract knowledge in the robot's perceptual world. 
This section discusses key neuro-symbolic methods in language-conditioned manipulation, categorized according to the framework by~\citet{Yu2021ASO}, namely \emph{Learning for reasoning}, \emph{Reasoning for learning}, and \emph{Learning-reasoning}.

\subsubsection{Learning for reasoning}
\hypertarget{learning-for-reasoning}{}
\textcolor{black}{In this category, neural networks serve as perception and feature extraction modules, converting unstructured data (e.g., images, language) into symbolic representations. A separate symbolic system then uses these symbols to perform high-level reasoning and planning. An early system proposed by ~\citet{tenorth2010understanding} sourced task knowledge from websites like wikiHow. It parsed natural language and used knowledge bases like WordNet~\citep{fellbaum1998wordnet} and Cyc ontology~\citep{matuszek2006introduction} to resolve word senses and map them to a formal logic representation, generating a symbolic plan. This plan was then translated into an executable language for the robot's planner for further optimization. \citet{she2014teaching} present a framework where a robotic arm learns new high-level actions through natural language. Their system first processes human instructions into a symbolic ``Grounded Action Frame'' by parsing the language and grounding it to objects perceived in the environment. This symbolic representation allows a classic STRIPS~\citep{fikes1971strips} planner to dynamically generate new sequences of primitive movements to accomplish the learned action in novel and more complex situations, demonstrating flexible application of the acquired knowledge.} 

\textcolor{black}{HiTUT~\citep{zhang-hierarchical2021} uses a unified transformer architecture to learn a hierarchical task structure from language and vision. Language is provided at two levels of granularity: high-level goal instructions guide sub-goal planning, while low-level step-by-step instructions inform navigation and manipulation actions. The neural model extracts task components from these inputs, enabling hierarchical planning. DANLI~\citep{zhang-etal-2022-danli} operates on a task-oriented dialogue history between a human commander and the robot. Its neural component, a ``task monitor'', processes the dialogue and action history to extract symbolic sub-goals, representing both completed and future steps. A traditional symbolic planner then uses these sub-goals and a dynamically built semantic map of the environment to generate low-level action plans, allowing the agent to reason about its progress and recover from errors. \citet{bartoli2022knowledge} focuses on long-term, incremental knowledge acquisition through human-robot interaction. Here, language serves as a verification and correction tool. The robot makes a prediction about a perceived object and asks the human a direct question (e.g., ``\textit{Is it true that this bottle has material metal?}''). The human's verbal response (``\textit{Yes}'', ``\textit{No}'', or a correction like ``\textit{Plastic}'') is used to directly update the robot's knowledge graph. This approach uses Knowledge Graph Embeddings and Continual Learning to generalize from human-provided feedback and prevent catastrophic forgetting over long interaction periods.}

\subsubsection{Reasoning for learning}
\hypertarget{reasoning-for-learning}{}
\textcolor{black}{\textit{Reasoning for Learning} indicates that the symbolic system provides structure, rules, or knowledge that constrains and guides a neural network's learning process or decision-making. The final output is typically generated by the neural component, but its behavior is shaped by symbolic reasoning.  \citet{Misra2014TellMD} propose a model that grounds free-form natural language instructions in a robot's environment. The language is first parsed into an intermediate symbolic representation of ``verb clauses''. Symbolic domain knowledge, encoded in the STRIPS formal language~\citep{fikes1971strips}, provides preconditions and effects for robot actions. This symbolic knowledge is integrated into a trainable energy function, modeled as a Conditional Random Field~\citep{sutton2012introduction}, which learns to map the verb clauses to a valid sequence of executable controller instructions that respects both the language command and the physical constraints of the world.  \citet{nguyen2019reinforcement} integrate the prior knowledge in KG, capturing spatial relationships between target objects, with the visual and textual information of observation and language instructions to enhance the agent's generalization ability in unseen environments. \citet{silver2022learning} learn neuro-symbolic skills from demonstrations without direct language commands during training. Instead, the symbolic system provides a predefined ``language'' of symbolic predicates (e.g., Grasped, Pressed) that structure the learning process. The system learns symbolic operators alongside neural policies and subgoal samplers from demonstrated trajectories. At test time, a bilevel planner uses these learned skills to solve new tasks specified by a formal symbolic goal, demonstrating how symbolic reasoning structures both learning and planning.} 

\subsubsection{Learning-reasoning approaches}
\hypertarget{learning-reasoning}{}
\textcolor{black}{This category includes integrated systems where neural and symbolic components work in a tight loop, mutually informing and enhancing each other to produce the final results. Both neural and symbolic components are critical to the reasoning process. \citet{chai2018itl} introduce an Interactive Task Learning framework that incorporates language into a rich, interactive dialogue between a human and a robot, complemented by physical demonstrations. A key concept is learning the physical causality of action verbs (e.g., the action ``slice'' results in the state ``in pieces'') to bridge symbolic actions and their perceptual outcomes~\citep{gao2016physical}. A hypothesis space of such verb representations can be acquired incrementally through continuous interaction with the environment~\citep{she2016incremental}. As the world is full of uncertainties, a dialogue policy can be learned to better capture the connection between symbolic representations and the perception from the world~\citep{she2017interactive}. Through this interaction, the agent simultaneously learns a grounded symbolic task structure and improves its perceptual understanding of cause-and-effect of actions, showing how neural and symbolic representations can \emph{co-develop}~\citep{gao2018action}.}

\textcolor{black}{\citet{namasivayam2023learning} propose a system that translates natural language instructions into executable programs. First, a symbolic Language Reasoner parses the instruction into a hierarchical program using a Domain-Specific Language. This symbolic program is then executed by a neural Visual Reasoner, which operates on neural object-centric representations of the scene to ground the program's symbols (e.g., ``the red block'') to specific objects. Finally, a neural Action Simulator predicts the physical outcome of the grounded actions. This architecture creates a tight loop where symbolic parsing guides neural grounding, which in turn informs neural prediction.} \citet{miao2023long} use a vision module, which takes the RGB images of the scene as input and predicts the labels, bounding boxes, and relation predicate labels of all entities to generate a scene graph. Then, the generated scene graph is utilized as the input of the regression planning network~\citep{xu2019regression} and the task instructions to output intermediate goals (plan).

\textcolor{black}{In summary, language instructions in the above parallels serve different roles. In \emph{learning-for-reasoning}, language specifies task structure and entities that the symbolic layer reasons over. In \emph{reasoning-for-learning}, language instantiates symbolic predicates or constraints that steer neural learning toward logically valid behaviors. In \emph{learning-reasoning}, language facilitates an interplay between symbolic and neural components, enhancing both learning and reasoning capabilities of the robot policy. While these classic neuro-symbolic frameworks provide interpretable and structured reasoning for robotic manipulation, they also face several limitations that can hinder their application in complex real-world environments, mainly including: 
\begin{itemize}[leftmargin=10pt]
    \item \textbf{Labor-intensive KG construction:} Approaches like DANLI~\citep{zhang-etal-2022-danli} or KGE-based continual learning~\citep{bartoli2022knowledge} presuppose a task-specific KG. Populating and maintaining these graphs quickly becomes a bottleneck in dynamic open-world settings where new objects and relations may appear frequently.
    \item \textbf{Manual symbol engineering and ontology drift:} The above systems rely on a hand-crafted inventory of symbols (such as object types~\citep{bartoli2022knowledge,kwak2022semantic}, predicates~\citep{Misra2014TellMD,silver2022learning}, operators~\citep{chai2018itl}). Extending the ontology to new domains requires human curation, ontology alignment, and often re-training of perception modules.
    \item \textbf{Limited coverage of commonsense and real-world knowledge:} Symbolic rules capture only what has been explicitly encoded. They lack the broad background knowledge, such as object materials, physical affordance, social conventions, and other contextual factors needed for robust behavior in unstructured environments.
    \item \textbf{Scaling issues in long-horizon tasks:} As plan length grows, search over discrete operators and grounding choices explodes combinatorially, leading to slow inference or heavy reliance on humans' sub-goal supervision~\citep{zhang-etal-2022-danli}.
\end{itemize}}

\hypertarget{llms}{}
\subsection{Empowered by large language models}
\tikzstyle{arrow} = [->,>=stealth]
\begin{figure}[!thbp]
    \centering
    \begin{adjustbox}{width=0.47\textwidth}
        \begin{tikzpicture}[node distance=2cm]
        \node[fill=Violet!30, circle,  minimum width=2.8cm, minimum height=2.8cm, text width=2cm, text centered](llm){Empowered by LLMs};
        \node[circle, minimum width=3.1cm, minimum height=3.1cm, draw=Violet!30, line width=0.5mm](llm_border){};

        \node[fill=RoyalBlue!30, circle, radius=3cm, text width=2cm, left=1.5cm of llm, text centered](planning){\hyperlink{planning}{\textcolor{black}{Planning}}};
        \node[fill=Salmon!30, circle, radius=3cm, text width=2cm, right=1.5cm of llm, text centered](reasoning){\hyperlink{reasoning}{\textcolor{black}{Reasoning}}};
        \draw [{Circle[length=3mm, width=3mm]}-{Triangle Cap[]. Fast Triangle[] Fast Triangle[] Fast Triangle[] Fast Triangle[] Fast Triangle[]}, draw=RoyalBlue!30, line width=1.3mm] ([xshift=0mm]llm.west) -- (planning.east);
        \draw [{Circle[length=3mm, width=3mm]}-{Triangle Cap[]. Fast Triangle[] Fast Triangle[] Fast Triangle[] Fast Triangle[] Fast Triangle[]}, draw=Salmon!30, line width=1.3mm] ([xshift=0mm]llm.east) -- (reasoning.west);
        \node[text width=4cm, above=1cm of planning](open-loop-planning){\hyperlink{open-loop-planning}{\textcolor{black}{Open-loop Planning}} \\ \citetitle{SayCan}{ahn2022can};~\citetitle{Language-planner}{huang2022language};~\citetitle{KNOWNO}{ren2023robots};\\\citetitle{TaPA}{wu2023embodied};~\citetitle{DEPS}{wang2023describe}};
        \draw[arrow, gray, dashed, line width=0.7mm] (planning.north) -- (open-loop-planning.south);
        \node[text width=4cm, below=1cm of planning](close-loop-planning){\hyperlink{closed-loop-planning}{\textcolor{black}{Closed-loop Planning}} \\ \citetitle{Text2motion}{lin2023text2motion};~\citetitle{SayPlan}{rana2023sayplan};~\citetitle{LLM-MCTS}{zhao2023large};~\citetitle{Alphablock}{jin2023alphablock};~\citetitle{RobLM}{chalvatzaki2023learning}};
        \draw[arrow, gray, dashed, line width=0.7mm] (planning.south) -- (close-loop-planning.north);
        \node[text width=3cm, above=1cm of reasoning, xshift=-1.7cm](summarization){\hyperlink{summarization}{\textcolor{black}{Summarization}}\\\citetitle{TidyingUp}{rasch2019tidy};\\\citetitle{OFSM}{yan2021quantifiable};\\\citetitle{Housekeep}{kant2022housekeep};\\\citetitle{Tidybot}{wu2023tidybot}};
        \draw[arrow, gray, dashed, line width=0.7mm] (reasoning.north west) -- (summarization.south);
        \node[text width=3cm, above=1cm of reasoning, xshift=1.5cm](code-generation){\hyperlink{code-generation}{\textcolor{black}{Code Generation}} \\ \citetitle{Voyager}{wang2023voyager};\\\citetitle{Code-as-policies}{liang2023code};~\citetitle{ScalingUp}{ha2023scaling};~\citetitle{ProgPrompt}{singh:progprompt:ar}~\citetitle{TranslatingGoals}{xie2023translating}};
        \draw[arrow, gray, dashed, line width=0.7mm] (reasoning.north east) -- (code-generation.south);
        \node[text width=3cm, below=1cm of reasoning, xshift=-1.2cm](prompt-engineering){\hyperlink{prompt-engineering}{\textcolor{black}{Prompt Engineering}} \\ \citetitle{Socratic Models}{zeng2022socratic};\\\citetitle{ECoT}{zawalski2024roboticcontrolembodiedchainofthought}};
        \draw[arrow, gray, dashed, line width=0.7mm] (reasoning.south west) -- ([xshift=-0.3cm]prompt-engineering.north);
        \node[text width=3cm, below=1cm of reasoning, xshift=2cm](iterative-reasoning){\hyperlink{iterative-reasoning}{\textcolor{black}{Iterative Reasoning}} \\ \citetitle{Inner monologue}{huang2022inner};~\citetitle{Voyager}{wang2023voyager};~\citetitle{AIC MLLM}{xiong2024autonomous}};
        \draw[arrow, gray, dashed, line width=0.7mm] (reasoning.south east) -- ([xshift=-0.5cm]iterative-reasoning.north);
        
        \end{tikzpicture}
    \end{adjustbox}

    \caption{\textcolor{black}{Illustration of approaches empowered by LLMs.}}
    \label{fig:llm-summary}
\end{figure}

\textcolor{black}{The limitations of classic neuro-symbolic methods listed above stem from the fact that classic pipelines rely on \emph{explicitly engineered} symbolic knowledge. Recent \emph{foundation models}, particularly LLMs, offer a complementary path. Trained on trillions of tokens of numerous unstructured data (such as internet text)~\citep{naveed2025comprehensive}, \textcolor{black}{LLMs encode broad textual priors and often exhibit useful instruction-following, commonsense understanding, and contextual reasoning without domain-specific retraining}. Famous LLMs like GPT-families~\citep{brown2020language,achiam2023gpt,hurst2024gpt}, Claude~\citep{bai2022constitutional}, LLaMA-2~\citep{touvron2023llama2}, Gemini~\citep{team2023gemini}, DeepSeek-R1~\citep{guo2025deepseek}, and Qwen3~\citep{yang2025qwen3}. Instead of extending a hand-written ontology, one can \emph{prompt} the model in natural language (or provide a few demonstrations~\citep{gao2023makes}) to obtain a structured plan, a piece of executable code, or a step-by-step chain-of-thought reasoning~\citep{wei2022chain}.}

\textcolor{black}{This capability allows LLMs to serve as a powerful tool that combines planner and reasoner for language-conditioned robots, bridging the gap between high-level human intent and a plausible sequence of executable steps. Early work in this area demonstrated this potential by using LLMs as task planners. For example, Saycan~\citep{ahn2022can} translates natural language instructions into intermediate action sequences. When combined with affordance functions that ground these actions in the robot's current environment, LLMs can generate behaviors that are both semantically relevant and physically feasible. Subsequent research has expanded on this, leveraging LLMs to solve a diverse set of long-horizon manipulation tasks in large-scale~\citep{rana2023sayplan} or open-ended environments like Minecraft~\citep{wang2023describe}. More generally, the pattern recognition abilities of LLMs can be exploited beyond just language-based reasoning by converting robot observations and actions to numerical text~\citep{dipalo2024kat}.}

While LLMs address many of the scalability issues of classic methods for robotic manipulation, they introduce their own potential and non-negligible issues.
\begin{itemize}[leftmargin=10pt]
    \item \textbf{The grounding problem}: LLMs are prone to confidently hallucinate predictions~\citep{rana2023sayplan}, and generate a plausible but not feasible plan~\citep{ahn2022can, singh:progprompt:ar}, because LLMs will plan an action involving objects unavailable in the real environment~\citep{huang2022language}. 
    \item \textbf{Ambiguity in language}: Natural language is highly ambiguous, especially in expressing spatial and geometrical relationships such as ``\textit{moving faster}''  or ``\textit{placing objects slightly left}''~\citep{liang2023code,mees2021composing,mees20icra_placements,mees17iros}. 
    \item \textbf{Lack of feedback and reactivity}: When operating as open-loop planners, LLMs are not inherently aware of the outcomes of their proposed actions. They can generate unsafe plans (e.g., putting metal in a microwave) and cannot dynamically replan when an action fails without a corrective mechanism to provide real-time feedback~\citep{ren2023robots}.
\end{itemize}

The following sections will review how recent works in robot manipulation are addressing these challenges, categorizing methods based on how they leverage LLMs for planning and reasoning, as illustrated in Figure~\ref{fig:llm-summary}. \textcolor{black}{Furthermore, we explore the methods that combine LLMs with symbolic systems to achieve interpretable task planning and reasoning.}
\\
\hypertarget{planning}{}
\subsubsection{Planning}
\label{sssec: planning}
Planning refers to a process wherein an agent decomposes a high-level task into subgoals or sequences, which are a set of learned actions to accomplish a specific objective~\citep{rana2023sayplan}. Recent research has highlighted the remarkable capabilities of LLMs in planning tasks, including semantic classification, common-sense reasoning, and contextual understanding. These capabilities can potentially be harnessed by embodied agents for task planning. \citet{huang2022language} posit that LLMs inherently possess the knowledge required to achieve goals without further training. Specifically, pre-trained autoregressive LLMs necessitate only minimal prompts~\citep{wang2023describe, singh:progprompt:ar, ren2023robots,ding2023task} or no prompts at all~\citep{ahn2022can,huang2022language} to generate coherent plans expressed in natural language. In contrast, traditional learning-based planning methods rely on intricate heuristics~\citep{vallati20152014} and extensive training datasets~\citep{ceola2019robot}. Collecting such vast data is often prohibitive, especially for diverse tasks and unpredictable real-world scenarios.

However, pre-trained LLMs encode a large amount of task-agnostic knowledge and lack state feedback from the environment, which leads to task planning with LLMs often struggling with the hallucination issue. LLMs tend to generate plausible but infeasible plans, such as an action involving an unavailable object in the environment~\citep{ahn2022can, singh:progprompt:ar}. \citet{ding2022robot} mention that grounding domain-independent knowledge into a specific domain with many domain-relevant constraints is challenging for task planning with LLMs. Therefore, we ask: \textit{how to ground planning into the environment?} i.e., \textit{how to enable LLMs to generate more feasible plans and executable actions?} Recent works integrate LLM with external components or leverage the code generation capability of LLM to remedy these problems. Here, we discuss \emph{open-loop} and \emph{closed-loop} planning. Figure~\ref{fig:open_close_loop} demonstrates the overall process of both approaches.
\input{tikz/open_close_loop2}

\hypertarget{open-loop-planning}{}
\noindent \textbf{Open-loop planning}\\[4pt]
In recent years, many researchers have designed LLM-based planners that integrate LLMs with different external components. These external components can provide LLMs with additional input information or embodied feedback from the environment, thus ensuring the output actions adhere to the constraints and achieve the goal. \citet{ahn2022can} leverage the affordance function to quantify the success rate of each action in the current state, and LLMs reorder the set of predicted actions and output the most accessible action. \citet{ren2023robots} combine LLMs with conformal prediction to measure and align uncertainty. LLM planning can be formulated as multiple-choice Q\&A. It generates a series of candidate actions in the next step. Then, LLMs choose the correct options by a conformal prediction threshold calculated based on a user-specified success rate. If the robot cannot select the only correct option, it will ask humans for help, thus aligning the uncertainty, e.g., ``\textit{Put a plastic bowl in the microwave}'' and ``\textit{Put a metal bowl in the microwave}'' for the task ``\textit{Heating up the food}''. Moreover, \citet{huang2022language} let two pre-trained LLMs play different roles in task planning. A pre-trained causal LLM decomposes the high-level task into a sensible mid-level action plan. The other pre-trained masked LLMs leverage semantic similarity to translate these mid-level actions into admissible learned actions, e.g., translating the action ``\textit{squeeze out a glob of lotion}'' into ``\textit{pour the lotion into right hand}''. \citet{wu2023embodied} utilize CLIP~\citep{radford2021learning} and a masked RCNN as an open-vocabulary object detector to collect multiple RGB images and predict a list of objects existing in the scene, showcasing the open-loop model guide planning with auxiliary information and mitigating the issues of LLMs to generate multi-step actions involving unavailable objects in the specific environment.

Nevertheless, these prior approaches still cannot solve the long-horizon tasks successfully, which is caused by two major issues. Firstly, the prior approaches adopt short-term or open-loop execution strategies, trusting LLMs to generate the correct strategies without accounting for the geometric dependence over a skill sequence~\citep{lin2023text2motion}, which is an essential factor in solving long-horizon tasks. Open-loop approaches, like Saycan~\citep{ahn2022can,huang2022language,ren2023robots,wu2023embodied}, have distinct planning and control components that are implemented separately. LLMs, as offline planners, never received embodied feedback to reflect on previous executions. Therefore, most open-loop approaches must assume faultless skill execution in solving long-horizon tasks~\citep{ahn2022can,huang2022language}, \textcolor{black}{or utilizing user cases to constrain the LLM's planning in specific domains~\citep{singh2025llm}}. These assumptions or constraints limit their scalability and the success rate of solving tasks in a new environment. Secondly, some approaches output a plan that can be viewed as a one-shot plan~\citep{wang2023describe}, i.e., the approaches have no replanning function. LLMs lacking state feedback do not replan the generated skill but only focus on reasoning over more accessible skills in the next step. Obviously, it is challenging for these open-loop approaches to generate a flawless one-shot plan that can directly solve long-horizon tasks. On the one hand, various complex preconditions and unforeseen accidents can occur in the real world, making the one-shot plan non-executable easily. On the other hand, many challenging long-horizon tasks, such as household tasks or rearrangement tasks, involve multiple objects and a series of chronologically linked subgoals, making it difficult to cover all of them in a one-shot plan.

\hypertarget{closed-loop-planning}{}
\noindent \textbf{Closed-loop planning}\\[4pt]
To address the aforementioned issues, more recent studies ~\citep{lin2023text2motion,zhao2023large,jin2023alphablock,chalvatzaki2023learning} integrate LLMs with external components in a closed loop, where these external components can provide embodied state feedback to LLMs. Then, LLMs can constantly replan more executable skills until the plan succeeds completely. This iterative replanning leverages the strong contextual reasoning of LLMs and continuous feedback through the closed loop to improve the scalability and generalizability~\citep{rana2023sayplan, ha2023scaling}. For example, compared to Saycan~\citep{ahn2022can}, which only accomplishes tasks in kitchen scenarios, Sayplan~\citep{rana2023sayplan} operates in a larger-scale environment that covers almost all daily office scenarios. Specifically, Sayplan leverages a hierarchical 3D scene graph to represent the environment, while a scene graph simulator generates textual feedback to LLMs, combining the scene graph's predicates, current states, and affordances to enhance the planning process. \citet{lin2023text2motion} propose Text2Motion to leverage Sequencing Task-Agnostic Policies (STAP)~\citep{agia2023stap} as geometric feasibility approaches. LLMs will plan a new skill if STAP finds the previous plan failing to adhere to geometric dependence. \citet{jin2023alphablock} and \citet{huang2022inner} utilize a pre-trained visual transformer as a scene descriptor, which can translate visual observations into real-time textual feedback for LLMs. \citet{kwon2024language} show that a single task-agnostic prompt could predict dense robot end-effector trajectories with closed-loop feedback to check performance and correct trajectories where necessary. \textcolor{black}{With an inner-feedback mechanism, \citet{wu2025selp} propose SELP to mitigate hallucination issues of LLMs for long-horizon tasks (i.g. producing unfeasible or unsafe plans), which incorporates Linear Temporal Logic (LTL) formulation to prune out LLM's unsafe plans, adhere the plan to follow the temporal constraint of natural language commands, and allow replanning when no valid plan exists. However, this method relies on a predefined LTL specification for specific environments, which limits its flexibility. Moreover, \citet{asuzu2025human} incorporate continuous human user feedback into the LLM planning process, which forms a closed-loop corrective mechanism to refine the LLM plans.}

\hypertarget{reasoning}{}
\subsubsection{Reasoning}

In general, reasoning refers to the ability of a policy to mimic human-like thinking and make inferences using observation embedding or external information. In robot manipulation, planning and reasoning are two crucial capabilities that embodied agents use to solve multi-step and long-horizon tasks. They are distinct but highly interconnected. Feasible reasoning at each step ensures the generation of an executable action plan, i.e., the reasoning is a prerequisite for planning. Some prior surveys do not make a clear distinction between planning and reasoning~\citep{tellex2020robots,DBLP:journals/corr/abs-2405-13245}. In our work, we categorize them into two different sections. In section \textit{Planning}\ (\S\ref{sssec: planning}), many works are developed for a closed world, assuming that complete knowledge of the world is provided and the agent can enumerate all possible states~\citep{ding2022robot}. LLM-based models utilizing auxiliary information ~\citep{ahn2022can,ding2023task} or feedback~\citep{wang2023describe,lin2023text2motion,jin2023alphablock} to solve spatial and geometric dependencies in action sequences. For unseen objects, LLM-based planners are trained to avoid them rather than to generalize them. Meanwhile, many other researchers~\citep{ding2022robot, kant2022housekeep} operate their agents in an open world, leveraging different types of reasoning to improve the generalizability of unseen objects or instructions. They improve the agent's performance by making it robust to unforeseen situations. We categorize these works into summarization, prompt engineering, and code generation in this section.

\input{tikz/summarization}
\hypertarget{summarization}{}
\noindent \textbf{Summarization}\\[4pt]
Summarization, also called inductive reasoning, is a cognitive ability to draw logical conclusions or provide a general strategy from limited information~\citep{funke2010complex}. Summarization of LLMs shows the potential of embodied agents in the household scenario. The rearrangement task of tidying up a room is challenging for classical methods~\citep{batra2020rearrangement,gan2021threedworld}. On the one hand, where objects are placed is highly personal, depending on different people's preferences and habits. On the other hand, it is impractical to enumerate all the objects that exist in the task-specific domain and to specify the goal state for every novel object. Thus, prior models of rearrangement tasks that specify target locations manually struggle to execute effectively in large-scale or real-world environments~\citep{rasch2019tidy,yan2021quantifiable}. To solve this issue, Housekeep utilizes a large-scale dataset of human preferences instead of learning from a small set of tidying samples~\citep{kant2022housekeep}. Consequently, the Housekeeper assesses the capability to reason the target location and rearrange unseen objects. \citet{wu2023tidybot} argue that such a rearrangement preference is still generic rather than personal. They have constructed a mobile manipulator, Tidybot, which reasons individual preference by few-shot prompting and summarizes a general strategy, e.g., through textual prompts ``\textit{yellow shirts go in the drawer}" and ``\textit{white socks go in the drawer}". Tidybot outputs ``\textit{lighter-colored clothes go in the drawer}", as shown in Figure \ref{fig:summarization}. Tidybot can decide where to place the unseen object in the test by executing a corresponding preference strategy.
While an agent with inductive reasoning can enhance their performance for unseen objects, a significant drawback is that the LLMs encode much task-agnostic knowledge, resulting in the failure to generate an entirely correct summary. For example, a rearrangement category may be too specific and not generalize well to unseen objects~\citep{wu2023tidybot}.\\[-10pt]

\begin{figure}[htbp]
    \centering
    \begin{adjustbox}{width=0.43\textwidth}
        \begin{tikzpicture}[node distance=2cm]
            \node (question)[text width=6cm, rectangle, rounded corners, 
            draw=JungleGreen!60, 
            label={[label distance=0cm]90:\small \textsf{Question}}]{\small \textsf{Prepare a cup of coffee.}};
            \node (agent) [inner sep=0pt, left of=question, circle, draw=black, thick, xshift=-2cm, label={[label distance=0cm]-90:\small \textsf{Agent}}]{\includegraphics[width=0.5cm]{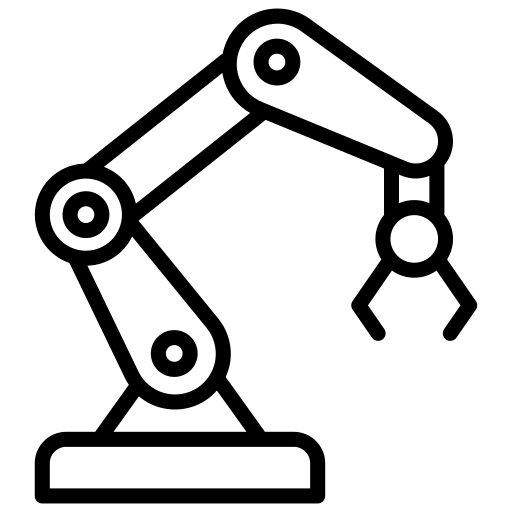}};
            \node (answer)[text width=6cm, rectangle, rounded corners, 
            draw=JungleGreen!60, 
            below of=text1, yshift=0.25cm, label={[label distance=0cm]90:\small \textsf{Chain-of-Thought}}]{\small \textsf{1. locate and gather a coffee mug, coffee grounds, a coffee maker and water.\\2. Ensure the coffee maker is ready.\\3. Add coffee grounds.\\4. Start the coffee maker.\\5. Wait for the coffee to brew.\\6. Pour the coffee.\\7. Serve the coffee and clean up.}};
            \node (chat-gpt) [inner sep=0pt, right of=answer, circle, draw=black, thick, xshift=2cm, label={[label distance=0cm]-90:\small \textsf{LLMs}}]{\includegraphics[width=0.5cm]{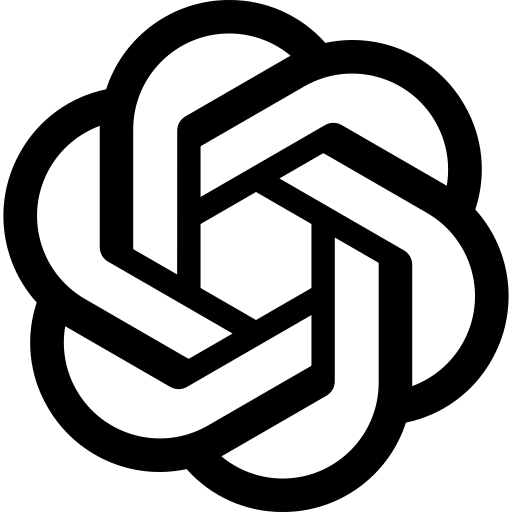}};
        \end{tikzpicture}
    \end{adjustbox}
    \captionsetup{justification=justified, singlelinecheck=false}
    \caption{Chain-of-Thought with few-shot \textcolor{gray}{prompting}.}
    \label{fig:chain-of-thought}
\end{figure}

\hypertarget{prompt-engineering}{}
\noindent \textbf{Eliciting reasoning via prompt engineering}\\[4pt]
LLMs are sensitive to prompting, so prompt engineering can elicit better reasoning from LLMs. The chain-of-thought (CoT) is one of the most well-known promptings~\citep{wei2022chain}. CoT decomposes a problem into a set of subproblems and then solves the sub-problems sequentially, where the answer to the next subproblem depends on the previous one. Consequently, CoT encourages LLMs to perform more intermediate reasoning steps before generating the final output. Figure \ref{fig:chain-of-thought} illustrates an example of CoT for better understanding. As a follow-up work, \citet{zhang2023building} input LLMs with additional communication information from other agents to generate high-level plans that involve multi-agent cooperation. In addition, Socratic models employ multimodal-informed prompting~\citep{zeng2022socratic}, such as utilizing visual language models and audio language models to incorporate perceptual information into the textual language input, thereby generating plans~\citep{liang2023code}. Socratic-model-based systems thus have access to open-ended reasoning, such as video Q\&A and forecasting, making these systems more robust to unseen objects. ECoT incorporates CoT into VLAs and performs multiple steps of reasoning about plans, sub-tasks, motions, and many others~\citep{zawalski2024roboticcontrolembodiedchainofthought, chen25training}.

\input{tikz/code_generation}

\hypertarget{code-generation}{}
\noindent \textbf{Code-generation}\\[4pt]
Beyond leveraging LLMs for planning, those trained on code completion have demonstrated the capability to synthesize policy code, orchestrating planning, policy logic, and control~\citep{chen2021evaluating,chowdhery2022palm,liang2023code}.
An example is shown in Figure \ref{fig:codegen}, where LLMs generate Pythonic code for an agent to perform a pick-and-place task. \citet{wang2023voyager} mention that programs can represent temporally extended and compositional actions. Otherwise, \citet{liang2023code} also argue that policy code generated by LLMs can run on the controller directly, avoiding the requirement of the language-conditioned plan to map every textual instruction into the executable action in the pre-trained skill library. \citet{liang2023code} utilize the prompting hierarchical code-gen approach to re-compose the original API calls, defining a more complex function flexibly, which can generalize better to unseen objects. 
Similar works, such as VOYAGER~\citep{wang2023voyager}, can code a new skill using skill retrieval and memorize it in the skill library, then refine learned skills to deal with unseen objects. However, LLMs for code generation struggle to interpret much longer and more complex commands than the sample and still may call functions that do not exist in the control primitive APIs. \citet{ha2023scaling} prompt LLMs to output a success function code snippet as a labeler. The LLMs can verify unlabeled trajectories through this success function and label them with success or failure. \citet{singh:progprompt:ar} design a Pythonic program prompt structure that ensures the generated plan has code formulation. This Pythonic plan inherits features from the code, such as obtaining state feedback via \textit{assert} and error collection via \textit{else}.

\textcolor{black}{However, these early Code-as-Policies methods~\citep{liang2023code,mu2023embodiedgpt,karli2024alchemist,mu2024robocodex} rely on open-loop control, meaning they cannot recover from execution errors or handle environmental uncertainties. To address this limitation, recent research incorporates closed-loop feedback and structured learning into code generation frameworks~\citep{meng2025data,meng2025growing}. For example, DAHLIA~\citep{meng2025data} introduces a dual-tunnel ``planner + reporter'' loop. The LLM first writes a multi-primitive code plan, guided by a CoT example curriculum that improves its reasoning. After robot execution, a vision-language reporter inspects before-and-after RGB-D frames and returns structured feedback (object, current pose, target pose, required action). The planner then patches only the failing parts and re-executes, iterating until success. While this LLM-based feedback improves robustness, it can be unreliable in extremely long-horizon tasks, as the LLM verifier may lack the understanding to identify subtle errors that are obvious to a human. Furthermore, this feedback is transient and does not contribute to the robot's long-term capability growth~\citep{meng2025growing}. The LYRA framework~\citep{meng2025growing} addresses this by integrating a human-in-the-loop for lifelong skill acquisition. It places a human verifier in the loop and stores every approved correction as a reusable skill function, indexed in an external vector memory. During future tasks, the agent retrieves the minimal skill set and few-shot plans, allowing the user to provide one-line hints that steer retrieval when ambiguity remains. This approach enables the robot to continually learn from its experiences and improve its performance over time.}

\textcolor{black}{In summary, early code-generation agents with LLMs~\citep{liang2023code,mu2023embodiedgpt,karli2024alchemist,mu2024robocodex} revealed the promise of ``from language to robot code'' for developing flexible and generalizable robot policies, but they were hindered by open-loop execution, shallow prompts, and forgetfulness. DAHLIA~\citep{meng2025data} addresses the first two issues with structured inter-loop feedback and CoT-guided example curricula, while LYRA~\citep{meng2025growing} complements it by turning corrections into persistent, retrievable skills under light human supervision. Together, these methods demonstrate a potential path for achieving reliable and scalable reasoning in language-conditioned manipulation. However, a fundamental limitation remains: the performance of these systems is ultimately capped by the reasoning and code-generation capabilities of the underlying LLM.  \citet{mialon2023augmented} demonstrate that different LLMs, and even different versions of the same model, possess varying inference capabilities, which directly impacts the quality of the generated code and, consequently, the robot's task success. For instance, on the GSM8K reasoning benchmark~\citep{cobbe2021training}, the code-optimized ``code-davinci-002'' model outperforms the text-optimized ``text-davinci-002'', highlighting that the choice of the foundation model is a critical factor for the success of code-as-policy approaches.}\\[-10pt]

\hypertarget{iterative-reasoning}{}
\noindent \textbf{Iterative reasoning}\\[4pt]
LLMs for planning or single-cycle reasoning often generate plans that are brittle to execution errors or environmental uncertainties, as they may call nonexistent functions or rely on flawed assumptions about the environment state. Iterative reasoning mitigates this issue by creating a closed loop where the model refines its output based on feedback from previous attempts iteratively. This process allows the agent to correct its course, circumventing hallucinations, and adapt to unexpected outcomes, as shown in Figure~\ref{fig:iterative}. Early examples of this paradigm focused on incorporating feedback. For instance, Inner Monologue~\citep{huang2022inner} uses feedback from a dedicated success detector and a scene descriptor to inform the LLM's next reasoning step, enhancing task completion. \textcolor{black}{A more cognitive form of iterative reasoning is \textit{self-reflection or self-correction}, where the LLM is prompted to analyze its own performance and generate corrective feedback for itself. This moves beyond simple error signals to a more cognitive process of introspection. For example,  REFLECT~\citep{reflect} attempts at this, where an LLM summarizes hierarchical sensor data into a textual form to obtain explanations for high-level task failures. However, it corrects errors only after an entire long-horizon task is completed, making it inefficient for real-time adjustments.} 

\input{tikz/iterative_plan}

\textcolor{black}{To address this, HiCRISP~\citep{ming2024hicrisp} introduces a hierarchical closed-loop system that enables error correction within individual steps of a task. It distinguishes between high-level planning failures and low-level action failures. When a low-level failure occurs, it first tries to correct using a predefined structure. If that fails, or if a high-level plan failure is detected, it generates a language prompt describing the error and the current state, which is fed back to the LLM to generate a corrected action or plan. While these methods correct high-level plans or skill choices, they often struggle to refine the low-level motor control parameters, like SE(3) pose for grasping. AIC MMLM~\citep{xiong2024autonomous} specifically targets this gap for articulated object manipulation. When a manipulation attempt fails, a Feedback Information Extraction module analyzes the failed interaction to infer its geometric cause. Similarly, the Self-Corrected (SC)-MLLM~\citep{liu2024self} also detects low-level pose errors and adaptively seeks feedback from ``experts'' (e.g., affordance models, grasp planners). Based on the expert's structured feedback (e.g., potential contact points or orientations), SC-MLLM re-evaluates the failure and generates a corrected action. Though these iterative reasoning methods enhance robustness, they still rely on the underlying LLM's reasoning capabilities and may require extensive computational resources or inference time due to multiple inference cycles. Moreover, this issue is exacerbated in long-horizon tasks, where multiple iterations may be needed at each step, leading to significant latency.} 

\hypertarget{llm-sp}{}
\subsubsection{\textcolor{black}{LLMs-driven  \textcolor{black}{structured} planning}}
\textcolor{black}{LLMs face challenges with long-horizon planning and reasoning, mainly due to their lack of interpretability and difficulty in handling complex constraints. To mitigate these issues, recent work has explored integrating LLMs with formal planning methods, such as symbolic systems and behavior trees, to leverage the strengths of both approaches.}

\hypertarget{llm-ss}{}
\noindent \textbf{\textcolor{black}{Combining LLMs with symbolic systems}}\\[4pt]
\textcolor{black}{Symbolic systems offer structured and interpretable planning but require extensive manual engineering. LLMs, on the other hand, posses vast commonsense knowledge but lack formal guarantees and can hallucinate. Combining them provides a promising solution for robust and grounded reasoning. One approach is to integrate LLMs with KGs, which provides a structured representation of entities and their relationships. For instance, PLANner~\citep{lu2022neuro} leverages an external KG (ConceptNet~\citep{speer2017conceptnet}) to construct commonsense-infused prompts, guiding the LLM to generate more plausible plans. Going a step further, LLMs can be used to autonomously build or augment these KGs. \citet{act13010028} use an LLM to create SkillKG, a task-centric manipulation KG, from which a robot can infer new task plans. Similarly, RoboEXP~\citep{jiang2024roboexp} interactively explores its environment to build an action-conditioned scene graph (a KG variant), which is then used for planning and reasoning. Another powerful symbolic formalism is the PDDL.}

\textcolor{black}{PDDL models define structured symbolic blueprints that enable off-the-shelf external planners (symbolic solvers) to generate robust and optimized solutions. Integrating LLMs with PDDL models enhances their capabilities by automating the difficult process of creating PDDL domain and problem files from natural language. Early works used LLMs to translate natural language goals into PDDL representations for constraint checking and goal verification~\citep{ding2022robot, xie2023translating}. However, this integration introduces several key challenges: (i) the slow speed of symbolic search, (ii) the symbol grounding problem, and (iii) the difficulty of representing complex real-world interactions. First, the \textbf{slow speed of symbolic search} may cause long delays before the robot can act. To overcome this, \citet{capitanelli2024framework} fine-tune GPT-3 on a domain-specific dataset of PDDL problem-plan pairs. This allowed the LLM to generate a plan action by action, enabling concurrent planning and execution and reducing the robot's perceived waiting time.} 

\textcolor{black}{The second challenge of \textbf{symbol grounding problem} is \textit{how to connect abstract PDDL predicates to the robot's physical reality}. To address this, IALP~\citep{wang2025instruction} develops a closed-loop system that augments user instructions with feasibility information derived from real-time sensor feedback. By using ``grounding mechanisms'' to check predicates, the PDDL problem itself is enriched with grounded, physically-aware knowledge before planning, ensuring the generated actions are viable. A specific but critical aspect of grounding arises in contact-rich manipulation, where visual perception alone is insufficient. LEMMo-Plan~\citep{chen2025lemmo} tackles this by incorporating multi-modal demonstrations, including tactile and force-torque data. This allows the LLM to reason about ``invisible'' events like cable tension or insertion forces, enabling it to define robust force-based skill conditions and segment demonstrations more accurately. Further refining the interface between language and symbolic systems, subsequent research explore different \textit{levels of abstraction}. Task-level PDDL may not be the most natural representation for an LLM to reason about. \citet{paulius2025bootstrapping} propose bootstrapping task planning with a higher Object-Level Plan. It uses an LLM to generate a plan schema describing object interactions, which is then systematically grounded into a set of PDDL subgoals for a traditional planner. This creates a cleaner separation of concerns, allowing the LLM to operate at a commonsense level while the symbolic planner handles the formal task-level constraints. The third challenge of \textbf{complex interaction representation} remains open, while behavior trees offer an alternative.}

\hypertarget{llm-bt}{}
\noindent \textbf{\textcolor{black}{Combining LLMs with behavior trees}}\\[4pt]
\textcolor{black}{While symbolic planners like PDDL may struggle with scalability, Behavior Trees (BTs) offer a more modular, reactive, and reusable alternative for specifying complex tasks~\citep{iovino2022survey}. However, manually designing BTs is a labor-intensive process that limits their scalability. To address this, recent work has focused on leveraging LLMs to automatically generate BTs from natural language instructions. This approach combines the structured, verifiable execution of BTs with the commonsense reasoning of LLMs, tackling key challenges in \emph{data inefficiency} (learn BTs from scratch) and \emph{runtime adaptability} (on-the-fly modifications). Early methods explored using LLMs for runtime plan adaptation. For example, LLM-bt~\citep{zhou2024llm} uses an LLM to generate high-level descriptive steps that are parsed into an initial BT. A dedicated algorithm then dynamically expands this tree at runtime if a failure is detected, allowing the robot to incorporate new actions to adapt to environmental changes. However, this approach still required hand-written parsing rules for each new task. To ensure optimality and correctness, LLM-OBTEA~\citep{chen2024integrating} introduces a two-stage framework where the LLM's role is limited to translating high-level instructions into a formal first-order logic goal. This goal is then fed to an Optimal BT Expansion Algorithm, which constructs a BT that is theoretically guaranteed to achieve the goal with minimal cost. A key limitation remained: the resulting BT could not autonomously debug missing preconditions discovered only during deployment. Subsequent work addressed this by casting the LLM as a ``common-sense repair agent''. BETR-XP-LLM~\citep{styrud2025automatic} uses the LLM to propose the minimal precondition and matching subtree required to resolve an execution failure, permanently integrating the fix into the policy. Because these patches are verifiable BT fragments, the robot's policy becomes more robust with every failure. Other research focuses on improving the initial generation process itself. \citet{ao2025llm} effectively exploite LLMs to directly generate BTs and explored four distinct in-context learning strategies for BT generation (one-step, iterative, human-in-the-loop, and recursive), enabling runtime feedback and replanning. They find that while richer interaction yields higher success rates, it also increases LLMs token and latency costs, highlighting a trade-off between performance and efficiency.}

\hypertarget{vlms}{}
\subsection{Empowered by vision-language models}
\label{sec:empowered-by-vision-language-models}

\textcolor{black}{In robotic manipulations, while LLMs excel at high-level reasoning and planning, they suffer from a fundamental limitation: they are disembodied. Operating purely on text, they lack direct perceptual access to the robot's environment. This leads to the classic ``grounding problem'', where the model's plans might involve objects that are not present or fail to account for the current observation of the world, resulting in hallucinations and infeasible actions. To bridge this gap, many approaches require intermediate modules, such as object detectors (like YOLO~\citep{redmon2016you}, MDETR~\citep{kamath2021mdetr}), to translate visual information into text that the LLM can comprehend. This process may lose critical information. VLMs offer a more direct and powerful solution. By being pre-trained on vast datasets of paired images and text, VLMs can inherently process and reason about visual information. This allows them to directly ground high-level language instructions in the images, connecting abstract concepts like ``the red block'' to specific image pixels. This section explores how the tight fusion of vision and language in VLMs enables more robust and capable robotic manipulation systems, focusing on contrastive learning and generative methods.}

\hypertarget{contrastive-learning}{}
\subsubsection{Contrastive learning approaches}
\textcolor{black}{Contrastive learning is widely used in VLMs to align the text and vision modalities. It learns representations where similar samples are brought closer together in the learned latent space, while dissimilar samples are pushed farther apart.}
The well-known approach CLIP \citep{radford2021learning} aligns the image embedding and the text embedding of its corresponding caption in the latent space. 
In the field of robot manipulation, CLIP-based approaches are extensively applied. 
CLIPORT~\citep{shridhar2022cliport}, a method that combines the broad semantic understanding of CLIP with the spatial precision of Transporter~\citep{zeng2021transporter}, is capable of solving a variety of language-specified tabletop tasks. In Dream2Real~\citep{dream2real}, 6-DoF language-based rearrangement tasks are achieved by enabling robots to ``imagine" virtual goal states and then evaluate them using CLIP. CLIP-Fields~\citep{shafiullah2023clipfields} is capable of mapping spatial locations to semantic embedding vectors trained using CLIP-based approaches, enabling the agent to conduct navigation and localization. 

EmbCLIP \citep{khandelwal2022:embodied-clip} investigates the effectiveness of CLIP visual backbones for Embodied AI tasks. 
Instruction2Act \citep{huang2023instruct2act} leverages CLIP model to accurately locate and classify objects in the environments. 
LAMP \citep{adeniji2023language} leverages R3M~\citep{nair2022r3m}, a reusable representation for robot manipulation, to calculate the rewards for RL.
MOO \citep{stone2023open} queries an OWL-ViT \citep{minderer2022simple} to produce a bounding box of the object of interest with the prompt ``An image of an X''. \citet{xiao2022robotic} enhance instructions using CLIP by fine-tuning it with robot manipulation data and natural language annotations, and then label a larger dataset with this CLIP model for further training.
LATTE utilizes CLIP encoders to better align visual and textual information~\citep{bucker2023latte}, grounding the instructions in the specific objects that need manipulation. Moving beyond image-level approaches, R+X~\citep{papagiannis2024rplusx} demonstrates how Gemini can be utilized to retrieve relevant videos of human demonstrations for a given language-based task, which is then used to condition a policy.
\\
\hypertarget{generative-approaches}{}
\subsubsection{Generative approaches}
\textcolor{black}{Generative approaches model data distributions to synthesize textual and visual content, conditioned on text, images, or both simultaneously. They are important for language-conditioned robot manipulation, since the generated content can be used for planning, reasoning, data augmentation, and future state estimation.}

\noindent \textbf{\textcolor{black}{Text generation}}\\[4pt]
\noindent
Auto-regressive text generation approaches~\citep{cho2021unifying, wang2022ofa} typically merge visual and textual information through a Transformer-based method and perform a sequence-to-sequence structure to generate text.  
A growing line of work uses such VLMs as planning and evaluation scaffolds. For example, \citet{patel2023pretrained} demonstrate that a pre-trained planner can convert textual task descriptions and execution videos into explicit plans. Complementing planning, language-augmented evaluators provide reward signals: \citet{du2023visionlanguage} fine-tune Flamingo~\citep{alayrac2022flamingo} as a success detector via VQA, using its judgments for reward design.
Beyond planning and evaluation, VLMs supply promptable perceptual priors and enable data-centric scaling. For instance, PR2L~\citep{chen2024visionlanguagemodelsprovidepromptable} demonstrates that VLMs can generate superior visual embeddings when prompted with task-relevant context, outperforming generic, non-promptable representations. Similarly, RoboPoint~\citep{yuan2024robopoint} addresses the VLM's struggle with precise action articulation by using a synthetic data pipeline to instruction-tune a VLM to predict keypoint affordances from language instructions, grounding abstract commands in concrete visual features without requiring real-world data. In parallel, NILS~\citep{blank2024scaling} leverages VLMs for data-centric scaling by automatically generating descriptive labels for unlabeled robot datasets, which are then used to train more capable language-conditioned policies.

\textcolor{black}{Some works focus on leveraging the generalist reasoning capabilities of large-scale VLMs. PaLM-E~\citep{driess2023palme}, a 562B parameter VLM, demonstrates that a single massive model could perform embodied reasoning, translating high-level commands into robot-executable plans \textcolor{black}{with little or no robot-task-specific finetuning}. This shows that scaling could directly yield more powerful robotic planners. An alternative to a single model is a modular approach. Socratic Models~\citep{zeng2022socratic} proposes a framework where multiple specialized models (e.g., for vision, audio, and language) communicate through multimodal prompts, collaboratively solving problems that no single model could handle alone. This enables more flexible and extensible reasoning. PIVOT~\citep{nasiriany2024pivotiterativevisualprompting} reframes manipulation as an iterative visual dialogue, where the VLM repeatedly answers questions about the scene to refine its proposed actions. This turns the VLM into an active reasoning partner that grounds its decisions through a continuous feedback loop.}

\noindent \textbf{\textcolor{black}{Image generation}}\\[4pt]
\textcolor{black}{The ability of generative models to generate realistic images or videos over a long horizon brings new opportunities for robotic manipulation. Such models commonly include text-to-image models like Make-A-Video~\citep{singer2023makeavideo}, DALL-E 2~\citep{ramesh2022hierarchical}, Stable Diffusion~\citep{rombach2022high}, and Imagen~\citep{saharia2022photorealistic}. Instead of just aligning or describing existing content, these models can synthesize novel visual data from language, serving several key roles in robotics.} One primary application is \emph{goal visualization}, where a text-to-image model translates a language instruction into a concrete goal image. This visual goal can then guide a low-level, goal-conditioned policy. For example, Dall-E-Bot~\citep{kapelyukh2023dall} utilizes a text-conditioned diffusion model to generate goal images for tabletop object rearrangement tasks. SuSIE~\citep{black2023zero} leverages pre-trained text-conditional image-editing models to generate a sequence of subgoals, which are then executed by a low-level controller. This approach decouples high-level semantic understanding from low-level control, allowing each component to be optimized independently. 

Another powerful use is \emph{data augmentation}. Given the high cost of collecting real-world robot data, generative models can create diverse, synthetic training examples. \citet{chen2024semantically} take a small offline dataset of expert demonstrations and use a text-to-image model to semantically bootstrap it into a much larger and more varied dataset. This augmented data can then be used to train a robot policy that generalizes better to unseen environments and tasks. Generative models can also function as \emph{world models} by predicting future video frames conditioned on language and current actions. These video prediction models can be used for planning or to learn an inverse dynamics model that maps desired outcomes back to the actions required to achieve them~\citep{gu2023seer,ko2023learning, du2024learning}. This allows the robot to ``imagine'' the consequences of its actions before executing them~\citep{nematoli20iros}. \textcolor{black}{Moreover, generative models can enhance policies through \emph{multi-modal conditioning}, especially when dealing with partially annotated datasets. GR-MG~\citep{li2025gr} first trains a policy that accepts both language and image goals. During inference, when only a text instruction is available, it uses an image-editing model~\citep{brooks2023instructpix2pix} to generate a corresponding goal image. The policy is then conditioned on both the original text and the generated image, making it more robust.} These approaches are powerful but resource-intensive. The generated images and videos can contain noise, artifacts, or physically implausible details, making it challenging to extract reliable signals for robot control~\citep{yuan2024general,mccarthy2024towards}.

\hypertarget{vlams}{}
\section{\textcolor{black}{Language in unified vision-language-action models}}
\label{sec:vla}
\begin{figure*}[ht!]
    \vspace{-1em}
    \centering
    \resizebox{.9\textwidth}{!}{
    \begin{tikzpicture}[mindmap]
    \selectcolormodel{cmyk}
        \node[circle, draw=RoyalBlue!50, fill=RoyalBlue!50, text=white, minimum size=2cm, text width=3cm, align=center] (n0)at(0,0){\textbf{\textsf{Vision-Language-Action Models}}};
        
        \node[circle, draw, fill, Lavender!50, text=white, minimum size=1.6cm, text width=2.1cm, align=center] (l150)   
        at ( -3.5cm,  2.5cm) 
        {\small \textbf{\textsf{I. Multimodal Sensing \& Fusion}}};
        \node[circle, draw, fill, Thistle!50, text=white, minimum size=1.6cm, text width=2.1cm, align=center] (l170)   
        at (-6.75cm,  1.5cm) 
        {\small \textbf{\textsf{I. Spatial understanding}}};        
        \node[circle, draw, fill, Orchid!50, text=white, minimum size=1.6cm, text width=2.1cm, align=center] (lm170) 
        at (-9cm,  -.75cm) 
        {\small \textbf{\textsf{I. Data Sources \& Augmentation}}};
        \path (n0) to[circle connection bar switch color=from (RoyalBlue!50) to (Orchid!50)] (lm170);
        \path (n0) to[circle connection bar switch color=from (RoyalBlue!50) to (Thistle!50)] (l170);
        \path (n0) to[circle connection bar switch color=from (RoyalBlue!50) to (Lavender!50)] (l150);

        \path (lm170) to[circle connection bar switch color=from (Orchid!50) to (Thistle!50)] (l170);
        \path (l170) to[circle connection bar switch color=from (Thistle!50) to (Lavender!50)] (l150);

        \draw[decorate, decoration={text along path,
            text={|\color{Orchid!80}\bfseries\sffamily\Large|Group I: Perception},
            text align={align=center}}] 
            (-10.5,-1) .. controls (-8.5,3.5) and (-6.5,4) .. (-3.5,3);
        
        \node[circle, draw, fill, Dandelion!50, text=white, minimum size=1.6cm, text width=2.1cm, align=center] (r30)  
        at (3.5cm, 2.5cm)
        {\small\textbf{\textsf{II. Long-horizon task solving}}};
        \node[circle, draw, fill, YellowOrange!50, text=white, minimum size=1.6cm, text width=2.1cm, align=center] (r10)  
        at ( 6.75cm,  1.5cm) 
        {\small\textbf{\textsf{II. Knowledge preserving}}};
        \node[circle, draw, fill, Orange!50, text=white, minimum size=1.6cm, text width=2.1cm, align=center] (rm10) 
        at ( 9cm,  -.75cm) 
        {\small\textbf{\textsf{II. Reasoning \& World Models}}};
        \path (n0) to[circle connection bar switch color=from (RoyalBlue!50) to (Dandelion!50)] (r30);
        \path (n0) to[circle connection bar switch color=from (RoyalBlue!50) to (YellowOrange!50)] (r10);
        \path (n0) to[circle connection bar switch color=from (RoyalBlue!50) to (Orange!50)] (rm10);

        \path (rm10) to[circle connection bar switch color=from (Orange!50) to (YellowOrange!50)] (r10);
        \path (r10) to[circle connection bar switch color=from (YellowOrange!50) to (Dandelion!50)] (r30);
        
        \draw[decorate, decoration={text along path,
            text={|\color{Orange!80}\bfseries\sffamily\Large|Group II: Reasoning},
            text align={align=center}}] 
            (3.5,3) .. controls (6.5,4) and (8.5,3.5) .. (10.5,-1);
        
        \node[circle, draw, fill, Cerulean!50, text=white, minimum size=1.6cm, text width=2.1cm, align=center] (rm30) 
        at (3cm, -3cm)
        {\textbf{\small\textsf{III. Policy execution}}};
        \path (n0) to[circle connection bar switch color=from (RoyalBlue!50) to (Cerulean!50)] (rm30);
        
        \node[circle, draw, fill, JungleGreen!50, text=white, minimum size=1.6cm, text width=2.1cm, align=center] (lm150) 
        at (-3cm, -3cm)
        {\small \textbf{\textsf{IV. Adaptation}}};
        \path (n0) to[circle connection bar switch color=from (RoyalBlue!50) to (JungleGreen!50)] (lm150);

    \end{tikzpicture}
    }
    \caption{\textcolor{black}{Hierachical taxonomy of Vision-Language-Action models according to optimization directions, which fllows Perception $\rightarrow$ Reasoning $\rightarrow$ Action $\rightarrow$ Adaptation: (I) Perception and state representation. (II) Reasoning, planning, and knowledge preserving. (III) Action generation and execution. (IV) Model learning and adaptation.}}
    \label{fig:vla-taxonomy}
    \vspace{-1em}
\end{figure*}

As the field has evolved, increasing attention has shifted from hierarchical language-conditioned systems to end-to-end vision-language-action models (VLAs). In the hierarchical approaches discussed in Section \ref{sec:language-cognitive-planning-reasoning}, large models such as LLMs and VLMs typically serve as planners, reasoners, or perceptual modules, while action execution is handled by separate low-level controllers, policy modules, or motion planners. By contrast, VLAs aim to learn a policy model that jointly represents visual observations, language instructions, and robot actions within \textcolor{black}{a unified embodied policy architecture}. In this paradigm, language is no longer limited to high-level reasoning or external task specification. Instead, \textcolor{black}{it is coupled with visual perception and action in the model's core representation and training process}.

\textcolor{black}{
The term VLA, however, is not used uniformly across the literature. Under a broad input-output definition, a system may be called a VLA if it receives visual observations and language instructions and produces robot actions~\citep{11164279}. This broad view is useful for tracing the historical development of observation-language-action systems. In this survey, however, the VLA category is used in a narrower analytical sense to preserve the distinction between the functional roles of language introduced in Sections~\ref{sec:language-as-policy-condition}--\ref{sec:language-cognitive-planning-reasoning}. Specifically, this section focuses on embodied policy models in which action generation is learned in close coupling with visual-linguistic representations through the model architecture and training objective. This coupling may be realized through discrete action tokens that are detokenized into executable robot commands~\citep{brohan2022rt,brohan2023rt}, continuous action heads, diffusion-based action generation~\citep{wen2025diffusionvla}, flow-matching-based controllers~\citep{black2024pi_0}, or action-expert modules attached to a VLM backbone.
}

\textcolor{black}{
This distinction is particularly relevant to the boundary between Section~\ref{sec:language-as-policy-condition} and the present section. Both language-conditioned policies and VLAs may receive visual observations and language instructions as inputs and output robot actions. The difference lies in the role played by language and in how the action-generation mechanism is integrated with the vision-language model. In language-conditioned policies, language primarily acts as an external task or behavior condition for a policy. In VLM-empowered methods, a pretrained VLM may provide semantic grounding, perception, affordance estimation, or reasoning, while a task-specific policy or controller generates executable actions. In contrast, the VLA models emphasized in this section integrate action generation more directly into the embodied vision-language policy, so that robot actions are learned as a central component of the model rather than only as the output of a downstream task-specific controller.
}

\textcolor{black}{
CLIPORT~\citep{shridhar2022cliport} provides an example. From a broad observation-language-action perspective, CLIPORT can be viewed as an early VLA-like system, as it integrates CLIP-based vision-language grounding with a manipulation policy that outputs robot actions~\citep{11164279}. In our taxonomy, however, CLIPORT is discussed primarily as a VLM-empowered language-conditioned visuomotor policy. The reason is that CLIP~\citep{radford2021learning} serves as a frozen semantic grounding module, while action generation is performed by a Transporter-style policy~\citep{zeng2021transporter}. Specifically, CLIPORT predicts dense pick and place affordance maps and selects executable actions through spatial maximization,
followed by a placement prediction conditioned on the selected pick location. Thus, language in CLIPORT conditions the policy and improves semantic grounding, but the robot action is generated through a task-specific pick-and-place affordance mechanism, rather than through an action representation integrated into a large VLM-based embodied policy. This placement is not meant to deny CLIPORT's observation-language-action attribute~\citep{11164279}, but to maintain a finer-grained distinction between VLM-empowered language-conditioned policies and recent unified VLA policy models in our taxonomy.}

\subsection{Overview}
Gato~\citep{reed2022a} represents a pioneering effort in this direction as a general-purpose agent. The authors tokenize text, images, discrete values, and continuous values, then train the model from scratch by predicting the masked tokens within sequences. RT-1~\citep{brohan2022rt}, RT-2~\citep{brohan2023rt}, Gemini Robotics~\citep{team2025gemini,abdolmaleki2025gemini}, and PI VLAs~\citep{black2024pi_0,pmlr-v305-black25a} utilize pre-trained vision-language models, which are co-fine-tuned on both large-scale vision-language datasets and low-level robot actions, to enhance generalization to novel objects and commands. RT-X~\citep{open_x_embodiment_rt_x_2023} is trained on an assembled dataset collected from 22 different robots, encompassing 527 skills across 160,266 tasks. It demonstrates positive transfer effects, where shared experiences from other platforms enhanced the capabilities of individual robots, suggesting that VLAs could benefit from cross-embodiment learning. In contrast, OpenVLA~\citep{kim2024openvlaopensourcevisionlanguageactionmodel} utilizes the pre-trained open-source Llama 2 (7B) along with features from DINOv2~\citep{Oquab2023DINOv2LR} and SigLIP~\citep{Zhai_2023_ICCV}, and is fine-tuned on a robot manipulation dataset, demonstrating that data diversity and new model components allow a small-scale model (7B) outperform the large-scale model RT-X (55B) in some tasks. 
Instead of relying on 2D inputs, 3D-VLA~\citep{zhen20243dvla3dvisionlanguageactiongenerative} is built on a 3D-based LLM and introduces a set of interaction tokens to engage with the embodied environment. 

As the field of VLAs continues to rapidly evolve with the advancement of foundation models, we present a systematic taxonomy to organize recent developments. Our taxonomy follows the flow: Perception $\rightarrow$ Reasoning $\rightarrow$ Action $\rightarrow$ Adaptation\footnote{In this taxonomy, we focus on delineating the primary contribution of each method (like reasoning aspect), even though most methods involve multiple aspects (such as perception, action).}, as illustrated in Figure \ref{fig:vla-taxonomy}:
\begin{itemize}
    \item \textcolor{black}{\textbf{Perception}: Methods that focus on optimizing how VLAs perceive and understand their environment, including: \emph{Data Sources and Augmentation}, \emph{3D Scene Representation and Grounding}, and \emph{Multimodal Sensing and Fusion}.}
    \item \textcolor{black}{\textbf{Reasoning}: Approaches that enhance the model's internal ``thought process'', that is, how it forms plans, leverages prior knowledge, and predicts outcomes to solve complex tasks, including: \emph{Internal World Models and Reasoning}, \emph{Preserving Foundational VLM Capabilities}, and \emph{Long-Horizon Planning}.}
    \item \textcolor{black}{\textbf{Action}: Methods that focus on the optimization regarding the ``output'' stage of the policy, concerning the form and mechanism of the robot's actions. This bridges the gap between the model's internal plan and its physical embodiment, primarily including: \emph{Action Generation and Execution}.}
    \item \textcolor{black}{\textbf{Adaptation}: Techniques for efficiently training, fine-tuning VLA models to enhance their adaptability to new situations and downstream tasks, including: \emph{Efficient Adaptation and Fine-Tuning}.}
\end{itemize}

\textcolor{black}{To synthesize the rapid progress in VLAs, Table \ref{tab:vla} summarizes the comparison of representative state-of-the-art methods across these four dimensions.}

\begin{table*}[htbp]
    \centering
    \caption{\textcolor{black}{Comparison of representative state-of-the-art methods for \textbf{Section \ref{sec:vla} Vision-language-action models}. This table categorizes large model-driven end-to-end policies based on their primary optimization directions: Perception, Reasoning, Action, and Learning \& Adapting. By tokenizing reasoning and low-level actions within a unified framework, these representative methods illustrate the cutting-edge strategies for overcoming embodiment gaps, enhancing internal planning, physical execution, and achieving efficient fine-tuning.}}
    \label{tab:vla}
    \small
    \begin{tabular}{p{0.12\linewidth}| p{0.1\linewidth}| p{0.21\linewidth}| p{0.21\linewidth}| p{0.21\linewidth}}
        \toprule[1pt]
        \textbf{Method} & \textbf{Optimization Direction} & \textbf{Core Mechanism} & \textbf{Key Advantages} & \textbf{Key Disadvantages} \\
        \midrule[0.2pt]
        EgoVLA\qquad\citep{yang2025egovla} & Perception & Pre-trains on egocentric human videos to predict actions, then fine-tunes with robot data. & Dramatically reduces reliance on large-scale, costly robot data collection for learning complex skills. & Inherently requires robust action retargeting methods to bridge the ``embodiment gap'' between humans and robots. \\
        \midrule[0.2pt]
        BridgeVLA \qquad\citep{li2025bridgevla} & Perception & Projects 3D point clouds into multi-view 2D images and predicts 2D heatmaps to seamlessly align 3D inputs and actions within a pre-trained 2D VLM. & Highly sample-efficient (3 trajectories per task) and generalizes robustly to visual disturbances and novel instructions. & Target keypoints can be occluded during the 2D orthographic projection, and performance drops on complex, multi-step long-horizon tasks. \\
        \midrule[0.2pt]
        CoT-VLA\qquad\citep{zhao2025cot} & Reasoning & Autoregressively generates intermediate subgoal images as a visual ``chain-of-thought'' before predicting the final actions. & Improves multi-step reasoning and complex instruction following by explicitly visualizing the goal state first (reason before acting). & High computational overhead due to image token generation, and struggles with out-of-distribution subgoal synthesis. \\
        \midrule[0.2pt]
        MemoryVLA\qquad\citep{shi2025memoryvla} & Reasoning & Incorporates explicit memory mechanisms to retain task context and visual history over extended interactions. & Mitigates context-loss in long-horizon tasks, allowing the robot to execute sequences that span extended timeframes. & Managing and retrieving from an expanding memory buffer adds architectural complexity and computational cost. \\
        \midrule[0.2pt]
        $\pi_0$\qquad\citep{black2024pi_0} & Action & Adds an action expert using flow-matching to a pre-trained VLM, outputting high-frequency, continuous actions. & Bridges semantic reasoning with precise continuous motor control, enabling highly dexterous, multi-stage tasks. & Requires massive and diverse training datasets (~10,000 hours). \\
        \midrule[0.2pt]
        ControlVLA\qquad\citep{li2025controlvla} &  Adaptationg & Integrates object-centric representations into VLA models, using zero-initialized layers to adapt without losing prior knowledge. & Highly data-efficient (requires only 10-20 demonstrations) and robust to unseen objects and backgrounds. & Relies on external visual models (GroundingDINO, SAM2) for tracking, and evaluations are currently limited to single-arm indoor manipulation.\\
        \midrule[0.2pt]
        OpenVLA-OFT\qquad\citep{kim2025fine} &  Adaptationg & Replaces autoregressive discrete prediction with an Optimized Fine-Tuning (OFT) recipe, integrating parallel decoding, action chunking, and continuous L1 regression. & Improves inference speed (25-50×) and reduces latency for real-time control, while boosting task success rates. & Struggles with complex, memory-dependent tasks and may lack the capacity to model highly multimodal action distributions. \\
        \bottomrule[1pt]
    \end{tabular}
\end{table*}

\subsection{Optimization for perception}
\textcolor{black}{A VLA model's ability to act intelligently is fundamentally rooted in its capacity to perceive and interpret the world accurately. The methods in this section focus on optimizing the ``input'' stage of the policy, addressing how VLAs can be trained on richer, more diverse, and more robust data sources to build a better understanding of their environment.}

\subsubsection{\textcolor{black}{Data sources and augmentation}}
\textcolor{black}{A foundational challenge in training a generalist VLA has been the scarcity and high cost of collecting large-scale, high-quality robot demonstration data. The requirement for physical robots and expert teleoperators fundamentally limits the scale, diversity, and complexity of tasks that can be captured, creating a data bottleneck for training powerful generalist policies. To overcome this challenge, a significant line of research has explored alternative data sources, moving beyond robot-specific demonstrations to leverage the vast and rich data generated by human activities. For example, EgoVLA \citep{yang2025egovla} proposed a paradigm of learning from egocentric human videos in a two-stage process. First, a VLA model is pre-trained on a curated dataset of egocentric human videos to predict human wrist and hand actions. Then, this model is fine-tuned with a small number of robot-specific demonstrations to bridge the ``embodiment gap'' between humans and robots, using inverse kinematics and action retargeting methods. This approach dramatically reduces the reliance on large-scale robot data collection for learning complex skills. H-RDT \citep{bi2025h} seeks to scale up this two-stage paradigm and addresses the challenges of transferring knowledge from a single embodiment (human) to multiple robot embodiments. It first pre-trains a 2B-parameter diffusion transformer on a human manipulation dataset, which provides a more consistent and powerful behavioral prior. Then, it applies modular action encoders/decoders that are re-initialized and fine-tuned for each target robot, preserving the rich manipulation knowledge learned during pre-training. This design turns rich but embodiment-mismatched human motion into a transferable visual-linguistic prior, yielding obvious improvements over training-from-scratch models ($\pi_{0}$~\citep{black2024pi_0}, RDT~\citep{liu2025rdt1bdiffusionfoundationmodel}).} 

\textcolor{black}{However, simply increasing the volume of data does not guarantee generalization. A critical line of inquiry investigates why policies trained on large, aggregated datasets often exhibit ``shortcut learning'', relying on spurious correlations instead of causal features, which leads to poor out-of-distribution performance. \citet{xing2025shortcut} provide a comprehensive analysis of this problem, identifying two primary causes: (1) limited diversity within individual sub-datasets and (2) dataset fragmentation, where significant distributional gaps exist between different sub-datasets (e.g., data from different labs or robots). The model learns to associate task-irrelevant features (like camera viewpoint or background) with specific tasks. Because those features are predictive within a fragmented dataset. To mitigate this, the work proposes not only principles for better data collection but also a practical solution for existing offline datasets: robotic data augmentation. By synthetically generating novel viewpoints or swapping objects between scenes, these augmentation strategies can increase intra-dataset diversity and bridge the gaps between fragmented sub-datasets, effectively breaking the spurious correlations that lead to shortcut learning. }

\subsubsection{\textcolor{black}{3D scene representation and grounding}}
\textcolor{black}{Many VLAs utilize 2D vision as their primary input modality, which lacks 3D spatial awareness (e.g., depth, scale, and spatial relationships). This ``spatial awareness'' gap hinders a robot's ability to perform precise manipulation. To overcome the limitations of 2D vision, research has focused on integrating 3D information into VLA frameworks, grounding the model's understanding in the geometric reality of the physical world. A direct approach is to augment 2D features with explicit spatial data. For instance, SpatialVLA~\citep{qu2025spatialvla} introduces Ego3D Position Encoding, which utilizes a depth model to calculate the 3D coordinates of each pixel and incorporates this information into the VLM's 2D visual features. This method enriches the model's input with spatial context without requiring complex architectural changes and extends this spatial reasoning to the action space via Adaptive Action Grids, making action tokens spatially meaningful. However, directly modifying the visual input stream risks disrupting the VLM's pre-trained vision-language alignment. To mitigate this, PointVLA~\citep{li2025pointvla} proposes a modular framework that injects 3D information with minimal disruption. It uses a separate lightweight encoder to process a 3D point cloud and injects the resulting geometric features into a frozen action expert module, thereby preserving the core VLM's integrity.}

\textcolor{black}{An alternative to injecting 3D data is to reformat it into a 2D representation that the VLM can natively understand. BridgeVLA~\citep{li2025bridgevla} implements this by projecting a 3D point cloud into multiple 2D orthographic images (e.g., top, front, and side views). These 2D images are then fed into the standard VLM backbone, and the model predicts actions as 2D heatmaps over these same projections. This approach unifies the input and output within a consistent 2D space, leveraging the VLM's powerful 2D understanding while grounding actions in a 3D-derived representation. After exploring methods that either enhance features with 3D data~\citep{qu2025spatialvla}, inject~\citep{li2025pointvla}, or reformat~\citep{li2025bridgevla}, a more holistic approach emerged that seeks to process both 2D and 3D modalities in parallel and fuse them at the decision-making stage. GeoVLA~\citep{sun2025geovla} processes 2D and 3D modalities in parallel. a standard VLM processes the 2D image and language to produce a fused vision-language embedding, while a separate network processes the point cloud to generate a separate 3D geometric embedding. The resulting embeddings are then concatenated and fed into a novel 3D-enhanced Action Expert, which uses a Mixture-of-Experts architecture ~\citep{jacobs1991adaptive, mees16iros} to fuse the information before generating the final action. This parallel design preserves the integrity of the pre-trained VLM while enabling a more expressive combination of 2D semantic and 3D geometric cues.} 

\textcolor{black}{The integration of 3D scene representation is a critical and rapidly evolving frontier for VLA models. A primary insight is that simply adding a 3D sensor is not enough, the core scientific challenge lies in bridging the ``modality gap'' between the 2D-centric perception of existing VLMs and the 3D geometry of the real world~\citep{mees19iros,kerr2023lerf}. The field is moving from straightforward feature augmentation~\citep{qu2025spatialvla}  toward more sophisticated and principled architectures that are modular~\citep{li2025pointvla}, aligned~\citep{li2025bridgevla}, or parallelized~\citep{sun2025geovla}. These advanced strategies aim to harness the semantic richness of 2D vision and the geometric precision of 3D data simultaneously, paving the way for robot policies with superior spatial awareness, precision, and adaptability to real-world complexities.}

\subsubsection{\textcolor{black}{Multimodal sensing and fusion}}
\textcolor{black}{While vision provides rich global context, it is insufficient for contact-rich tasks like assembly or insertion, where a sense of touch is crucial. To overcome this limitation, a key area of research focuses on integrating non-visual sensory data, such as tactile and force feedback, into VLA models. This introduces challenges in representing diverse sensory signals, fusing them with vision-language features, and grounding abstract commands in physical forces.} \textcolor{black}{Early work focused on integrating tactile data and refining the learning process. VTLA~\citep{zhang2025vtla} is proposed to fuse vision, tactile, and language by tokenizing tactile image sequences alongside other modalities. It addresses the mismatch between the VLM's classification objective and continuous robot control by using a two-stage training process with Direct Preference Optimization (DPO) to refine the policy. While integrating tactile data as another perceptual channel is a crucial first step, a deeper challenge lies in grounding the VLM's abstract knowledge directly into physical force control. To better ground language in physical interaction, Tactile-VLA~\citep{huang2025tactile} augments the policy's output to predict both target force and target position, enabling it to understand force-related adverbs like ``gently'' or ``hard''. It also introduces a CoT mechanism to interpret tactile feedback upon failure and generate corrective actions.}

\textcolor{black}{A practical barrier is the heterogeneity of tactile sensors, where some are vision-based (e.g., GelSight), while others are force-based, each with different data structures and characteristics. To address this, OmniVTLA~\citep{cheng2025omnivtla} creates a unified tactile representation by using a dual-path encoder for different sensor types ( vision-based tactile sensors and force-based sensors), and uses cross-modal contrastive learning to align the tactile representations with their linguistic and visual counterparts. This pre-alignment ensures that tactile information is processed within a shared semantic space before being fused into the main VLA model. As models begin incorporating multiple modalities, the focus shifts to how to fuse them most effectively. Simple concatenation or early fusion can disrupt the VLM's powerful pre-trained representations. ForceVLA~\citep{yu2025forcevla} introduces a dynamic ``late fusion'' mechanism using a force-aware Mixture-of-Experts (MoE) module. This allows the model to adaptively route sensory inputs to specialized expert networks, enabling a more effective integration of real-time force feedback.}

\textcolor{black}{In summary, incorporating a ``sense of touch'' is essential for enabling robust manipulation in contact-rich environments. Research has progressed from simple data fusion to more principled approaches that address learning objectives (VTLA~\citep{zhang2025vtla}), semantic grounding (Tactile-VLA~\citep{huang2025tactile}), sensor heterogeneity (OmniVTLA~\citep{cheng2025omnivtla}, FuSe-VLA~\cite{jones24fuse}), and dynamic fusion (ForceVLA~\citep{yu2025forcevla}). This progression underscores that the future of generalist robots lies not just in seeing and hearing, but in feeling and physically reasoning about their physical interactions with the world. This requires architectures can intelligently and adaptively fuse multiple sensing modalities.}

\subsection{Optimization for reasoning}
Approaches that enhance the model's internal ``thought process'', that is, how it forms plans, leverages prior knowledge, and predicts outcomes to solve complex tasks.

\subsubsection{\textcolor{black}{Long-horizon planning}}
\textcolor{black}{While VLAs excel at short-horizon skills, they often struggle with long-horizon tasks due to their reactive, step-by-step characteristics, which leads to compounding errors over time. To address this, a key area of research has focused on endowing VLAs with planning capabilities, enabling them to reason about multi-step sequences. This has led to three primary strategies: hierarchical decomposition, phase-aware control, and memory-augmented reasoning. The first strategy, \textbf{hierarchical decomposition}, trains the VLA to function as its own high-level planner. LoHoVLA~\citep{yang2025lohovla} and DexVLA~\citep{wen2025dexvla} explore this by training a unified model to first generate a linguistic sub-task (e.g., ``\textit{pick up the green block}'') and then use that self-generated instruction to predict the corresponding action. This ``think before you act'' process is enhanced in LoHoVLA~\citep{yang2025lohovla} with a closed-loop control mechanism that efficiently replans at the sub-task level when execution fails, making error correction more robust. While DexVLA~\citep{wen2025dexvla} trains the model on demonstrations annotated with fine-grained sub-step reasoning, compelling the VLM backbone to function as an implicit high-level policy that decomposes direct prompts into a sequence of executable steps.}

\textcolor{black}{While hierarchical decomposition provides a plan, it does not solve the \textbf{skill chaining challenge}, where dynamic coupling and error propagation between sub-tasks degrade performance. To address this, Long-VLA~\citep{fan2025long} introduces a more fine-grained and \textbf{phase-aware control} strategy. It segments each sub-task into a ``moving phase'' and an ``interaction phase'' and uses a novel phase-aware input masking to force the model to attend to the most relevant camera view for each phase (e.g., a wide view for navigation, a gripper view for manipulation). This allows a single model to smoothly transition between different stages of a sub-task, improving robustness. An alternative to explicit planning is to use \textbf{memory-augmented reasoning} to handle the non-Markovian properties of long-horizon tasks, where action history is crucial. Mainstream VLAs are effectively ``memoryless'', which can cause failures when an action produces little visual change. MemoryVLA~\citep{shi2025memoryvla} solves this by incorporating an explicit memory system (Perceptual-Cognitive Memory Bank) that stores both low-level perceptual details and high-level semantic summaries from past steps. By retrieving and fusing relevant memories with the current observation, the model obtains a continuous stream of historical context, allowing it to handle temporal dependencies without relying on a predefined task hierarchy.}

\textcolor{black}{In summary, long-horizon planning approaches range from integrating explicit hierarchical reasoning within unified models to developing complex mechanisms for temporal consistency. LoHoVLA~\citep{yang2025lohovla} and DexVLA~\citep{wen2025dexvla} represent a ``planning-as-reasoning'' paradigm, where the VLM is trained to generate its own textual sub-goals to guide behavior. While Long-VLA~\citep{fan2025long} focuses on the execution dynamics within and between these sub-goals, proposing a phase-based perceptual adaptation to improve skill chaining. Moreover, MemoryVLA~\citep{shi2025memoryvla} offers a more generalized solution by introducing an explicit memory module, tackling the core non-Markovian property of long-horizon tasks.}

\noindent\textbf{\small Preserving foundational VLM capabilities}\\[4pt]
\textcolor{black}{A central motivation for building VLA models is to leverage the vast semantic knowledge and reasoning skills embedded in pre-trained VLMs. However, a fundamental challenge emerged: fine-tuning these powerful, generalist models on narrower, domain-specific robotics data often leads to catastrophic forgetting. This phenomenon causes the VLA to lose the very conversational and reasoning capabilities that motivated its creation in the first place, degrading the model's ability to understand novel instructions and generalize. To overcome this, some works have focused on developing architectural and training strategies to preserve, reactivate, and effectively transfer these foundational VLM capabilities into the embodied domain. Early efforts focused on creating a synergy between the reasoning of autoregressive models and the precision of diffusion-based policies. DiffusionVLA~\citep{wen2025diffusionvla} proposes a unified framework that integrates an autoregressive component for reasoning with a diffusion model for control. It introduces a ``reasoning injection module'' to embed self-generated textual reasoning into the diffusion policy, enriching the learning process with explicit reasoning signals.}

\textcolor{black}{Subsequent research identifies that performance degradation stems from direct conflicts within the model's parameter space. ChatVLA~\citep{zhou2025chatvla} identifies two key challenges: ``spurious forgetting'', where robot-centric training overwrites crucial visual-text alignments, and ``task interference'', where the competing objectives of control and understanding degrade performance. To address this, ChatVLA introduces a MoE architecture to isolate task-specific parameters, minimizing interference. Building on this, ChatVLA-2~\citep{zhou2025vision} refines this approach by using a dynamic MoE and a two-stage training pipeline to ensure the model's actions reliably follow its internal reasoning. Taking a different perspective, some work focuses on the training dynamics as the root cause of knowledge degradation. \citet{driess2025knowledge} propose knowledge insulation, which stops the gradient flow from the action expert back into the VLM backbone. This technique insulates the VLM's knowledge from disruptive gradients while the VLM is simultaneously fine-tuned using a safer parallel learning signal: an autoregressive next-token prediction loss on discretized actions. Contrasted to MoE-based methods, this approach modifies the training graph rather than the core model architecture, allowing the VLM and action expert to be trained in parallel without interference.}

\textcolor{black}{Another line of research recognized that preserving VLM capabilities is fundamentally tied to the training data. InstructVLA~\citep{yang2025instructvla} argues that ``catastrophic forgetting'' is exacerbated by the lack of instructional diversity in robotics datasets. It introduces Vision-Language-Action Instruction Tuning, a paradigm centered on a curated dataset rich with diverse instructions, captions, and question-answer pairs. This data-centric scheme ensures the model is continuously exposed to VLM-style tasks during training. Similarly, GR-3~\citep{cheang2025gr} demonstrates the power of this data-centric approach at scale by employing extensive co-training with web-scale vision-language data alongside robot and human trajectories. This shows that large-scale co-training is an effective method for producing a generalist policy that retains its foundational reasoning capabilities.}

\textcolor{black}{In summary, preserving VLM knowledge can mitigate the catastrophic forgetting issue of VLAs. The path evolved from simply using reasoning outputs (DiffusionVLA~\citep{wen2025diffusionvla}) to architecturally isolating task-specific parameters (ChatVLA~\citep{zhou2025chatvla}, ChatVLA-2~\citep{zhou2025vision}), modifying training dynamics to protect the VLM backbone (Knowledge Insulation~\citep{driess2025knowledge}), and enriching the training data to prevent skill atrophy (InstructVLA~\citep{yang2025instructvla}, GR-3~\citep{cheang2025gr}). While architectural solutions like MoE and training-dynamic solutions like gradient stopping both effectively tackle task interference, they represent different strategies: MoE creates separate ``specialists'' within one model, while insulation shields the generalist ``brain'' (reasoning) from the specialist ``limb'' (action). The data-centric co-training approach is complementary and has become a cornerstone of recent state-of-the-art models, suggesting a growing consensus that to keep a VLM's abilities, one must continue to train it on VLM-style data.}

\subsubsection{\textcolor{black}{Internal world models and reasoning}}
\textcolor{black}{A key limitation of standard VLAs that utilize a behavioral cloning training scheme is their reactive and ``black box'' attribute, where they map observations directly to actions without explicit intermediate reasoning or planning. This approach struggles with long-horizon tasks that require anticipating future outcomes. To address this, recent research integrates predictive world models into VLAs, enabling them to reason about the future and ground their actions in an understanding of environmental dynamics.}

\textcolor{black}{One line of work focuses on using visual foresight to guide action generation. \citet{tian2025predictive} first predicts a future visual state based on the current observation and language goal, and then use an inverse dynamics model to determine the actions needed to reach that state. By training both the visual prediction and action generation modules jointly in an end-to-end manner, the model learns a synergistic relationship where visual prediction informs action planning, and the need to generate plausible actions refines the model's understanding of world dynamics. CoT-VLA~\citep{zhao2025cot} introduces a Visual Chain-of-Thought. The model generates a subgoal image as an explicit reasoning step, which is used in conjunction with the current observation and language instruction to condition the final action prediction. This strategy allows the model to ``think visually'' about how to accomplish a task before taking action.}

\textcolor{black}{WorldVLA~\citep{cen2025worldvla} unifies the action model and world model into a single autoregressive framework, which is trained to perform two complementary tasks: (i) as an action model, it predicts actions given the current state and a goal, and (ii) as a world model, it predicts the next visual state given the current state and an action, creating a synergistic learning process that mutually enhances each other. The world model's need to understand actions for accurate future prediction improves the action model's grounding, while the action model's interaction with the environment provides rich data for learning world dynamics. However, predicting entire future images is computationally expensive and can include irrelevant information (like static backgrounds). To improve efficiency, DreamVLA~\citep{zhang2025dreamvla} proposes forecasting a compact latent ``world embedding'' instead of raw pixels. This embedding captures critical aspects of the future, such as object dynamics, depth, and high-level semantics, providing the policy with a concise and relevant representation for planning. To prevent interference between these different knowledge types, the model employs a block-wise structured attention mechanism, disentangling the representations and ensuring that the final action is conditioned on a clean understanding of the predicted future.}

\textcolor{black}{In summary, the integration of internal world models into VLAs marks a shift from reactive to predictive control. Early methods like Seer~\citep{tian2025predictive} and CoT-VLA~\citep{zhao2025cot} established the paradigm of using pixel-level future predictions to guide actions, framing it respectively as a control problem and a cognitive reasoning step. WorldVLA~\citep{cen2025worldvla} advances this by creating a bidirectional link, where the model learns to predict future states from actions and, conversely, actions from future states, thereby building a more complete understanding of environmental dynamics. The most recent evolution, represented by DreamVLA~\citep{zhang2025dreamvla}, addresses the computational inefficiency of pixel-level prediction by forecasting a compact representation of the future. This trajectory reflects a move from concrete visual prediction towards more abstract and efficient future reasoning.}

\subsection{\textcolor{black}{Optimization for action}}
\textcolor{black}{The representation and generation of robot actions are critical bottlenecks in VLA performance. A primary challenge with early VLA models was their reliance on discrete representations of continuous actions (like OpenVLA~\citep{kim2024openvlaopensourcevisionlanguageactionmodel}), which limited the fidelity and smoothness of robot movements for dexterous manipulations. To mitigate this issue, $\pi_{0}$~\citep{black2024pi_0} is proposed to shift away from discrete action tokens toward a continuous generative process, which models the distribution of entire action sequences using \emph{flow matching}. By adding a dedicated ``action expert'' module to a standard VLM backbone, $\pi_{0}$ could generate precise and fluent continuous action chunks (50 Hz trajectories), enabling it to master highly dexterous tasks. While the flow matching approach effectively solves the action precision problem, it introduces computational overhead during training.}

\textcolor{black}{\citet{pertsch2025fast} observe that the poor performance of autoregressive VLA results from redundancy, as consecutive action tokens are highly correlated, which provides a weak learning signal for the next-token prediction objective. Their Frequency-space Action Sequence Tokeniser (FAST) compresses the action sequence by converting it into the frequency domain using the Discrete Cosine Transform, thereby removing temporal redundancy before tokenization. Their results demonstrate that a VLA trained with FAST could match the performance of the $\pi_{0}$ flow-matching VLA while reducing training time by up to 5x.} \textcolor{black}{With precision and efficiency addressed separately, a broader question remains: ``\emph{how can a single model generalize its actions to unseen scenarios}''. The answer requires a method that can efficiently train a single model on massive heterogeneous data sources while still enabling fine-grained and real-time control at deployment. $\pi_{0.5}$~\citep{pmlr-v305-black25a} answers this by \emph{jointly} learning discrete and continuous action heads. In the pre-training stage, it uses the FAST tokenizer~\citep{pertsch2025fast} to handle diverse robotic action data at a large scale. During the post-training stage, a flow-matching action expert (as in $\pi_{0}$) is employed for high-fidelity continuous control and efficient real-time inference. At run-time $\pi_{0.5}$ first predicts a \emph{high-level sub-task} (text) and then refines it into continuous motor commands, achieving open-world manipulation such as cleaning an unseen kitchen.} 

\textcolor{black}{Some works explore the action generation problem from the model architectural perspective. Both autoregressive decoders and the external continuous diffusion heads used in many state-of-the-art models created an architectural disconnect between the VLM backbone and the action generation module. To create a more cohesive and scalable architecture, Discrete Diffusion VLA ~\citep{liang2025discrete} introduces a framework to model discretized action chunks using a discrete diffusion process within a single unified transformer. This approach trains the entire model with the same cross-entropy objective used by the VLM backbone, thereby preserving pre-trained priors while enabling parallel and iterative refinement of the action chunk. This contrasted with all previous methods by treating action decoding not as a separate task for an external module, but as a native capability of the core transformer, paving the way for better scaling and architectural consistency.} \textcolor{black}{These approaches provide a perspective for action generation in VLAs: \emph{control precision} demands continuous action decoders, \emph{scalability} requires smarter tokenization to reduce redundancy in action sequences, and \emph{generalization} needs a hybrid that combines large-scale pre-training with fine-grained control.}

\subsection{\textcolor{black}{Optimization for Adaptation}}
\textcolor{black}{Although VLAs unify vision, language, and action within a single policy, the grounding between language instructions and executable actions remains sensitive to deployment conditions. When applied to new scenes, tasks, embodiments, or object distributions, a VLA must adapt to downstream settings while preserving the semantic and instruction-following priors inherited from large-scale vision-language pretraining. This section focuses on adaptation methods that improve or preserve language-action grounding in manipulation, especially under limited data, real-time control requirements, and interactive deployment.}

\textcolor{black}{A practical requirement for adaptation is the improvement of the deployability of VLA policies, ensuring responsive enough for real robot control. For example, OpenVLA-OFT~\citep{kim2025fine} revisits the fine-tuning design of OpenVLA~\citep{kim2024openvlaopensourcevisionlanguageactionmodel} and shows that parallel decoding, action chunking, and continuous action prediction can substantially improve the responsiveness of language-conditioned control, which improves success rate by 20\% on LIBERO and boosting control frequency 26x against OpenVLA. By reducing inference latency and producing temporally coherent action sequences, such designs help language-specified commands remain executable under the timing constraints of real robot control.} \textcolor{black}{However, improving control efficiency alone does not guarantee robust language grounding under new objects, scenes, or referring expressions. When only limited downstream demonstrations are available, adaptation should improve task-specific behavior without overwriting the semantic knowledge that supports open-vocabulary instruction following. ControlVLA~\citep{li2025controlvla} mitigates this issue by injecting object-centric visual information into a largely frozen VLA backbone, enabling adaptation to new task-relevant objects and language references while preserving pretrained language priors.}

\textcolor{black}{Nevertheless, supervised fine-tuning is constrained by the quality of the demonstrations it imitates, which may be suboptimal, incomplete, or unsafe in downstream manipulation settings. To mitigate this, some works use post-training or interaction to align language-specified goals with successful manipulation behavior. post-training methods such as ConRFT~\citep{chen2025conrft} and RIPT-VLA~\citep{tan2025interactive} refine pretrained VLAs through task rewards, online interaction, or human intervention. By incorporating downstream feedback, these methods correct failures in alignment with language goals, improve instruction-following, and strengthen the mapping from linguistic intent to physical action. Overall, language-grounded adaptation mitigates a central challenge for VLAs, e.g., transferring broad linguistic and visual priors to new manipulation settings while maintaining reliable language-action alignment.}

\textcolor{black}{Additionally, Appendix~\ref{sec:apxa} (Table~\ref{tab:vla_table}) provides a consolidated classification of representative VLA models discussed in this section. It summarizes the models according to the optimization dimensions introduced in this section: perception, reasoning, action generation, and learning/adaptation, while also comparing their observation modalities, action-generation paradigms, policy mechanisms, pretraining settings, and evaluation scenarios.}

\section{Comparative analysis}
\label{sec:comparative_analysis}

\textcolor{black}{The preceding sections systematically categorized language-conditioned methodologies based on the functional role of language within the control loop. While this taxonomy clarifies \textit{how} language is integrated, it is equally important to analyze the structural design choices and practical trade-offs that determine a system's real-world feasibility. Critical dimensions often affect the success of a method beyond its theoretical formulation, such as the granularity of action spaces, the cost of data acquisition, system latency, and the validity of evaluation setups. Furthermore, to fully contextualize the utility of language, it is necessary to compare it with alternative task-specification modalities, such as goal images or videos. In this section, we conduct a detailed comparative analysis of representative approaches, evaluating them along five cross-cutting axes designed to highlight the fundamental engineering decisions in the field.}

\begin{itemize}
    
    \item \textbf{Action granularity:} We analyze whether methods output high-level skills (a sequence of actions) or low-level torques/poses (fundamental commands that directly control a robot's motors/limbs). This is a critical design choice that determines a system's precision, ability to handle contact-rich tasks, and real-time feasibility. It reveals the trade-off between long-horizon planning (coarse actions) and fine-grained control (fine actions, e.g., joint torques).
    
    \item \textbf{Data and supervision regimes:} We compare methods based on their data sources (e.g., expert demos, play data, web data) and supervision signals (e.g., human labels, learned rewards, FM-generated code). This axis is crucial as it directly impacts annotation cost, scalability, and a method's ability to generalize to out-of-distribution scenarios.
    
    \item \textbf{System cost and latency:} We now explicitly compare the computational and memory costs of different approaches, both at training and inference time. This is a vital organizing principle because real-world robots must operate under strict hardware and latency constraints. This axis highlights the practical gap between theoretically powerful models and deployable systems.
    
    \item \textbf{Environments and evaluations:} We compare the benchmarks, tasks, and hardware used for evaluation. \textcolor{black}{We further discuss the manipulation task categories to characterize task difficulty.} This axis aims to understand the external validity of a method's claims, highlighting the critical distinction between results achieved in controlled simulations versus the complexities of the real world.
    \item \textcolor{black}{\textbf{Task Specification:} We compare methods based on whether they condition on language instructions or other modality specifications (e.g., goal images/videos). This axis reveals the trade-offs between the expressiveness and flexibility of language versus the directness and precision of visual goals.}

\end{itemize}

\textcolor{black}{To support this cross-sectional comparison, Appendix \ref{sec:apxb} (Tables \ref{tab:approaches}-\ref{tab:approaches-v2}) provides a unified tabular summary of representative language-conditioned robotic manipulation approaches discussed across Sections \ref{sec:language-for-state-evaluation}-\ref{sec:language-cognitive-planning-reasoning}. The tables consolidate key implementation and evaluation dimensions, including benchmarks or simulation engines, language and perception modules, real-world validation, and whether a method relies on foundation models, reinforcement learning, imitation learning, or motion planning.}

\subsection{Action granularity}
Action granularity denotes the resolution of the control signal produced by the learned component, defining the end-to-end coverage of the learning method. We consider three levels: skill-level (selects pre-defined skills such as ``open drawer''), trajectory-level action (predicts waypoints/poses), and low-level control (outputs high-rate joint torques/velocities). 

\subsubsection{Skill-level actions}
Skill-level actions represent the nature composition of the language instruction. Representative approaches issue high-level skills or API calls rather than time-parameterized waypoints/poses of the motion planner or torques of the robot's joints. The planner-level LLMs such as SayCan~\citep{ahn2022can}, Code as Policies~\citep{liang2023code}, Inner Monologue~\citep{huang2022inner}, ReKep~\citep{huang2024rekep} follows this design. Additionally, task and motion planning (TAMP) approaches, such as PDDLStream~\citep{Garrett2018PDDLStreamIS}, provide the backbone that binds these symbolic steps to geometric feasibility. This granularity matches language well, where skills are ``natural language units'', showing strong constitutionality and interpretability. The main failure mode is skill library uncoverage, which means the policy-asked skill primitive is missing in the library. In practice, systems address these failures with affordance or value filters~\citep{ahn2022can, huang2024rekep}, schema validation for generated code~\citep{liang2023code, singh:progprompt:ar}, and scene-aware success checks with re-planing~\citep{ahn2022can, liang2023code, huang2022inner}.

\textcolor{black}{
\subsubsection{Trajectory-level actions}
Trajectory-level actions involve outputting end-effector waypoints or trajectories, rather than skills. 
Many language-conditioned IL approaches~\citep{lynch2020language,jang2022bc, wang2023mimicplay, mees23hulc2, zhou2023languageconditioned} fall here: they map an image observation and language instruction to either a next waypoint or an entire trajectory to achieve the described goal. 
Similarly, some language-conditioned RL methods~\citep{bing2023meta, pang2023natural, aljalbout2025limt} optimize a trajectory that satisfies a language-specified reward or success condition. 
Diffusion models have also emerged as powerful trajectory generators, learning a distribution over motion sequences that can be sampled by iterative refinement.
For instance, Diffusion Policy~\citep{chi2023diffusion} and subsequent variants~\citep{ze20243d, lu2024manicm} generate multi-step actions by denoising random trajectories, guided by language or goal images. 
This granularity lies between skill and low-level control, which is expressive enough for contact and geometry while still amenable to language conditioning via goals, keyframes, or affordance hints. The main failure modes are covariate shift that yields trajectory drift and compounding error over long horizons. Common mitigations include receding-horizon replanning~\citep{nair2022learning,huang2023voxposer}, which optimizes a short-horizon trajectory, executes only the first slice, resenses, and replans; and intervention-based dataset aggregation~\citep{jang2022bc}, which logs brief human corrections during rollouts and incorporates them into the training set to mitigate distribution (covariate) shift.
}

\subsubsection{Low-level control}
\textcolor{black}{A challenge of many language-conditioned manipulation tasks, such as dexterous manipulations~\citep{ma2024eureka}, or handing fragile/deformable objects~\citep{kobayashi2025bi}, which are defined by physical contact and force dynamics, rather than just spatial positions. This motivates the need for integrating language with low-level torque/force control, which directly commands the robot's joint torques at high control frequency (e.g., 100Hz+) that allows robots to perform contact-rich tasks with greater precision and compliance. Current methods integrate this capability in different ways, including reinforcement learning, imitation learning, and foundation models, which present unique challenges and opportunities.} \textcolor{black}{A primary bottleneck for reinforcement learning is the difficulty of manual reward design. Crafting reward functions that guide a policy to learn a complex, dynamic, and contact-rich skill (e.g., pen spinning) is extremely hard. EUREKA~\citep{ma2024eureka} mitigates this issue by using the LLM's coding and zero-shot generation capabilities to perform evolutionary optimization over the reward function code itself. EUREKA successfully generates reward functions that enabled a simulated robot hand to learn complex low-level manipulation skills (like spinning a pen). While RL is powerful, imitation learning from human demonstrations is often more data-efficient. The core problem for IL in this domain is that standard kinesthetic teleoperation only captures position data, failing to record the demonstrator's force intent. To solve this, Bi-LAT~\citep{kobayashi2025bi} introduces a framework based on bilateral control, a teleoperation setup that records not only joint position and velocity but also joint torque data. The key insight is then to condition this multimodal policy on natural language instructions (like ``softly/strongly twist the sponge''), explicitly modulating its force output. }

\textcolor{black}{The rise of foundation models introduces new challenges, such as creating a unified model for contact physics. ManiFoundation Model~\citep{xu2024manifoundation} addresses this by framing manipulation as a problem of ``contact synthesis''. It takes object and robot point clouds as input and predicts the optimal contact points and contact forces required to achieve a desired target motion, offering a general solution that bypasses direct torque prediction. Another challenge is integrating low-level torque control into large pre-trained VLA models (which are typically pose-based). TA-VLA~\citep{zhang2025elucidating} explores this by elucidating the design space for creating ``torque-aware'' VLAs, providing several insights. They find that (i) torque signals are best integrated into the action decoder, as they align better with other proprioceptive signals like joint angles, (ii) torque history can be summarized into a single token for temporal context, and (iii) performance improves by adding an auxiliary task of predicting future torque, which helps the model build a physically grounded internal representation.}

\textcolor{black}{Despite these advances, many language-conditioned policies still operate one step higher, predicting the trajectory of the end-effector poses and delegating torques to the execution layer. We identify two primary reasons for this: data labeling and the sim-to-real gap. Collecting clean, time-aligned torque labels with accurate force sensing and per-joint calibration is expensive, whereas trajectory data can be easily obtained via teleoperation and is less noisy. Additionally, simulating friction, compliance, and actuator dynamics can be challenging. Torque-level policies trained in simulation often overfit to the simulator's specific physics, whereas trajectory outputs tracked by a robust low-level controller can result in a more seamless transfer.}

\vspace{-1em}
\textcolor{black}{
\subsection{Data and supervision regime}
The data and supervision regime specifies where training signals come from and how they are provided. We use it to compare annotation cost, reachable-state coverage, and out-of-distribution (OOD) behavior across methods. 
\subsubsection{Data sources}
Language-conditioned manipulation relies on two broad categories of data: (i) Robotic interaction data (direct experience) and (ii) Web-scale data (human-curated content). Robotic data is often further split by how structured it is:
}

\textcolor{black}{
\noindent
\textbf{Expert demonstrations:} Demonstrations of tasks (via teleoperation~\citep{mandlekar2018roboturk}, kinesthetic teaching~\citep{calinon2007active}, or VR control~\citep{zhang2018deep}) paired with language instructions labeled by experts~\citep{jang2022bc,brohan2022rt}. They are high-quality and task-directed, making them effective for imitation learning. However, they are costly to collect and scale. Each demonstration must be carefully performed and often manually labeled with a description. Despite the cost, several works have compiled demonstration datasets for dozens of manipulation tasks (e.g., RT-1~\citep{brohan2022rt}’s dataset with 130k demos), enabling supervised policies to learn a range of skills from ``open drawer'' to ``flip light switch''.
}

\textcolor{black}{
\noindent 
\textbf{Unstructured play data:} This refers to large collections of robot experience without strict task labels – e.g., a robot or a human teleoperator freely interacts with objects, producing a wide variety of behaviors. Such play or on-robot logs are much easier to gather at scale since they don’t require labeling each sequence with a specific instruction. Despite the convenience of collecting these data, making use of them conditioned on language becomes a challenge. Recent techniques address this by relabeling play data with language post-hoc. For instance, Learning from Play~\citep{pmlr-v100-lynch20a} introduces relabeling unlabeled trajectories with goal images.
\citet{lynch2020language,lynch2023interactive} combine these play data with goal images and 1\% language-labeled data with language instruction to train a language-conditioned policy, similar to HULC and HULC++~\citep{mees2022matters,mees23hulc2}. Such play+language strategies significantly expand state coverage and diversity compared to limited expert demos and have therefore also been explored for extracting affordances~\citep{borja22icra} and learning goal-conditioned policies via offline reinforcement learning~\citep{rosete2022corl}. 
}

\textcolor{black}{
\noindent
\textbf{Web-scale data:} Internet data provides broad priors that cut robot data needs. Image–text pretraining~\citep{radford2021learning, alayrac2022flamingo} combines visual and textual modalities and is helpful for language grounding in current perception. Text-only pretraining~\citep{chowdhery2022palm, brown2020language} offers a rich source of common-sense knowledge and reasoning abilities. 
\subsubsection{Supervision}
Supervision indicates how robots learn from data. We classify supervision by the form of the learning signal it provides to the learner entity. This yields three types of supervision paradigms for language-conditioned manipulation systems. Some approaches utilize a combination of these supervision paradigms. 
}

\textcolor{black}{
\noindent
\textbf{Targets labels} The targets here are the per-sample values (actions, subgoals, success/value labels) the learner is asked to match or predict during training. For example, recent generalist policies learn from these labels via imitation learning. RT-2~\citep{brohan2023rt} co-finetunes a web-scale VLM jointly with robot trajectories and then supervises action outputs, improving zero-shot instruction following in the real world. OpenVLA~\citep{kim2024openvlaopensourcevisionlanguageactionmodel}, Octo~\citep{octomodelteam2024octo} follow the same supervised recipe, leveraging strong pretrained vision backbones and more training data. 
}

\textcolor{black}{
\noindent
\textbf{Outcome evaluations:} In outcome-driven supervision, the learner optimizes a scalar evaluation from execution (rewards, success, preferences, penalties). The scalar evaluation rates the quality of a state, action, or trajectory. It is used to drive learning without prescribing an exact target action. The learner then maximizes (or minimizes) this number via RL/MPC or on-policy updates. The evaluation can be manually defined~\citep{silva2021lancon} or realized by VLMs as success detectors~\citep{du2023visionlanguage}, LLMs for reward generation~\citep{yu2023language, ma2024eureka}. 
}

\textcolor{black}{
\noindent 
\textbf{Auxiliary supervision}
Auxiliary self-supervision shapes the learner without supervising the primary task directly. These additional learning objectives have two main advantages: 
(i) Ensuring important information
flows in the neural networks, which is critical for decision-making. (ii) Guiding the direction of gradient descent during training can potentially lead the neural network to a more optimal location in the parameter space. We classify them into three categories: \textbf{(i) visual attention}, which learns what/where to act via language-conditioned affordances and attention (e.g., region–text alignment~\citep{shridhar2022cliport, mees23hulc2}, handle/keypoint and graspability maps~\citep{kite, Mo_2021_ICCV, Contact_GraspNet}); \textbf{(ii) reconstruction}, which learns how to represent robust scenario representation through reconstruct image/video~\citep{shridhar2023perceiver,nair2020contextual, rahmatizadeh2018vision, ebert2018visual}, and \textbf{(iii) future prediction}, which learns what will happen next via next-states prediction~\citep{paxton2019prospection, ebert2018visual, bu2024closedloop} in pixel or latent space and keyframe prediction~\citep{xian2023chaineddiffuser} to stabilize long-horizon behavior and expose subgoals for planning.
}

\textcolor{black}{
\noindent
Many systems blend supervision paradigms across training phases. We write the objective as
\begin{equation}
    \mathcal{L}_\text{final} = \lambda_1 \mathcal{L}_\text{targets} + \lambda_2 \mathcal{L}_\text{evaluations} + \lambda_3\mathcal{L}_\text{auxiliary}\text{,}
\end{equation}
where $\mathcal{L}_{\text{targets}}$, $\mathcal{L}_{\text{evaluations}}$, and $\mathcal{L}_{\text{auxiliary}}$ correspond to fixed targets, outcome evaluations, and auxiliary losses, respectively. $\lambda_1,\lambda_2$, and $\lambda_3$ are time-dependent curriculum weights. A common schedule is to warm-start with fixed targets (imitation) for stability ($\lambda_1\!\uparrow$), anneal in outcome evaluations (rewards/preferences/validators) for robustness and long-horizon credit assignment ($\lambda_2\!\uparrow$ as $\lambda_1\!\downarrow$), and co-train with auxiliaries to improve grounding and invariances ($\lambda_3$ small but nonzero throughout).
}

\textcolor{black}{
\subsection{System cost and latency}
\subsubsection{Training cost}
}

\textcolor{black}{
Scaling models and data improve capability and generalization, as shown by LLMs. Recent robotics systems increasingly ride these trends. However, scaling significantly raises the training cost. More GPU hours are needed to leverage foundation models in the robotic domain. Three training strategies are identified: (i) Training from scratch, (ii) Fine-tuning, and (iii) Prompt engineering.
}

\textcolor{black}{
\noindent
\textbf{Training from scratch:} This approach builds language-conditioned manipulation models from the ground up, without leveraging any prior pre-training on external data. Large models are initialized with random weights and trained end-to-end on robot-specific datasets. The training cost is enormous because the model must learning vision, language, and control solely from task data. For example, the Gato~\citep{reed2022a} agent is a pioneering vision-language-action model trained from scratch on tokenized images, text, and actions, achieving multi-task skills entirely from its training set. Such scratch-trained models demonstrate that it is possible to integrate language and control without foundations, while they also show the requirements for large training datasets and GPU resources~\citep{brohan2022rt, open_x_embodiment_rt_x_2023} to achieve downstream generalization ability. 
}

\textcolor{black}{
\noindent
\textbf{Fine-tuning:} Rather than start from scratch, this strategy adapts pre-trained foundation models to the robot manipulation domain. A model that has already learned broad visual and linguistic patterns is used as a backbone, then fine-tuned on robot-specific data (possibly with additional new layers or using LoRA~\citep{lora} strategy). The key advantage is leveraging prior knowledge: the foundation model’s understanding of semantics and vision greatly reduces the needed robot data and improves generalization to new conditions. For instance, Google’s RT-2~\citep{brohan2023rt} policy builds on pre-trained vision-language representations (PaLI-X~\citep{chen2023pali} and PaLM-E~\citep{driess2023palme}) and fine-tunes them with robotic data, \textcolor{black}{resulting in substantially better performance on novel objects and instructions compared to a scratch-trained policy, especially when those examples remain close to the pre-training distribution.} The trade-offs include overfitting and forgetting that the model can lose some of its prior knowledge if the domain is narrow and the fine-tuning data is limited.}

\textcolor{black}{
\noindent
\textbf{Prompt engineering :} In this strategy, the FMs remain frozen and is invoked at inference via prompting. An LLM proposes a sequence of subtasks that a robot executes using a library of skill primitives (e.g., SayCan~\citep{ahn2022can} or Code-as-Policies~\citep{liang2023code}). Because the FM is not fine-tuned, no additional training on robot data is required. The planner simply queries a pretrained LLM, which can generate coherent plans with prompts. The prompted, frozen-FM planners can call very large LLMs/VLMs and inherit broad open-ended knowledge so that they show strong generalization across instructions and scenes. However, since the model is decoupled from the embodiment, the planning is typically open-loop and prone to hallucinated or infeasible steps, struggling on long-horizon tasks.}

\textcolor{black}{
\subsubsection{Inference cost}
\begin{table}[]
    \caption{The inference cost and latency for sampled approaches. \emph{Cloud} denotes inference on remote compute; ‘–’ indicates not reported; ‘*’ marks values inferred from architectural details. }
    \centering
    \fontsize{7}{8}\selectfont 
    \setlength{\tabcolsep}{1pt}
    \begin{tabular}{l c c c c c}
        \toprule
        System & Model & Params & Hardware & Latency & Cloud\\
        \midrule
        RT-1~\citeauthor{brohan2022rt} & VLA policy & 35M & - & $\sim$3Hz & \textcolor{Salmon}{\xmark}\\
        RT-2-55B~\citeauthor{brohan2023rt} & VLA policy & 55B & TPU & 1-3Hz&\textcolor{JungleGreen}{\cmark}\\
        RT-2-5B~\citeauthor{brohan2023rt} & VLA policy & 5B & - & 1-3Hz & \textcolor{Salmon}{\xmark} \\
        OpenVLA~\citeauthor{kim2024openvlaopensourcevisionlanguageactionmodel} & VLA policy & 7B & RTX4090 & 3-6Hz & \textcolor{Salmon}{\xmark}\\
        CLIP-RT~\citeauthor{clip-rt} & VLA policy & 1B & - & 8-16Hz & \textcolor{Salmon}{\xmark}\\
        PaLM-SayCan~\citeauthor{ahn2022can} & LLM planner & 540B & TPU & - & \textcolor{JungleGreen}{\cmark}\\
        PaLM-E~\citeauthor{driess2023palme} & Embodied VLM & 562B & TPU & - & \textcolor{JungleGreen}{\cmark}\\
        \midrule
        Diffusion Policy~\citeauthor{chi2023diffusion} & Diffusion policy & $\sim$200M$^*$ & - & $\sim$1Hz & \textcolor{Salmon}{\xmark}\\
        DP3~\citeauthor{ze20243d} & Diffusion policy & $\sim$150M$^*$ & - & 5-6Hz & \textcolor{Salmon}{\xmark}\\
        ManiCM~\citeauthor{lu2024manicm} & Diffusion policy & $\sim$200M$^*$ & RTX4090 & $\sim$50-60Hz & \textcolor{Salmon}{\xmark}\\
        \bottomrule        
    \end{tabular}
    \label{tab:inference cost}
    \vspace{-1em}
\end{table}
}

\textcolor{black}{
\noindent
At deployment time, the inference time largely depends on the model scale and the computation resources. Systems that prompt a frozen LLM/VLM inherit the reasoning breadth of very large FMs but typically run open-loop, so plan proposals arrive on a seconds-scale and must be grounded with affordance/validation or closed-loop replanning. For end-to-end VLA policies, fine-tuning larger backbones improves generalization but increases per-step compute and memory, making multi-Hz closed-loop control harder unless the model is compressed or carefully engineered, as shown in Table~\ref{tab:inference cost}. Diffusion policy may incur additional runtime tax, since actions are sampled with multiple denoising steps, which can be mitigated by utilizing fast samplers~\citep{chi2023diffusion, ze20243d} (e.g., DDIM/DPMSolver) or consistency distillation~\citep{lu2024manicm}. We flag real-time performance as a core hurdle for large LLM/VLM-based systems. Although cloud computation can be used for inference to address on-board hardware limits, it makes performance contingent on network reliability, introducing latency jitter and drop-out failure modes. 
}

\subsection{Environments and evaluations}

\begin{table*}[t]
    \centering
    \caption{Comparison of Simulation Engines.}
    \begin{adjustbox}{width=\linewidth,center}
    \begin{tabular}{ p{4cm} p{10.5cm} p{5cm} }
        \toprule
        Simulator & Description & Use Cases \\
        \midrule
        PyBullet \citep{coumans2021} & With its origins rooted in the Bullet physics engine, PyBullet transcends the boundaries of conventional simulation platforms, offering a wealth of tools and resources for tasks ranging from robot manipulation and locomotion to computer-aided design analysis. & \citet{shao2021concept2robot} and \citet{mees23hulc2, mees2022matters} leverage PyBullet to build a table-top environment to conduct object manipulation tasks. \\
        \midrule
        MuJoCo \citep{todorov2012mujoco} &  MuJoCo, short for ``Multi-Joint dynamics with Contact", originates from the vision of creating a physics engine tailored for simulating articulated and deformable bodies. It has evolved into an essential tool for exploring diverse domains, from robot locomotion and manipulation to human movement and control. &  \cite{lynch2020language, pmlr-v100-lynch20a}  build a table-top environment based on MuJoCo. The famous benchmark Meta-world also builds on this simulator. \\
        \midrule
        CoppeliaSim \citep{rohmer2013v} & CoppeliaSim  offers a comprehensive environment for simulating and prototyping robotic systems, enabling users to create, analyze, and optimize a wide spectrum of robotic applications. Its origins as an educational tool have evolved into a full-fledged simulation framework, revered for its versatility and user-friendly interface. & \citet{stepputtis2020language} leverage CoppeliaSim to construct an environment where the agent can manipulate cups and bowls with different sizes and colors. \\
        \midrule
        NVIDIA Omniverse &  NVIDIA Omniverse offers real-time physics simulation and lifelike rendering, creating a virtual environment for comprehensive testing and fine-tuning of robot manipulation algorithms and control strategies, all prior to their actual deployment in the physical realm. &e.g., LEMMA benchmark~\citep{gong2023lemma} builds a multi-robot environment based on NVIDIA Omniverse. \\
        \midrule
        Unity & Unity is a cross-platform game engine developed by Unity Technologies. Renowned for its user-friendly interface and powerful capabilities, Unity has become a cornerstone in the worlds of video games, augmented reality (AR), virtual reality (VR), and simulations. &  Alfred~\citep{shridhar2020alfred}, which is based on Ai2-thor~\citep{kolve2017ai2} hausehold environment, use Unity game engine. \\
        \midrule
        UniSim \citep{yang2024learninginteractiverealworldsimulators} & Different from physical simulators, UniSim is learned by real-world interactions through generative modeling. It leverages diverse, naturally rich datasets such as abundant objects in image data, densely sampled actions in robotics data, and varied movements in navigation data to simulate both high-level instructions and low-level controls. & UniSim represents the pioneering attempt to build a simulator based on generative models. \\
        \bottomrule
    \end{tabular}
    \end{adjustbox}
    \label{tab:simulation engine}
\end{table*}

\hypertarget{simulation-engine}{}
\hypertarget{benchmarks}{}
\subsubsection{Simulations and benchmarks}
A variety of task domains and benchmarks are developed to assess the novel methodologies’ performance. 
In this section, we delve into the environmental settings and benchmarks within the realm of language-conditioned robot manipulation. Simulation plays a pivotal role in accelerating advancements in robot manipulation by enabling rapid prototyping, thorough testing, and exploring complex scenarios that might be challenging to replicate in the real world. We introduce several simulators widely used in the field of robot manipulations, as shown in Table \ref{tab:simulation engine}. Benchmarks are important for assessing and advancing the capabilities of robotic systems. We explore the fundamental significance of benchmarks in the domain of robot manipulation based on the above simulation engine or a real-world setting. A comparison is shown in Table \ref{tab:benchmarks}, and illustrations of these benckmarks as shown in Figure \ref{fig:benckmark}. Benchmarks in both simulation and real-world environments are listed as follows. 
 \begin{table*}[t]
    \centering
    \caption{Comparison of existing language-conditioned robot manipulation benchmarks.}
    \begin{adjustbox}{width=\linewidth,center}
    \begin{tabular}{ l c c c c c c c c c }
        \toprule
        \multirow{2}{*}{Benchmark} & {Simulation Engine or} & \multirow{2}{*}{Embodiment} & \multirow{2}{*}{Data Size} & \multicolumn{3}{c}{Observations} & \multirow{2}{*}{Tool used} & \multirow{2}{*}{Multi-agents} & \multirow{2}{*}{Long-horizon}\\  
        \cmidrule{5-7} 
         & Real-world Dataset & & & RGB & Depth & Masks & & \\
        \midrule
        
        CALVIN~\citep{9788026} & PyBullet & Franka Panda & 2400k & \textcolor{JungleGreen}{\cmark} & \textcolor{JungleGreen}{\cmark} & \textcolor{Salmon}{\xmark} &\textcolor{Salmon}{\xmark} & \textcolor{Salmon}{\xmark} & \textcolor{JungleGreen}{\cmark} \\
        
        Meta-world \citep{yu2020meta} & MuJoCo & Sawyer & - & \textcolor{JungleGreen}{\cmark} & \textcolor{Salmon}{\xmark} & \textcolor{Salmon}{\xmark} &\textcolor{Salmon}{\xmark} & \textcolor{Salmon}{\xmark} & \textcolor{Salmon}{\xmark} \\
        
        RLBench~\citep{james2020rlbench} & CoppeliaSim & Franka Panda & - & \textcolor{JungleGreen}{\cmark} & \textcolor{JungleGreen}{\cmark} & \textcolor{JungleGreen!80}{\cmark} &\textcolor{Salmon}{\xmark} & \textcolor{Salmon}{\xmark} & \textcolor{JungleGreen}{\cmark}\\
        VIMAbench~\citep{jiang2023vima} & PyBullet & UR5 & 650K & \textcolor{JungleGreen}{\cmark} & \textcolor{Salmon}{\xmark} & \textcolor{Salmon}{\xmark} & \textcolor{Salmon}{\xmark} & \textcolor{Salmon}{\xmark} &  \textcolor{JungleGreen}{\cmark} \\ 
        LoHoRavens \citep{zhang2023lohoravens} & PyBullet & UR5 & 15k & \textcolor{JungleGreen}{\cmark} & \textcolor{JungleGreen}{\cmark} & \textcolor{Salmon}{\xmark} & \textcolor{Salmon}{\xmark} & \textcolor{Salmon}{\xmark} & \textcolor{JungleGreen}{\cmark}\\
        ARNOLD~\citep{gong2023arnold} & NVIDIA Omniverse & Framka Panda & 10k & \textcolor{JungleGreen}{\cmark} & \textcolor{JungleGreen}{\cmark} & \textcolor{JungleGreen}{\cmark} & \textcolor{Salmon}{\xmark} & \textcolor{Salmon}{\xmark} & \textcolor{JungleGreen}{\cmark}\\
        RoboGen~\citep{wang2023robogen} & PyBullet & Multiple & - & \textcolor{JungleGreen}{\cmark} & \textcolor{Salmon}{\xmark} & \textcolor{Salmon}{\xmark} & \textcolor{JungleGreen}{\cmark} & \textcolor{JungleGreen}{\cmark} & \textcolor{JungleGreen}{\cmark}\\
        \textcolor{black}{LIBERO}~\cite{liu2023libero} & MuJoCo & Franka Panda & 6.5k & \textcolor{JungleGreen}{\cmark} & \textcolor{Salmon}{\xmark} & \textcolor{Salmon}{\xmark} & \textcolor{Salmon}{\xmark} & \textcolor{Salmon}{\xmark} & \textcolor{JungleGreen}{\cmark} \\

        \midrule 
        
        Open X-Embodiment~\citep{open_x_embodiment_rt_x_2023} & Real-world Dataset & Multiple & 2419k & \textcolor{JungleGreen}{\cmark} & \textcolor{JungleGreen}{\cmark} & \textcolor{Salmon}{\xmark}  & \textcolor{Salmon}{\xmark}  & \textcolor{Salmon}{\xmark} & \textcolor{JungleGreen}{\cmark} \\
        \textcolor{black}{DROID}~\citep{khazatsky2024droid} & Real-world Dataset & Franka Panda & 76k & \textcolor{JungleGreen}{\cmark} & \textcolor{JungleGreen}{\cmark} & \textcolor{Salmon}{\xmark}  & \textcolor{Salmon}{\xmark}  & \textcolor{Salmon}{\xmark} & \textcolor{JungleGreen}{\cmark} \\
        \textcolor{black}{Galaxea Open-world}~\citep{jiang2025galaxea} & Real-world Dataset & Galaxea R1 Lite & 100k & \textcolor{JungleGreen}{\cmark} & \textcolor{JungleGreen}{\cmark} & \textcolor{Salmon}{\xmark}  & \textcolor{Salmon}{\xmark}  & \textcolor{Salmon}{\xmark} & \textcolor{JungleGreen}{\cmark} \\
        \bottomrule
    \end{tabular}
    \end{adjustbox}
    \label{tab:benchmarks}
    \vspace{-1em}
\end{table*}

\noindent \textbf{CALVIN}~\citep{9788026}: Composing Actions from Language and Vision (CALVIN)  is an open-source simulated benchmark to learn long-horizon language-conditioned tasks constructed on PyBullet. It has four different yet structurally related environments so that it can be used for general playing and evaluating specific tasks. 34 different tasks can be performed under such environments. It has two different evaluation metrics, namely Multi-Task Language Control (MTLC) and Long-Horizon MTLC (LH-MTLC). MTLC aims to verify how well the learned multi-task language-conditioned policy generalizes to 34 manipulation tasks; LH-MTLC aims to verify how well the learned multi-task language-conditioned policy can accomplish several instructions in a row. \citet{mees2022matters, mees23hulc2}  and \citet{zhou2023languageconditioned} evaluate their models on this benchmark.

\input{tikz/benckmark}

\noindent \textbf{Meta-World}~\citep{yu2020meta}: Meta-World is a collection of 50 diverse robot manipulation tasks built on the MuJoCo physics simulator. It contains two widely used benchmarks, namely, ML10 and ML45. The ML10 contains a subset of the ML45 training tasks, which are split into 10 training tasks and 5 test tasks, and the ML45 consists of 45 training tasks and 5 test tasks. It used to be a platform to evaluate the performance of meta-reinforcement learning. \citet{bing2023meta, silva2021lancon, goyal2021pixl2r, nair2022learning} use Meta-world to test the performance of their language-conditioned models.

\noindent \textbf{RLBench}~\citep{james2020rlbench}: This benchmark includes 100 completely unique, hand-designed tasks ranging in difficulty, which share a common Franka Emika Panda robot arm, featuring a range of sensor modalities, including joint angles, velocities, and forces, an eye-in-hand camera, and an over-the-shoulder stereo camera setup. RLBench is built around a CoppeliaSim/V-REP, allowing for drag-and-drop style scene building and a whole host of robotic tools, such as inverse/forward kinematics, motion planning, and visualizations. \citet{shridhar2023perceiver} use RLbenchmark to evaluate their models' performance on 18 tasks.

\noindent \textbf{VIMAbench}~\citep{jiang2023vima}: \textcolor{black}{VIMAbench is a simulation benchmark built on the Ravens simulator~\citep{zeng2021transporter}  that supports multimodal prompts (text + images) for 17 tabletop manipulation task templates. 
It provides nearly 650,000 expert trajectories along with a four-level hierarchical evaluation protocol that progresses from randomized object placements to entirely novel tasks, enabling a systematic assessment of an agent’s generalization capability.}
The benchmark aims to support the development of generalist robot learning agents that can handle a wide variety of tasks specified via intuitive multimodal interfaces.
\textcolor{black}{For example, \citet{li2023mastering} applies VIMAbench to assess a new policy for multimodal language-conditioned robot control and reports meaningful performance gains.}

\noindent \textbf{LoHoRavens}~\citep{zhang2023lohoravens}: LoHoRavens is built upon the Ravens simulator~\citep{zeng2021transporter} and contains ten long-horizon language-conditioned tasks in total. 
The tasks are split into seen tasks and unseen tasks to evaluate the robot’s generalization performance. 
Unlike VIMAbench, which primarily focuses on short-horizon tasks, this benchmark emphasizes long-horizon scenarios where the robot must perform at least five sequential pick-and-place actions to accomplish a high-level instruction.
\textcolor{black}{\citet{meng2025data} build upon this benchmark to validate their code generation framework.}

\noindent \textbf{ARNOLD}~\citep{gong2023arnold}: ARNOLD is built on NVIDIA Omniverse, equipped with photo-realistic and physically accurate simulation, covering 40 distinctive objects and 20 scenes. 
It comprises 8 language-conditioned tasks that involve understanding object states and learning policies for continuous goals, and provides 10,000 expert demonstrations with diverse template-generated language instructions based on thousands of human annotations. 
\textcolor{black}{Its main strength lies in its high visual realism and detailed annotations, which enable richer language-grounded manipulation studies. However, its limited scale and task diversity make it less suitable for developing broadly generalizable policies across varied robots and environments. For example, \citet{lee2025dynscene}~use ARNOLD to train and evaluate their framework in dynamic scenes.}

\noindent \textbf{RoboGen}~\citep{wang2023robogen}: RoboGen is an automated pipeline that utilizes the embedded common sense and generative capabilities of the foundation models (ChatGPT, Alpaca~\citep{taori2023alpaca}) for automatic task, scene, and training supervision generation, leading to diverse robotic skill learning at scale. 
RoboGen considers tasks including rigid object manipulation, soft body manipulation, and legged locomotion. In addition, RoboGen integrates several functions, including Task Proposal, Scene Generation, Training Supervision Generation, and Skill Learning.

\textcolor{black}{
\noindent \textbf{LIBERO}~\citep{liu2023libero}: LIBERO is a simulation benchmark originally for lifelong language-conditioned robot manipulation.
It introduces a procedural task generation pipeline that automatically creates infinite manipulation scenarios with natural language instructions, enabling studies of knowledge transfer under distribution shifts in objects, spatial layouts, and task goals. 
The benchmark consists of four curated task suites based on Franka Panda (LIBERO-SPATIAL/-OBJECT/-GOAL/-100), totaling 130 high-quality, language-conditioned tasks with human-teleoperated demonstrations. 
Although LIBERO was originally designed for lifelong imitation learning, it has now become a widely used simulation benchmark for evaluating the reasoning and task generalization capabilities of VLA models~\citep{kim2025fine,zhao2025cot,cen2025worldvla}.}

\noindent \textbf{Open X-Embodiment}~\citep{open_x_embodiment_rt_x_2023}: Open X-Embodiment is a real-world large-scale dataset, which is assembled from 22 different robots collected through a collaboration between 21 institutions, demonstrating 527 skills and 160266 tasks. 
Over 2M+ robot trajectories are stored in RLDS data format, which stores data in serialized TFRecord files and supports diverse action spaces and input modalities across different robot setups, including varying configurations of RGB cameras, depth sensors, and point clouds.
\textcolor{black}{Open X-Embodiment is currently one of the few real-world datasets offering large-scale, language-conditioned robot manipulation data with broad diversity in both robot embodiments and task distributions. Recent VLA models have widely adopted it for large-scale pretraining \citep{kim2024openvlaopensourcevisionlanguageactionmodel,black2024pi_0,pmlr-v305-black25a}.}

\textcolor{black}{
\noindent \textbf{DROID}~\citep{khazatsky2024droid}: Distributed Robot Interaction Dataset (DROID) is a real-world, large-scale, in-the-wild robot manipulation dataset comprising 76k teleoperated trajectories ($\approx$350 hours) collected across 564 scenes, 52 buildings, and 86 tasks.
Each episode records synchronized observations from three RGB stereo camera streams (two external ZED 2 cameras and one wrist-mounted ZED Mini), depth maps, camera calibration, low-level robot states, and crowd-sourced natural language instructions.
DROID is collected with a standardized Franka Panda platform with a Robotiq 2F-85 gripper via VR-based teleoperation, providing diverse scenes and viewpoints that support reproducibility, 3D reasoning, and robust policy generalization.
$\pi_{0.5}$~\citep{pmlr-v305-black25a} utilize this dataset for fine-tuning and real-world evaluation.}

\textcolor{black}{
\noindent \textbf{Galaxea Open-World Dataset}~\citep{jiang2025galaxea}: This dataset is a real-world open-world manipulation dataset containing 100k teleoperated trajectories ($\approx$500 hours) across 150 tasks and 50 diverse scenes with over 1,600 unique objects. 
Data is collected using the Galaxea R1 Lite, a mobile bimanual robot with 23 DoF, equipped with a head stereo camera and dual wrist cameras. 
All demonstrations are recorded through isomorphic teleoperation and annotated with fine-grained subtask-level language labels, ensuring consistent embodiment, multimodal alignment, and high-quality data for policy learning.
This real-world dataset provides a valuable foundation for fine-tuning humanoid-based, bi-manual VLA models.}

\textcolor{black}{
While recent progress in language-conditioned robot manipulation has produced a growing number of simulators, benchmarks, and real-world datasets, the field still lacks a unified evaluation protocol that can fairly quantify how accurate language grounding and contextual understanding contribute to task performance \citep{liu2023libero}.
Current benchmarks often focus on task success rates \citep{yu2020meta,james2020rlbench,jiang2023vima} but overlook semantic comprehension and context knowledge transfer between tasks \citep{pmlr-v305-black25a,meng2025preserving}.
Furthermore, there remains a pressing need for large-scale, real-world datasets that encompass a wider variety of embodiments, environments, and task distributions to support robust and generalizable language-conditioned robotic manipulation learning \citep{open_x_embodiment_rt_x_2023,pmlr-v305-black25a}.}

\textcolor{black}{
\subsubsection{Manipulation Task Families}
\label{sec:manipulation_task_spectrum}
Language-conditioned manipulation methods are evaluated on diverse task families with substantially different sources of difficulty. Therefore, direct comparison based only on reported task success can be misleading, as performance may reflect not only algorithmic capability but also the complexity of the underlying benchmark. For example, benchmarks dominated by object recognition and short-horizon pick-and-place primarily evaluate semantic grounding and visuomotor control, whereas tasks involving contact-rich interaction, deformable objects, or extended action sequences impose additional requirements on physical interaction modeling, temporal reasoning, and error recovery. To avoid conflating algorithmic capability with task difficulty, we group the manipulation tasks considered in this survey into three broad families, namely, \emph{Object-centric manipulation}, \emph{Interaction-centric manipulation}, and \emph{Long-horizon manipulation} (Table~\ref{tab:manipulation_task_categories}). This grouping is evaluation-oriented rather than a strict taxonomy of robot motions. The three task families correspond to different dominant sources of difficulty in language-conditioned manipulation.
}

\begin{table*}[t]
\centering
\small
\setlength{\tabcolsep}{4pt}
\caption{\textcolor{black}{Three main categories of manipulation tasks in language-conditioned robot manipulation. Some benchmarks span multiple task families.}}
\label{tab:manipulation_task_categories}
{
\begin{tabular}{l m{0.27\linewidth} m{0.24\linewidth} m{0.23\linewidth}}

\toprule
\textbf{Task category} & 
\textbf{Task Examples} & 
\textbf{Representative Benchmarks} &
\textbf{Illustration} \\
\midrule

\textbf{Object-centric} &
Object relocation, \newline Stacking, \newline Pose Manipulation &
Meta-World~\citep{yu2020meta},
\newline
RLBench~\citep{james2020rlbench},
\newline
VIMAbench~\citep{jiang2023vima} &
\includegraphics[width=1.8cm]{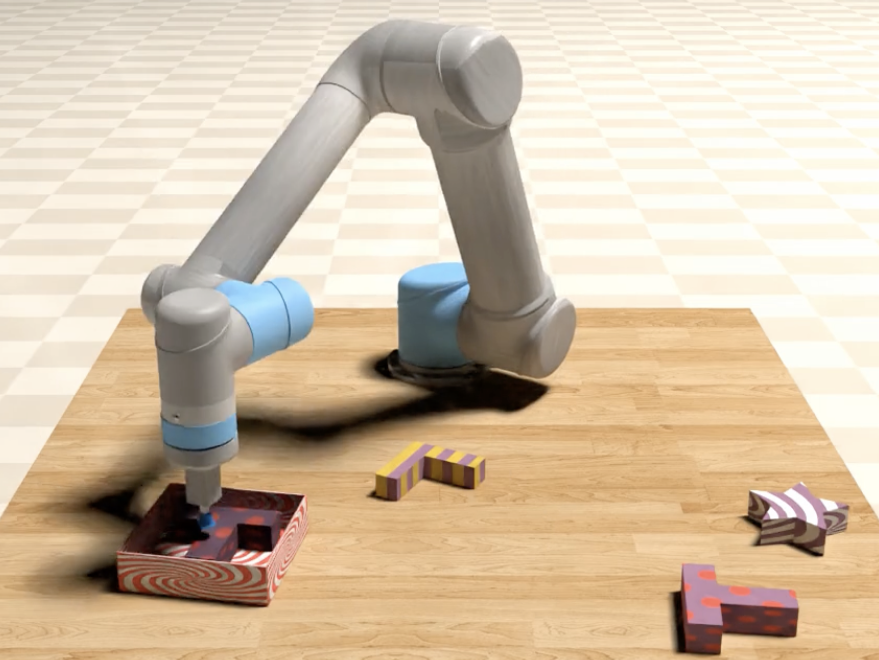}
\includegraphics[width=1.8cm]{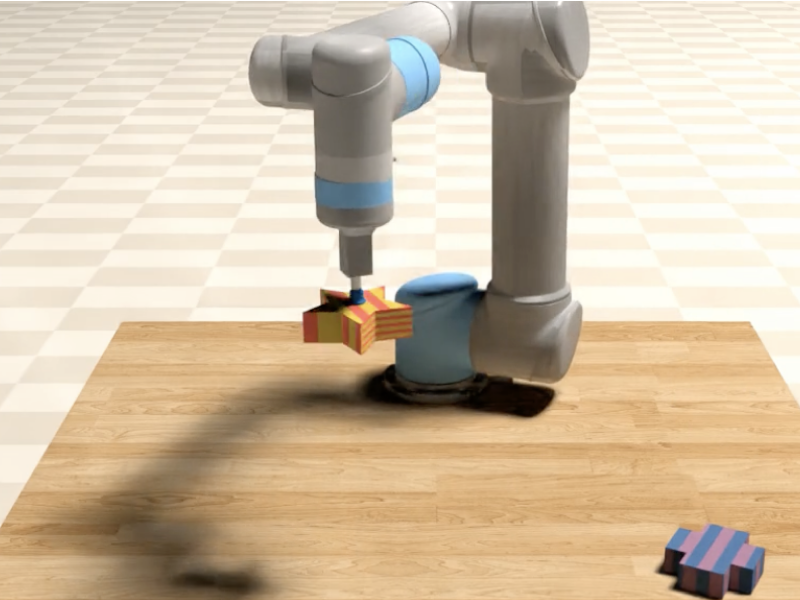}
\\
\hline

\textbf{Interaction-centric} &
Articulated-object Manipulation, 
\newline Deformable-object Manipulation, 
\newline Tool-mediated Manipulation &
ARNOLD~\citep{gong2023arnold}, 
\newline RoboGen~\citep{wang2023robogen}, 
\newline Open X-Embodiment~\citep{open_x_embodiment_rt_x_2023} &
\includegraphics[width=1.8cm]{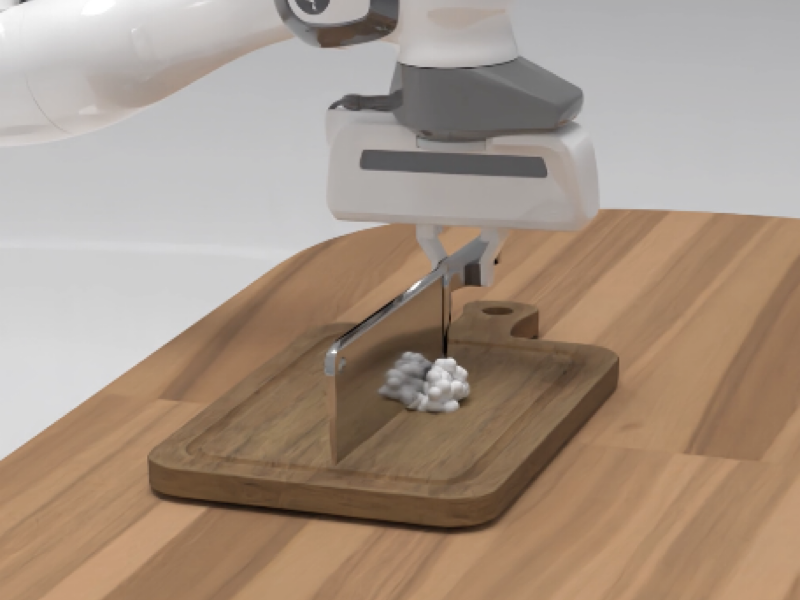}
\includegraphics[width=1.8cm]{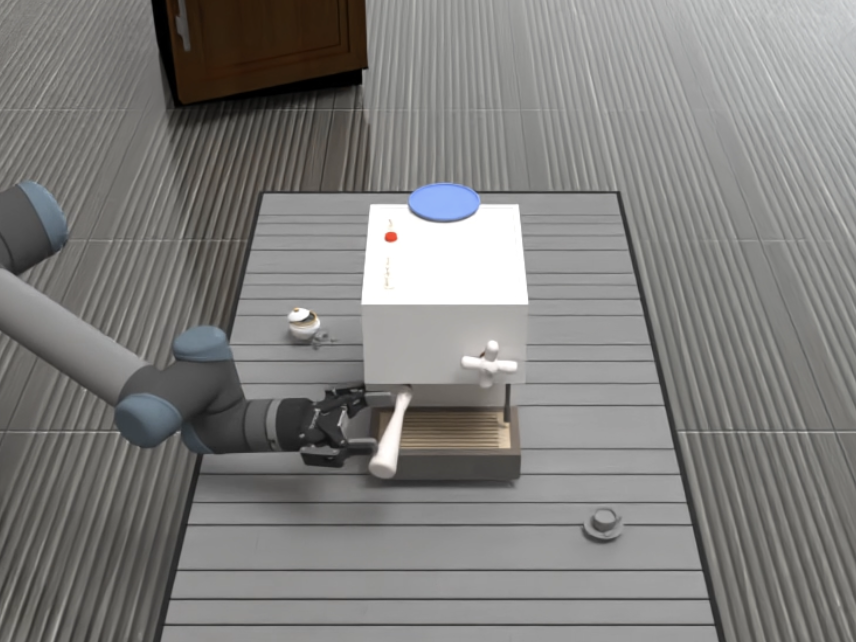}

\\
\hline

\textbf{Long-horizon} &
Multi-step Object Rearrangement, 
\newline Household Manipulation,
\newline Mobile Manipulation
&
CALVIN~\citep{9788026}, 
\newline LoHoRavens~\citep{zhang2023lohoravens}, \newline LIBERO~\citep{liu2023libero}, 
\newline Open X-Embodiment~\citep{open_x_embodiment_rt_x_2023}, DROID~\citep{khazatsky2024droid}, Galaxea Open-World~\citep{jiang2025galaxea} &
\includegraphics[width=1.8cm]{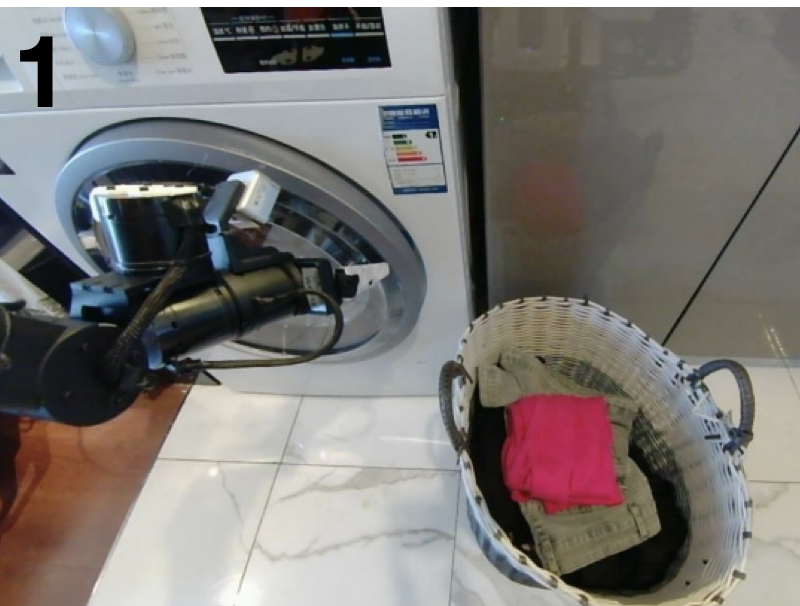}
\includegraphics[width=1.8cm]{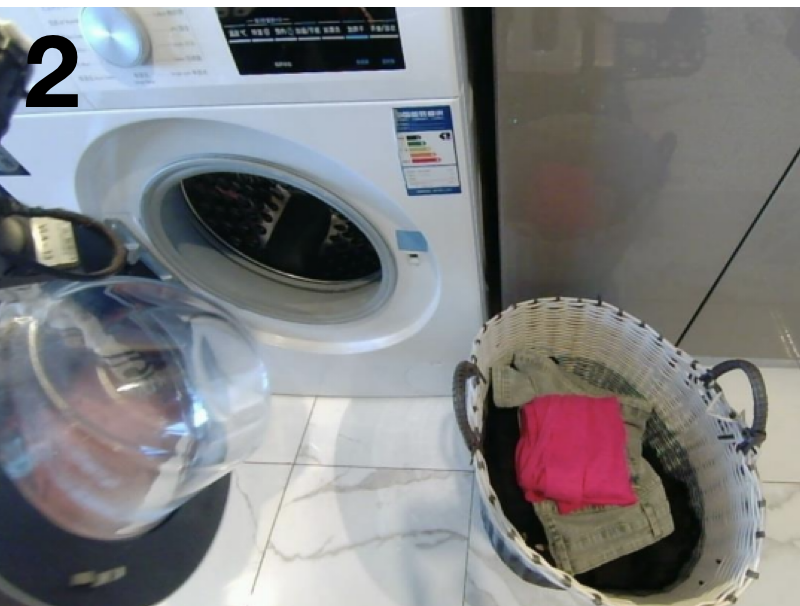}
\newline
\includegraphics[width=1.8cm]{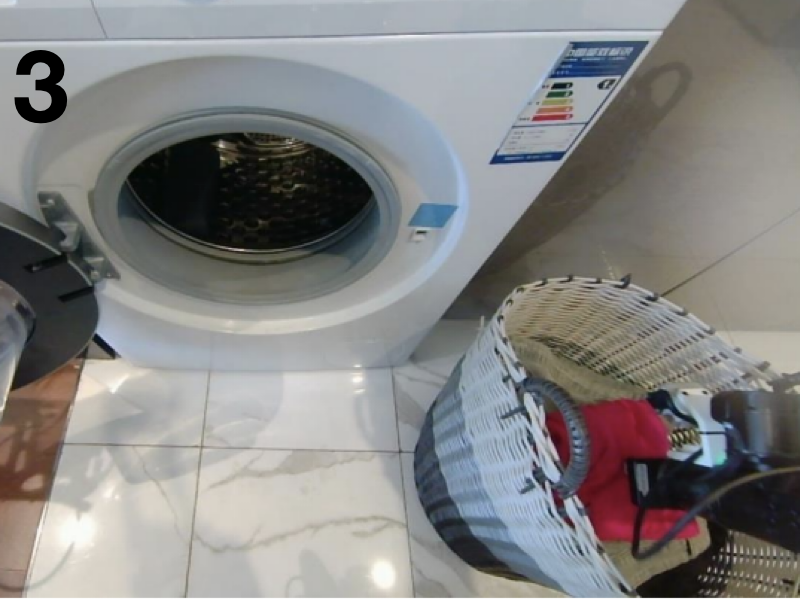}
\includegraphics[width=1.8cm]{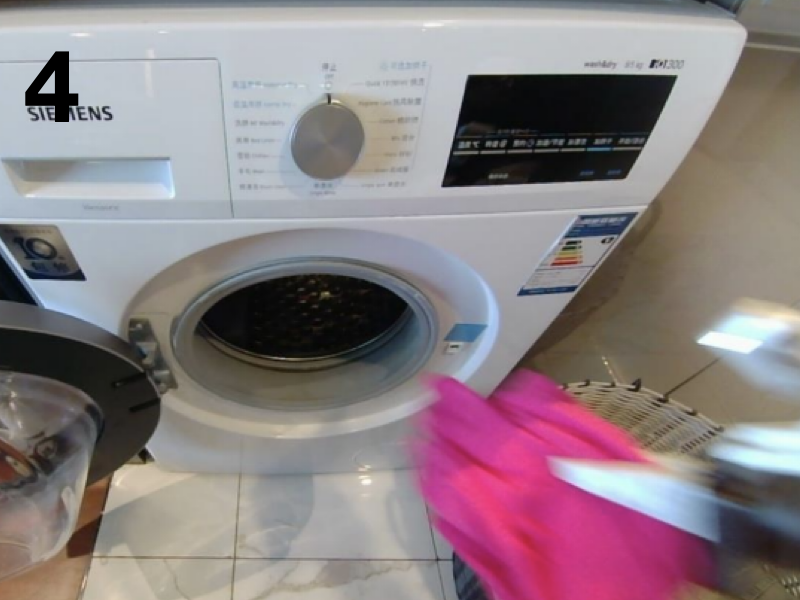}
\\
\bottomrule

\end{tabular}
}
\end{table*}

\textcolor{black}{
\textbf{Object-centric manipulation} refers to tasks in which the main objective is to identify, grasp, move, arrange, or reorient one object. 
Typical examples include object relocation, object sorting, stacking, and tabletop rearrangement. These tasks usually require grounding language instructions to object identities, attributes, spatial relations, grasp points, and target poses. Representative instructions include ``move the red block into the bowl'', ``Stack the red block on the blue one''. This category is widely used to evaluate visual-language grounding. Some early benchmarks, such as Meta-World~\citep{yu2020meta} and RLBench~\citep{james2020rlbench}, mainly focus on this category to evaluate whether an agent can identify the target object, ground the language instruction, and determine how to approach, grasp, and place the object. In these settings, the physical interaction between the robot end-effector and the object is often simplified. The manipulated objects are often rigid and geometrically simple, reducing the complexity of contact dynamics and allowing the evaluation to focus primarily on object grounding, spatial reasoning, and short-horizon visuomotor control.}

\textcolor{black}{
\textbf{Interaction-centric manipulation} focuses on the interaction of the agent with the physical world. It includes articulated-object manipulation, deformable-object manipulation, and tool-mediated Manipulation. Representative instructions include ``insert the plug'', ``fold the cloth'', ``use the spatula to flip the object'', and ``twist the sponge gently''. Compared with object-centric manipulation, these tasks require stronger physical grounding, including 3D spatial reasoning, force control, tactile feedback, deformation modeling, and tool-object affordance reasoning.  Representative benchmarks include ARNOLD~\citep{gong2023arnold}, with language-guided manipulation of articulated household objects such as drawers, cabinets, doors, ovens, and refrigerators; and RoboGen~\citep{wang2023robogen}, with automatically generated tasks involving articulated objects, tool use, and deformable or soft-object interactions. Large-scale real-world datasets such as Open X-Embodiment~\citep{open_x_embodiment_rt_x_2023}, DROID~\citep{khazatsky2024droid}, and Galaxea Open-World~\citep{jiang2025galaxea} also contain these tasks across diverse scenes and robot embodiments.
}

\textcolor{black}{
\textbf{Long-horizon manipulation} focuses on tasks that require the robot to complete a sequence of interdependent manipulation steps to satisfy a high-level instruction. It includes multi-step object rearrangement, sequential household manipulation, and mobile manipulation. Representative instructions include ``tidy up the table'', ``prepare a cup of coffee'', ``bring me the red cup from the kitchen'', and ``open the drawer, pick up the blue block, place it inside, and then close the drawer''. These tasks require task decomposition, subgoal sequencing, memory, semantic reasoning, navigation-manipulation coupling, closed-loop replanning, and failure recovery. Representative benchmarks include CALVIN~\citep{9788026}, LoHoRavens~\citep{zhang2023lohoravens}, and LIBERO~\citep{liu2023libero}, which focus on long-horizon tabletop tasks, requiring to complete multiple sequential tasks and evaluate the completeness. Open X-Embodiment~\citep{open_x_embodiment_rt_x_2023}, DROID~\citep{khazatsky2024droid} and Galaxea Open-World~\citep{jiang2025galaxea} provide diverse long-horizon manipulation demonstrations across different scenes, objects, and embodiments in the real world. This category is therefore particularly useful for evaluating whether a trained agent can maintain robust execution under long-horizon error accumulation and decompose complex high-level instructions into feasible subtask sequences.
}

\textcolor{black}{In summary,  these categories correspond to increasingly rich requirements in semantic grounding, physical interaction, and temporal reasoning; however, they are not mutually exclusive and should not be interpreted as a strict linear difficulty hierarchy. For instance, an articulated-object task may also require long-horizon planning, and deformable-object manipulation may be more challenging than a short multi-step rigid-object task. Nevertheless, this taxonomy provides a useful lens for interpreting reported performance, i.e., strong results on object-centric benchmarks primarily indicate effective visual-language grounding and short-horizon visuomotor control, whereas interaction-centric and long-horizon benchmarks further evaluate physical interaction modeling, planning, memory, and closed-loop robustness.}

\subsection{Task Specification}
\label{sec:lang_vs_visual}

\textcolor{black}{While this survey primarily focuses on \emph{language-conditioned} manipulation, it is important to position language as one task-specification modality within a broader multi-modal landscape. In parallel to language, a substantial body of work specifies tasks using \emph{goal images} or \emph{demonstration videos} (e.g., image/video-conditioned manipulation). These visual modalities occupy analogous functional roles in the manipulation control loop, directly aligning with our established taxonomy: (i) defining \emph{state evaluation} signals (rewards/costs) to quantify task progress (Section~\ref{sec:language-for-state-evaluation}), (ii) serving as a \emph{policy condition} that directly shapes action generation (Section~\ref{sec:language-as-policy-condition}), and (iii) providing structural priors for \emph{cognitive planning} in long-horizon tasks (Section~\ref{sec:language-cognitive-planning-reasoning}). In this subsection, we compare language and image/video conditioning through the lens of these three functional roles, highlighting the inherent strengths, limitations, and synergistic potential of each modality.}

\subsubsection{\textcolor{black}{State evaluation: Language-derived objectives \emph{vs.} video-derived progress signals}}

\textcolor{black}{In our taxonomy (Section \ref{sec:language-for-state-evaluation}), language for state evaluation refers to converting instructions into numerical feedback signals to define goals and quantify task progress. A fundamental challenge here is defining these signals when explicit reward engineering is intractable. Image and video-based approaches excel at providing \textbf{implicit and dense} progress signals in such scenarios. By learning state evaluation functions directly from visual demonstrations, these methods can map physical state changes to reward-like metrics without manual specification. For instance, XIRL~\citep{zakka2022xirl} learns a visual representation and a temporal progress measure from videos that transfer across different embodiments. Such visual-derived evaluations are particularly effective when success is defined by perceivable physical changes (e.g., whether a peg is fully inserted or gears are aligned~\citep{chen2023multimodality}) and when geometric precision is paramount.}

\textcolor{black}{Conversely, language-conditioned state evaluation specifies objectives through \textbf{explicit compositional semantics} (e.g., \emph{``place the red mug into the left drawer and close it''}). A central advantage of language is its ability to enforce abstract constraints that are difficult to infer solely from a visual goal or demonstration. Specifically, language seamlessly encodes \textbf{negative constraints} (e.g., \emph{``\textit{stay away from the yellow bottle}''}~\citep{sharma2022correcting} or \emph{``\textit{watch out for the vase}''}~\citep{huang2023voxposer}) and \textbf{safety protocols} (e.g., grounding modifiers like ``\textit{slowly}'' and ``\textit{upright}'' into trajectory optimizer parameters~\citep{park2019efficient}). Furthermore, instructions capture \textbf{relational preferences}, such as explicit keypoint constraints~\citep{huang2024rekep} or comparative directives like ``\textit{move farther from the stove}''~\citep{pmlr-v270-yang25e}, by mapping linguistic descriptions into quantitative rewards or costs. While visual evaluation excels at verifying \emph{physical configurations}, language provides an editable and human-readable interface for defining \emph{behavioral boundaries}. However, language inherently under-specifies exact geometric details (e.g., precise insertion angles), necessitating visual grounding for fine-grained verification.}

\subsubsection{\textcolor{black}{Policy conditioning: Language instructions \emph{vs.} Perceptual grounding}} 

\textcolor{black}{As a condition for policy execution, language instructions serve as high-level behavior specifiers, whereas goal images and videos offer direct \textbf{perceptual grounding}. Image-conditioned policies are optimal for tasks where the objective is strictly defined by a target spatial configuration, effectively resolving geometric ambiguity. Video-conditioned imitation extends this capability by implicitly conveying complex spatio-temporal dynamics~\citep{11365992}, such as approach trajectories, contact timing, and velocity profiles, which are difficult for language to communicate precisely. For instance, BC-Z~\citep{jang2022bc} demonstrates transfer to holdout tasks by conditioning on diverse task specifications, including human videos and language, showing that multimodal conditioning can improve generalization within the evaluated task family.}

\textcolor{black}{Language-conditioned policies abstract away these low-level physical details to offer three distinct practical advantages:
\begin{itemize}
\item \textbf{Low-bandwidth specification:} Language delivers human intent compactly, eliminating the need to record or render new visual demonstrations for every task variation.
\item \textbf{Compositional generalization:} Language supports the systematic recombination of skills, objects, and spatial relations, a critical requirement for scaling to open-world environments~\citep{wang2024poco,zhou2023languageconditioned}.
\item \textbf{Interactive editability:} Language allows for real-time goal refinement and corrective interaction during execution, allowing human operators to steer the policy through verbal corrections~\citep{olaf}, or dialogue-based clarification~\citep{zhang-etal-2022-danli}, rather than re-collecting demonstrations or re-specifying visual goals.
\end{itemize}
These properties make language the ideal modality for user-friendly interfaces and scalable deployment. Nonetheless, because language frequently under-specifies spatial realities, hybrid conditioning (fusing language embeddings with goal images or visual keyframes) often yields the most robust policies~\citep{jang2022bc,mees2022matters}.}

\subsubsection{\textcolor{black}{Planning: Language-based decomposition \emph{vs.} Visual imitation}} 

\textcolor{black}{In Section \ref{sec:language-cognitive-planning-reasoning}, we discussed how language serves as an ``internal reasoning medium'' for planning, enabling task decomposition~\citep{wei2022chain}, strategy formulation~\citep{silver2022learning}, and open-loop planning with LLMs~\citep{wang2023describe, rana2023sayplan}. In contrast, video-conditioned methods utilize visual demonstrations as dense \textbf{kinematic and structural priors}. A continuous video implicitly contains the necessary subgoals and physical transitions required for long-horizon tasks. One-shot visual imitation methods~\citep{lu2025visual} capitalize on this by extracting multi-stage operational plans directly from demonstrations, bypassing the need for explicit symbolic planners. This is particularly effective for tasks requiring complex physical manipulation that is difficult to articulate linguistically (e.g., folding cloth).}

\textcolor{black}{Language-conditioned planning, however, excels in scenarios that are \textbf{branching, constraint-rich, or highly interactive}. While video demonstrations typically show a single successful path, language naturally encodes causal logic and conditional behaviors (e.g., \emph{``if the drawer is stuck, pull harder; otherwise, close it gently''}). Furthermore, language acts as a bridge to high-level reasoning modules, allowing systems to decompose tasks logically, propose alternative plans, and query users for clarification~\citep{bartoli2022knowledge}. The key distinction lies in the abstraction level: visual demonstrations define the specific \emph{physical path}, while language organizes the \emph{logical structure}. To bridge the gap between high-level logic and low-level execution, effective systems often anchor linguistic subgoals to visual keyframes.}

\subsubsection{\textcolor{black}{Summary}}

\textcolor{black}{Overall, image and video-conditioned manipulation provide stronger \textbf{perceptual grounding}, effectively conveying precise \textbf{geometric configurations} and \textbf{implicit manipulation skills}. Conversely, language-conditioned manipulation offers unmatched \textbf{text semantic expressiveness} for constraints, \textbf{compositionality} for open-vocabulary generalization, and \textbf{interactive editability} for non-expert human collaboration.}

\textcolor{black}{Rather than viewing these as competing paradigms, the contemporary consensus points toward a unified multimodal task-specification framework: utilize language as the primary interface to dictate \emph{what} to do (e.g., high-level goals, logical constraints, and user preferences), while using image/video conditioning to specify \emph{how precisely} to execute it (e.g., spatial geometry, kinematics, and contact-rich dynamics). This distinction motivates the development of hybrid VLA architectures that fuse semantic linguistic instructions with visual demonstrations, enabling robust and adaptable manipulation in unstructured environments~\citep{pmlr-v305-black25a,li2026matters,torne2026mem}.}

\textcolor{black}{
\section{Discussion}
\label{sec:discussion}
This section discusses three heat debates in the community: Are large-scale vision-language-action models the most direct path to generalization? Can learned world models provide the necessary foresight for robust planning? And what are the trade-offs between model scaling and the strict real-time constraints of physical control? While these topics are relevant across robotics, they are particularly critical in language-conditioned robot manipulation. Here, language must be grounded in perception to inform physical interaction, creating a tightly coupled system where a failure in one domain may cascade and cause total task failure. Therefore, these advanced techniques bring new capabilities but also introduce challenges that should be addressed to achieve reliable and generalizable robot manipulation.}

\textcolor{black}{\subsection{Are VLAs the right path forward?}
The question of whether VLA models represent the right direction for robotics ultimately depends on whether the scaling principles that have succeeded in language and vision domains can also extend to embodied AI. In pure language modeling, the assumptions of effectively unlimited, homogeneous data and strong structural regularities make scaling laws viable: larger models and more data yield consistent improvements~\citep{achiam2023gpt,hurst2024gpt}.
However, this assumption becomes questionable when extended to robotic learning, where vision, language, and action must be tightly fused to enable embodied understanding and control.}

\textcolor{black}{
On the one hand, current VLA models are still smaller and less data-rich than modern foundation models in computer vision or natural language processing. 
While today’s LLMs usually contain over 100B parameters and are trained on trillions of tokens, most existing VLAs remain under 7B parameters and rely on only a few million robot trajectories~\citep{kim2024openvlaopensourcevisionlanguageactionmodel,open_x_embodiment_rt_x_2023}. 
As a result, it is still uncertain whether scaling up model size or dataset volume alone will yield meaningful improvements in robotic manipulation performance.
On the other hand, robotic learning is different from pure language or vision tasks because perception, reasoning, and control are all tightly linked to the physical world. 
Even a small error in the physical world can lead to task failure, whereas minor ambiguities in vision or language tasks rarely have catastrophic consequences.
Moreover, robotic data arises from physical interactions that are highly diverse, costly to collect, and difficult to reproduce, thus biases in data cannot be compensated for simply by collecting more samples.}

\textcolor{black}{
Consequently, an alternative direction for current VLA approaches is to operate under the paradigm of finite data with strong structural dependencies rather than relying purely on scale ~\citep{wagenmaker2025steering,kerr2025eye,parimi2025diffusion}.
In this setting, structure should guide scaling, ensuring that model capacity is grounded in physical and causal understanding.
Historically, advances in VLA have emerged from the expansion of task complexity rather than from model growth~\citep{open_x_embodiment_rt_x_2023,yang2025lohovla}. Moving from single-arm to dual-arm manipulation~\citep{gbagbe2024bi,li2025information}, from fixed-base to mobile platforms~\citep{wu2025momanipvla}, from rigid to deformable object handling~\citep{gao2025vla}, or from standalone manipulation to human-robot collaboration~\citep{xia2025robotic}, each introduces new structural dimensions and coordination mechanisms.
In this sense, genuine scaling in robotics lies in expanding the task manifold, namely the complexity, diversity, and structure of interactions, rather than merely increasing model parameters.
Empirical findings show that enlarging model scale alone does not yield consistent generalization improvements, as perception and control remain tightly coupled to embodiment and environment dynamics~\citep{lin2025datascalinglawsimitation}.
Looking forward, VLA frameworks should evolve beyond pure scaling toward task-driven structural induction, hybrid learning that integrates symbolic reasoning with physical priors, and modular architectures where perception, language, and action modules co-adapt coherently.
Whether such hybrid and structure-aware scaling can ultimately achieve the robustness and versatility of human-like embodied intelligence remains an open question.}

\textcolor{black}{\subsection{Are world models the right path forward?}
A world model is a learned predictor of environmental dynamics that enables a robot to ``imagine'' how states will evolve under candidate actions. The Dreamer family, e.g., DreamerV3~\citep{hafner2025dreamerv3}, shows that such imagination-based learning can yield strong sample efficiency and long-horizon behavior across diverse control domains (from visual game to continuous-control benchmarks) by optimizing policies in latent rollouts. Extending this idea to language, conditioning imagined rollouts on instructions turns free-form text into goal-aware futures that support compositional generalization, recombining known skills to follow new commands~\citep{aljalbout2025limt}. 
}

\textcolor{black}{
Beyond these, world models contribute to improved safety and generalization in language-conditioned robotics. Because candidate action sequences can be evaluated off-line in the model’s ``imagination'', the robot gains a capacity for what-if reasoning and safety filtering before execution. The system can proactively reject plans that appear risky or infeasible under the learned dynamics~\citep{wu2023daydreamer,huang2024safedreamer}, thereby avoiding hazardous trials in the real world. This ability to trial plans in simulation provides a measure of safety through imagination that purely reactive policies lack. Moreover, a world model encapsulates shared dynamics priors that aid generalization across tasks and contexts~\citep{ai2025review}. Knowledge of physical cause-and-effect learned in one scenario can be carried over to new objects~\citep{xie2019improvisation}, tasks, or even robotic embodiments, reducing the amount of task-specific experience needed for each new instruction.
}

\textcolor{black}{
However, there are notable drawbacks and challenges associated with world-model-based approaches. 
A primary concern is model bias and distributional brittleness. Because the world model is only an approximate dynamics model, small errors in contact, friction, or object properties can accumulate during imagined rollouts~\citep{hoque2022visuospatial}. When the system encounters out-of-distribution scenarios, it may propose semantically plausible plans (they satisfy the instruction) yet physically inconsistent (violating kinematics, stability, or contact constraints).
In practice, this means a plan that looks sound in simulation can fail when executed on the real robot. For example, due to subtle unmodeled dynamics or unseen environmental factors, such out-of-distribution errors show an inherent fragility: the world model’s predictions are  reliable within its training data's distribution. Ensuring the model’s validity in the real world, especially given the open-ended attribute of language instructions, remains an open problem. Robustness techniques and online adaptation may be required to mitigate these issues, but they add extra complexity to the system~\citep{chen2025vision}. 
Another downside of world models is the computational overhead and the difficulty of grounding imagined rollouts in real perception~\citep{DBLP:journals/corr/abs-2507-05169}. Training and deploying high-capacity generative models for dynamics prediction can be resource-intensive, straining real-time control loops.
}

\textcolor{black}{
In summary, rather than definitive drawbacks, current limitations reflect an early methodological stage. Open problems include: (i) injecting structured priors (physics, geometry, task constraints, and commonsense) into the learned dynamics, (ii) quantifying and propagating model uncertainty to mitigate bias and rollout drift, and (iii) meeting real‑time control deadlines with compute‑efficient models. } 

\textcolor{black}{\subsection{Does scaling help under real-time constraints?}
While scaling up models clearly improves representation quality and broadens the range of instructions they can handle, controlling physical robots is constrained by strict real-time deadlines. Each control cycle has a fixed time budget within which sensing, policy inference, and actuation must be completed. 
If a larger model increases inference time beyond this budget, the control loop can miss its deadline, leading to jitter and degraded stability, especially during contact-rich interactions. 
Using planner-based LLMs or VLMs for decision-making often adds noticeable latency (on the order of seconds) unless results are aggressively cached. 
Similarly, fully end-to-end VLA models tend to demand more computation time and memory per step on embedded hardware. 
Diffusion-based controllers also incur additional ``denoising tax'' due to iterative sampling, unless faster samplers or model distillation techniques are applied~\citep{lu2024manicm}. 
Offloading computation to the cloud can provide greater processing capacity, but it introduces unpredictable network delays and potential failure modes that are difficult to bound in safety-critical settings.
}

\textcolor{black}{
Consequently, simply using a ``bigger'' model is only helpful if its benefits can be realized within the strict timing deadlines of the control loop. 
In practice, robotics stacks often mitigate latency by compressing or distilling large neural network backbones and by pruning input tokens or model memory caches to reduce inference time. 
Many systems also favor multi-view 2D encoders (processing information from multiple camera views) over heavy 3D volumetric perception to keep the feedback loop fast. 
A split architecture is frequently effective: a lightweight onboard controller handles immediate feedback control and safety interlocks, while more computationally intensive deliberation (e.g., high-level planning or sub-goal generation) runs asynchronously or off-board when possible. 
We note that latency is not consistently treated as a first-class evaluation metric in the literature, and often it is mentioned only in footnotes. 
To assess real-world feasibility, we strongly encourage authors to report latency prominently as a primary performance metric.
}

\section{Limitations and future directions}
\label{sec:future_directions}
While vision and language models provide a strong foundation for language-conditioned robotic manipulation, several limitations remain for future research, such as \emph{generalization capability} and \emph{real-world safety}.

\subsection{Generalization capability}\label{sec:discuss-generalization}
The generalization capability of language-conditioned robot manipulation systems remains a central challenge for developing agents that can operate robustly across diverse, real-world scenarios~\citep{mon2025embodied}. 
Although current systems exhibit impressive performance within specific training domains ~\citep{wu2023tidybot,driess2023palme,black2024pi_0}, their performance often degrades when faced with unseen language instructions, novel objects, unfamiliar task compositions, dynamic environments, or new robot embodiments. 
This challenge reflects the tightly coupled but uneven generalization properties of language, perception, and control: language is compositional and ambiguous, while physical environments are diverse and nonstationary. As a result, the mapping from language to action may overfit to limited data distributions, leading to unexpected behaviors when the robot encounters new situations, object configurations, or embodiment-specific dynamics. More fundamentally, such failures stem from insufficient semantic grounding, data sparsity, weak mechanisms for the continual adaptation of knowledge, and the lack of aligned representations that transfer across tasks, spatial and temporal contexts, and varying robot dynamics.

To achieve robust generalization in language-conditioned robot manipulation, we argue that future research could focus on the following key aspects:
(i) \textbf{Data}: Developing large-scale, diverse, and better multimodally aligned robot manipulation datasets across language, perception, and action so as to enhance semantic grounding and coverage. 
(ii) \textbf{Lifelong learning}: Integrating lifelong learning frameworks that enable continual adaptation and knowledge retention across evolving tasks and linguistic contexts. 
(iii) \textbf{Cross-embodiment alignment}: Establishing cross-embodiment alignment mechanisms to bridge semantic and control representations across heterogeneous robot morphologies and dynamics. \textcolor{black}{Finally, these three aspects jointly affect the practical effectiveness of zero-shot capability.}

\textcolor{black}{\subsubsection{Language-conditioned datasets and evaluation}\label{sec:discuss-data}}
\textcolor{black}{
The generalization of language-conditioned robot manipulation heavily depends on the availability of diverse, language-annotated datasets and standardized benchmarks.}

\textcolor{black}{\textbf{Data availability}:
Training large-scale language-conditioned manipulation models inherently demands extensive datasets~\citep{open_x_embodiment_rt_x_2023,khazatsky2024droid,jiang2025galaxea}. 
While VLMs and LLMs benefit from web-scale data, collecting comparable volumes in the domain of robotic manipulation remains infeasible and highly labor-intensive.}
Some approaches~\citep{pmlr-v100-lynch20a,lynch2020language,mees2022matters,zhou2023languageconditioned,wang2023mimicplay} leverage play data instead of expert-labeled data to reduce labeling costs.
Unlabeled play data is often readily available in large quantities and can be acquired at a lower cost. 
It is also possible to gather play data by integrating computer gaming platforms, potentially providing access to data from millions of users. 
\citet{xiao2022robotic} utilize VLMs to augment the scenarios or instructions of video demonstration data. 
Nevertheless, relying solely on play data and generative VLMs may not be sufficient to train a robust FM for robot manipulation. 
We foresee the need for alternative techniques to gather large-scale demonstration data to enhance robot manipulation capabilities.

\textcolor{black}{\textbf{Beyond web-scale data}:}
Do we really need vast amounts of data to excel at manipulation and language understanding? 
\textcolor{black}{While web-scale datasets have fueled progress in FMs, human proficiency in manipulation arises from much smaller, context-rich experiences.}
Deep learning models are highly data-hungry, requiring massive amounts of information to generalize well. 
When there is limited data, deep learning models are more likely to overfit to spurious correlation patterns (short-cuts) that exist in the training set but do not generalize to real-world scenarios. 
This highlights a key limitation: relying solely on data-driven approaches may not be sufficient to solve manipulation and other complex tasks.
Neuro-symbolic approaches~\citep{DBLP:series/faia/BesoldGBBDHKLLPPPZ21}, which combine neural networks with symbolic reasoning, offer a promising alternative. 
These models can enhance data efficiency and reduce the need for massive datasets by integrating predefined human-readable knowledge, such as knowledge graphs or knowledge bases. This suggests that FMs trained on enormous data are not the only path to achieving intelligence. Exploring how to leverage limited data to train models with better generalization capabilities is an important direction.

\textcolor{black}{\textbf{Benchmarking}:}
Quantifying generalization in real-world settings is challenging, as model performance strongly depends on the underlying hardware specifications and physical constraints of the robotic platform.
For this reason, most of the approaches evaluate their results in simulators and lack a unified, standardized real-world evaluation benchmark.
\textcolor{black}{
However, strong performance in simulation does not necessarily translate to real-world success, as the sim-to-real gap arises from discrepancies in visual appearance, sensor noise, physical dynamics, and the interpretation of linguistic context between virtual and physical environments.
To reduce the gap, a promising approach is the \textit{real-to-sim-to-real}~\citep{torne2024reconciling}, where real-world interaction data are used to calibrate simulators that closely replicate physical environments. 
Policies trained in these aligned simulations are then redeployed in the real world, forming a feedback loop that enhances transfer robustness and consistency across domains. 
In parallel, establishing a standardized, language-conditioned real-world benchmark is essential for fair and reproducible evaluation. 
Such a benchmark should integrate natural-language instructions, multimodal observations, and diverse robot embodiments to assess both task success, semantic understanding, and real-time performance, ultimately advancing grounded language-conditioned manipulation.}

\textcolor{black}{\subsubsection{Lifelong learning}\label{sec:discuss-lifelong}
Lifelong learning represents a transformative direction for adaptive, continuously improving robotic agents. 
Unlike recent episodic learning, which assumes fixed datasets and tasks, lifelong learning allows agents to incrementally acquire new skills while avoiding catastrophic forgetting of prior ones. 
In language-conditioned contexts, this means not only adapting to new tasks but also maintaining semantic consistency across evolving linguistic instructions. Integrating replay buffers~\citep{chaudhry2019efficient}, modular architectures~\citep{mallya2018packnet}, and progressive neural expansion strategies~\citep{kirkpatrick2017overcoming} enables robots to preserve earlier knowledge while incorporating new behaviors, leading to more scalable, context-aware generalization. 
Future work should explore how lifelong learning can interface with FMs and policy distillation in language-conditioned robotic manipulation, forming persistent, adaptive intelligence capable of reasoning over both new language goals and prior embodied experiences.}

\subsubsection{Cross-embodiment alignment}\label{sec:discuss-crossembodiment}

\textcolor{black}{Language is inherently embodiment-agnostic~\citep{wang2024scaling}, as the same instruction, such as ``\textit{pick up the block}'', can be given to robots with very different kinematic structures and sensing configurations. However, robots differ significantly in perception systems, actuation mechanisms, and control dynamics. As a result, policies trained on one robot platform (e.g., Franka Panda) may fail when transferred to another robot (e.g., KUKA iiwa) even when the same language instruction is provided.}

\textcolor{black}{From the perspective of the taxonomy introduced in Section \ref{sec:Taxonomy}, the embodiment gap affects multiple stages of language-conditioned manipulation systems.} \textcolor{black}{First, \textbf{at the perception level}, language instructions must be grounded in observations collected by different sensor configurations. Robots may rely on different camera placements, sensing modalities, or workspace geometries, which can lead to inconsistent visual grounding of the same language command~\citep{khazatsky2024droid}. Consequently, perception modules must learn representations that can align semantic language descriptions with heterogeneous sensory inputs.}

\textcolor{black}{Second, \textbf{at the control level}, policies conditioned on language instructions must adapt to different kinematic structures and action spaces. A manipulation policy learned on one robot often implicitly encodes embodiment-specific dynamics. When transferred to another robot with different degrees of freedom, joint limits, or end-effector configurations, the learned policy may no longer produce valid actions~\citep{black2024pi_0}. Cross-embodiment alignment, therefore, requires mechanisms that map shared semantic goals to embodiment-specific control strategies.}

\textcolor{black}{
Third, \textbf{at the planning and reasoning level}, robots must maintain consistent semantic interpretation of language instructions while executing tasks using different embodiments. High-level reasoning modules may generate task plans such as object rearrangement or multi-step manipulation sequences~\citep{kim2024openvlaopensourcevisionlanguageactionmodel}. Ensuring that such plans remain executable across robots with different capabilities requires bridging the gap between abstract language reasoning and embodiment-specific constraints.}

\textcolor{black}{
To address these challenges, cross-embodiment learning aims to align shared semantic representations across heterogeneous robot morphologies. Recent work, such as HPT~\citep{wang2024scaling} and embodiment-aware dynamics modeling~\citep{schaldenbrand2023-icra, schaldenbrand24-cofrida} demonstrates that learning shared latent spaces across robots can significantly reduce the adaptation effort required when transferring policies to new embodiments. Developing such unified representations will be essential for enabling language-conditioned manipulation systems to generalize across diverse robotic platforms and environments.}

\textcolor{black}{\subsubsection{Effectiveness of zero-shot capability}}

\textcolor{black}{In language-conditioned robot manipulation, zero-shot capability is not a uniform property but depends on what is being generalized. Current methods show the strongest zero-shot behavior at the semantic and planning levels, where pre-trained language or vision-language models can interpret unseen instructions, retrieve relevant skills, propose subgoals, or construct reward/cost signals without task-specific retraining. For example, ZSRM~\citep{pmlr-v162-mahmoudieh22a} derives rewards by matching goal text with images through CLIP, thereby enabling zero-shot instruction-to-reward transfer, although its performance is fundamentally limited by CLIP's weak spatial reasoning~\citep{radford2021learning}. At the planning level, SayCan~\citep{ahn2022can} translates natural-language commands into semantically relevant and physically feasible skill sequences by combining LLM reasoning with affordance scores, while Code as Policies~\citep{liang2023code} can flexibly recompose API calls for previously unseen objects. However, such zero-shot planning remains fragile, since SayCan assumes faultless skill execution and Code-as-Policies may generate infeasible steps or even call nonexistent functions.}

\textcolor{black}{At the policy level, zero-shot generalization is more limited and usually depends on recombining previously learned skills or leveraging broad pre-trained representations. For instance, Lancon-learn~\citep{silva2021lancon} uses language to reweight reusable skill modules and can recombine them for zero-shot task generalization, but the joint learning of routing and control is fragile and sample-hungry. Similarly, RT-2~\citep{brohan2023rt} fine-tunes pre-trained vision-language representations for robot control and achieves better zero-shot performance on novel objects and instructions than scratch-trained policies. Nevertheless, these gains remain strongly tied to the coverage of the pre-training distribution and the availability of embodied robot data.}

\textcolor{black}{As the problem moves closer to physical execution, zero-shot capability weakens further. Robust zero-shot execution remains difficult for long-horizon, contact-rich, dynamic, or cross-embodiment tasks, where errors in grounding, perception, dynamics, and feedback accumulation become dominant. Open-loop planners such as SayCan~\citep{ahn2022can} often struggle in new environments because they lack embodied feedback, whereas closed-loop systems such as SayPlan~\citep{rana2023sayplan} improve robustness through replanning and textual scene feedback, but their performance is still limited by the reasoning ability and latency of the underlying LLM. More fundamentally, as discussed in our section on cross-embodiment alignment, even the same instruction can fail to transfer across robots with different sensors, kinematics, and control dynamics. Therefore, the practical effectiveness of zero-shot capability in current systems should be understood as \emph{partial and conditional} rather than universal: strongest for semantic specification and high-level planning, promising but still fragile for policy transfer, and still limited for reliable real-world manipulation under substantial novelty.}

\textcolor{black}{\subsection{Real-world safety} 
Ensuring real-world safety in language-conditioned robot manipulation is critical, especially as robots increasingly interact with humans in dynamic and unstructured environments.
We summarize three major concerns, namely ambiguity in language (Sec. \ref{sec:discuss-language}), failure recovery (Sec. \ref{sec:discuss-recovery}), and real-time performance (Sec. \ref{sec:discuss-realtime}).}

\subsubsection{Ambiguity in language}\label{sec:discuss-language}
One of the main safety concerns arises from language grounding and language use. Natural language instructions alone can be ambiguous. They are often underspecified and depend on rich context for a correct interpretation~\citep{clark1996using}. For instance, given the instruction ``\textit{Remove the chemicals from the table.}'', the user may intend for the robot to carefully relocate the containers of chemicals from the table to a designated storage area, ensuring safety by avoiding spills or accidents. However, this instruction might be interpreted as physically removing chemicals by tipping over or mishandling the containers, resulting in a hazardous situation with chemical spills and potential harm. \textcolor{black}{To mitigate this risk, robots need to understand the intent of human users and interpret the instructions based on shared experience and context. This often requires robots to take extra effort to proactively engage in clarification dialogue with humans to establish a common ground~\citep{chai2014collaborative}.} For instance, if instructed to ``\textit{Remove the chemicals from the table}'', the robot could respond with ``\textit{Do you want me to move the chemicals to storage, or should I dispose of them}?'' Furthermore, implementing feedback loops to confirm its interpretation before acting is essential. For example, the robot might state, ``\textit{I understood that I should carefully move the containers to the storage area. Is this correct?}'' This ensures clarity and prevents potential errors.

Despite recent progress, language grounding to 3D world still faces significant challenges~\citep{hong20233d, huang2024embodied}. Vision-language models are prone to hallucination~\citep{zhou2024analyzing,chen2024multi} and incapable of taking spatial perspectives when interpreting spatial expressions~\citep{zhang2025do}. Future work will need to find ways to improve language grounding, for example, through an agentic architecture that orchestrates various reasoning and perception modules~\citep{yang2024llm}, or through large-scale data synthesis~\citep{yang20253d} for training better 3D-LLMs or VLMs. 

\subsubsection{Recovering from failures}\label{sec:discuss-recovery}
\textcolor{black}{
Safety issues also emerge when robots fail during task execution due to software and hardware limitations.
On the software side, LLM hallucinations can lead to serious task failures, such as referencing nonexistent objects, producing logically inconsistent plans, or generating unsafe control code~\cite{rana2023sayplan,liang2023code,mu2023embodiedgpt}.
On the hardware side, edge-device inference limitations can delay actions~\citep{brohan2022rt}; motor overheating may trigger shutdowns in high-dynamic motions; depth sensors are prone to interference from lighting or reflective surfaces; and limited battery capacity restricts operating time in the wild.
To address these challenges, future research should focus on two complementary directions:
(1) At the software level, establishing closed-loop feedback mechanisms such as self-verification for LLM outputs and human-in-the-loop supervision~\citep{dai2024racer,bucker2024groundingrobotpoliciesvisuomotor,olaf,shi2024yell,meng2025growing} to detect and correct reasoning or planning errors in real time; and
(2) At the hardware level, advancing computational efficiency, thermal management, and sensor fusion to improve stability and reliability during continuous real-world operation.}

\subsubsection{Real-time performance}\label{sec:discuss-realtime}
\textcolor{black}{
Real-time responsiveness is another essential aspect of safe manipulation.
Large-scale models such as LLMs, VLMs, and VLAs often incur long inference times, which extend the control loop latency and hinder a robot’s ability to adapt to fast-changing environments.
Model compression methods, including pruning, quantization~\citep{DBLP:journals/ijon/LiangGWSZ21}, and knowledge distillation~\citep{DBLP:journals/ijcv/GouYMT21}, can accelerate inference while preserving accuracy.
Hybrid frameworks that delegate time-critical decisions to lightweight local models while offloading complex reasoning to larger models can balance responsiveness and intelligence.}

\textcolor{black}{Beyond on-device optimization, cloud-based inference offers a scalable path forward, where computationally intensive reasoning is performed remotely and distributed to multiple robots~\citep{brohan2022rt,brohan2023rt,rth2024rss}.
However, this introduces new latency and safety challenges.
For instance, in industrial settings, hundreds of robots may simultaneously query a shared cloud model, placing enormous pressure on network infrastructure.
Each agent typically requires control updates at frequencies exceeding 50 Hz~\citep{rth2024rss}, demanding not only high inference throughput on the cloud side but also ultra-fast data transmission and bus communication speeds to ensure timely feedback.
Even minor network delays or bandwidth bottlenecks can propagate through the control loop, resulting in unsafe behavior or degraded task performance.
Moreover, communication security risks present another critical safety concern in cloud-connected language-conditioned robotic systems.
Risks include hijacking or maliciously tampering with user commands, altering the model’s output action patterns, or even infiltrating and corrupting the internal weights of large-scale models.
Such attacks could lead to unsafe or unpredictable robot behaviors, jeopardizing both human safety and system integrity.
Therefore, ensuring secure, encrypted communication protocols, robust authentication mechanisms, and integrity verification of transmitted data and model parameters is essential.
These measures not only protect robots from external interference but also help maintain the reliability, traceability, and real-time performance required for safe deployment of language-conditioned policy in unstructured environments.}

\section{Conclusion}
\label{sec:conclusion}
\textcolor{black}{In summary, this survey presents an overview of the current language-conditioned robot manipulation approaches. 
Our analysis focuses on the primary ways language is integrated into the robotic systems, namely \emph{language for state evaluation}, \emph{language as a policy condition}, \emph{language for cognitive planning and reasoning}, and \emph{language in unified vision-language-action models}.
Furthermore, our comparative analysis offers a multifaceted perspective along five axes: \emph{action granularity}, \emph{data and supervision regimes}, \emph{system cost and latency}, \emph{environment and evaluation} and \emph{task representations}. 
We also examine the central debates in language-conditioned robot manipulation concerning VLAs, world models, and scaling. Finally, we articulate the key challenges and delineate future directions, with particular emphasis on \emph{generalization capability} and \emph{real-world safety}.}

\section{Acknowledgement}
The authors thank the International Max Planck Research School for Intelligent Systems (IMPRS-IS) for supporting Hongkuan Zhou. The authors also thank Yufei Duan for helpful insights with the design of Figure 2.

\titleformat{\section}{\sffamily\large\bfseries}{Appendix}{1em}{}
\appendix
\section{}
\subsection*{A. Tabular comparison of vision-language-action models}
\label{sec:apxa}
\begin{table*}[ht!]
\centering
\caption{\textcolor{black}{Classification of VLA models and design dimensions in Section \ref{sec:vla}.}}
\begin{threeparttable}
\resizebox{\textwidth}{!}{%
\begin{tabular}{ l l c c c c c c c c c c c c c c c c }

\toprule
\multicolumn{1}{c}{Optimization} & \multicolumn{1}{c}{Article} & \multicolumn{1}{c}{Time} & \multicolumn{2}{c}{Observation} & \multicolumn{4}{c}{Action} & \multicolumn{3}{c}{VLM policy} & \multicolumn{2}{c}{Pretraining} & \multicolumn{3}{c}{Scenarios} \\
 \cmidrule(lr){2-2} \cmidrule(lr){3-3} \cmidrule(lr){4-5} \cmidrule(lr){10-12} \cmidrule(lr){13-14} \cmidrule(lr){15-17}
\multicolumn{1}{c}{Direction} &  &  &  &  & \multicolumn{4}{c}{Generation} & CoT & FP & MEM & MD & CE & MS & RW & CE \\

\midrule

\cellcolor{Orchid!40}I: Data Source \& & EgoVLA~(\citeauthor{yang2025egovla}) & 2025-07 & \multicolumn{2}{c}{RGB, ROB, TX} & \multicolumn{4}{c}{DP} & \textcolor{Salmon}{\xmark} & \textcolor{Salmon}{\xmark} & \textcolor{JungleGreen}{\cmark} & \textcolor{JungleGreen}{\cmark} & \textcolor{JungleGreen}{\cmark} & \textcolor{JungleGreen}{\cmark} & \textcolor{Salmon}{\xmark} & \textcolor{JungleGreen}{\cmark} \\
\cellcolor{Orchid!40}Augmentation & H-RDT~(\citeauthor{bi2025h}) & 2025-08 & \multicolumn{2}{c}{RGB, ROB, TX} & \multicolumn{4}{c}{FM} & \textcolor{Salmon}{\xmark} & \textcolor{Salmon}{\xmark} & \textcolor{Salmon}{\xmark} & \textcolor{JungleGreen}{\cmark} & \textcolor{JungleGreen}{\cmark} & \textcolor{JungleGreen}{\cmark} & \textcolor{JungleGreen}{\cmark} & \textcolor{JungleGreen}{\cmark} \\
\cellcolor{Orchid!40} & Shortcut~(\citeauthor{xing2025shortcut}) & 2025-08 & \multicolumn{2}{c}{RGB, TX} & \multicolumn{4}{c}{-} & \textcolor{Salmon}{\xmark} & \textcolor{Salmon}{\xmark} & \textcolor{Salmon}{\xmark} & \textcolor{JungleGreen}{\cmark} & \textcolor{JungleGreen}{\cmark} & \textcolor{JungleGreen}{\cmark} & \textcolor{JungleGreen}{\cmark} & \textcolor{Salmon}{\xmark} \\

 \midrule

\cellcolor{Thistle!40}I: Spatial & SpatialVLA~(\citeauthor{qu2025spatialvla}) & 2025-01 & \multicolumn{2}{c}{RGB, ROB, TX} & \multicolumn{4}{c}{AR} & \textcolor{Salmon}{\xmark} & \textcolor{Salmon}{\xmark} & \textcolor{Salmon}{\xmark} & \textcolor{JungleGreen}{\cmark} & \textcolor{JungleGreen}{\cmark} & \textcolor{JungleGreen}{\cmark} & \textcolor{JungleGreen}{\cmark} & \textcolor{JungleGreen}{\cmark} \\
\cellcolor{Thistle!40}Understanding & PointVLA~(\citeauthor{li2025pointvla}) & 2025-05 & \multicolumn{2}{c}{RGBD, ROB, TX} & \multicolumn{4}{c}{DM} & \textcolor{Salmon}{\xmark} & \textcolor{Salmon}{\xmark} & \textcolor{Salmon}{\xmark} & \textcolor{JungleGreen}{\cmark} & \textcolor{JungleGreen}{\cmark} & \textcolor{JungleGreen}{\cmark} & \textcolor{JungleGreen}{\cmark} & \textcolor{JungleGreen}{\cmark} \\
\cellcolor{Thistle!40} & BridgeVLA~(\citeauthor{li2025bridgevla}) & 2025-06 & \multicolumn{2}{c}{RGB, ROB, TX} & \multicolumn{4}{c}{DP} & \textcolor{Salmon}{\xmark} & \textcolor{Salmon}{\xmark} & \textcolor{Salmon}{\xmark} & \textcolor{Salmon}{\xmark} & \textcolor{Salmon}{\xmark} & \textcolor{JungleGreen}{\cmark} & \textcolor{JungleGreen}{\cmark} & \textcolor{Salmon}{\xmark} \\
\cellcolor{Thistle!40} & GeoVLA~(\citeauthor{sun2025geovla}) & 2025-08 & \multicolumn{2}{c}{RGBD, ROB, TX} & \multicolumn{4}{c}{DM} & \textcolor{Salmon}{\xmark} & \textcolor{Salmon}{\xmark} & \textcolor{Salmon}{\xmark} & \textcolor{Salmon}{\xmark} & \textcolor{Salmon}{\xmark} & \textcolor{JungleGreen}{\cmark} & \textcolor{JungleGreen}{\cmark} & \textcolor{Salmon}{\xmark} \\

\midrule

\cellcolor{Lavender!40}I: Multimodal & VTLA~(\citeauthor{zhang2025vtla}) & 2025-05 & \multicolumn{2}{c}{RGB, TX} & \multicolumn{4}{c}{AR} & \textcolor{Salmon}{\xmark} & \textcolor{Salmon}{\xmark} & \textcolor{JungleGreen}{\cmark} & \textcolor{Salmon}{\xmark} & \textcolor{Salmon}{\xmark} & \textcolor{Salmon}{\xmark} & \textcolor{JungleGreen}{\cmark} & \textcolor{Salmon}{\xmark} \\
\cellcolor{Lavender!40} Sensing \& & Tactile-VLA~(\citeauthor{huang2025tactile}) & 2025-07 & \multicolumn{2}{c}{RGB, ROB, TX} & \multicolumn{4}{c}{FM} & \textcolor{JungleGreen}{\cmark} & \textcolor{Salmon}{\xmark} & \textcolor{Salmon}{\xmark} & \textcolor{Salmon}{\xmark} & \textcolor{Salmon}{\xmark} & \textcolor{JungleGreen}{\cmark} & \textcolor{Salmon}{\xmark} & \textcolor{Salmon}{\xmark} \\
\cellcolor{Lavender!40} Fusion & OmniVTLA~(\citeauthor{cheng2025omnivtla}) & 2025-08 & \multicolumn{2}{c}{RGB, ROB, TX} & \multicolumn{4}{c}{FM} & \textcolor{Salmon}{\xmark} & \textcolor{Salmon}{\xmark} & \textcolor{JungleGreen}{\cmark} & \textcolor{Salmon}{\xmark} & \textcolor{Salmon}{\xmark} & \textcolor{Salmon}{\xmark} & \textcolor{JungleGreen}{\cmark} & \textcolor{Salmon}{\xmark} \\
\cellcolor{Lavender!40} & ForceVLA~(\citeauthor{yu2025forcevla}) & 2025-09 & \multicolumn{2}{c}{RGBD, ROB, TX} & \multicolumn{4}{c}{FM} & \textcolor{JungleGreen}{\cmark} & \textcolor{Salmon}{\xmark} & \textcolor{Salmon}{\xmark} & \textcolor{Salmon}{\xmark} & \textcolor{Salmon}{\xmark} & \textcolor{JungleGreen}{\cmark} & \textcolor{JungleGreen}{\cmark} & \textcolor{Salmon}{\xmark} \\
\cellcolor{Lavender!40} & FuSe VLA~(\citeauthor{jones24fuse}) & 2025-01 & \multicolumn{2}{c}{RGBD, ROB, TX} & \multicolumn{4}{c}{AR} & \textcolor{Salmon}{\xmark} & \textcolor{Salmon}{\xmark} & \textcolor{Salmon}{\xmark} & \textcolor{Salmon}{\xmark} & \textcolor{Salmon}{\xmark} & \textcolor{JungleGreen}{\cmark} & \textcolor{JungleGreen}{\cmark} & \textcolor{Salmon}{\xmark} \\

\midrule

\cellcolor{Dandelion!40}II: Long-horizon & LoHoVLA~(\citeauthor{yang2025lohovla}) & 2025-05 & \multicolumn{2}{c}{RGB, ROB, TX} & \multicolumn{4}{c}{AR} & \textcolor{JungleGreen}{\cmark} & \textcolor{Salmon}{\xmark} & \textcolor{Salmon}{\xmark} & \textcolor{Salmon}{\xmark} & \textcolor{Salmon}{\xmark} & \textcolor{JungleGreen}{\cmark} & \textcolor{Salmon}{\xmark} & \textcolor{Salmon}{\xmark} \\
\cellcolor{Dandelion!40} task solving & Long-VLA~(\citeauthor{fan2025long}) & 2025-08 & \multicolumn{2}{c}{RGB, TX} & \multicolumn{4}{c}{DM} & \textcolor{Salmon}{\xmark} & \textcolor{Salmon}{\xmark} & \textcolor{Salmon}{\xmark} & \textcolor{JungleGreen}{\cmark} & \textcolor{Salmon}{\xmark} & \textcolor{JungleGreen}{\cmark} & \textcolor{JungleGreen}{\cmark} & \textcolor{Salmon}{\xmark} \\
\cellcolor{Dandelion!40}& DexVLA~(\citeauthor{wen2025dexvla}) & 2025-08 & \multicolumn{2}{c}{RGB, ROB, TX} & \multicolumn{4}{c}{DM} & \textcolor{JungleGreen}{\cmark} & \textcolor{Salmon}{\xmark} & \textcolor{Salmon}{\xmark} & \textcolor{JungleGreen}{\cmark} & \textcolor{JungleGreen}{\cmark} & \textcolor{JungleGreen}{\cmark} & \textcolor{JungleGreen}{\cmark} & \textcolor{JungleGreen}{\cmark} \\
\cellcolor{Dandelion!40}& MemoryVLA~(\citeauthor{shi2025memoryvla}) & 2025-08 & \multicolumn{2}{c}{RGB, TX} & \multicolumn{4}{c}{DM} & \textcolor{JungleGreen}{\cmark} & \textcolor{Salmon}{\xmark} & \textcolor{JungleGreen}{\cmark} & \textcolor{JungleGreen}{\cmark} & \textcolor{JungleGreen}{\cmark} & \textcolor{JungleGreen}{\cmark} & \textcolor{JungleGreen}{\cmark} & \textcolor{JungleGreen}{\cmark} \\

\midrule

\cellcolor{YellowOrange!40}II: Knowledge & DiffusionVLA~(\citeauthor{wen2025diffusionvla}) & 2024-12 & \multicolumn{2}{c}{RGBD, TX} & \multicolumn{4}{c}{DM + AR} &  \textcolor{JungleGreen}{\cmark} & \textcolor{Salmon}{\xmark} & \textcolor{Salmon}{\xmark} & \textcolor{JungleGreen}{\cmark} & \textcolor{Salmon}{\xmark} & \textcolor{JungleGreen}{\cmark} & \textcolor{JungleGreen}{\cmark} & \textcolor{JungleGreen}{\cmark} \\
\cellcolor{YellowOrange!40} Preserving & ChatVLA~(\citeauthor{zhou2025chatvla}) & 2025-02 & \multicolumn{2}{c}{RGB, TX} &  \multicolumn{4}{c}{DM} & \textcolor{Salmon}{\xmark} & \textcolor{Salmon}{\xmark} & \textcolor{Salmon}{\xmark} & \textcolor{JungleGreen}{\cmark} & \textcolor{Salmon}{\xmark} & \textcolor{JungleGreen}{\cmark} & \textcolor{JungleGreen}{\cmark} & \textcolor{Salmon}{\xmark} \\
\cellcolor{YellowOrange!40} & ChatVLA-2~(\citeauthor{zhou2025vision}) & 2025-05 & \multicolumn{2}{c}{RGB, TX} &  \multicolumn{4}{c}{DM} &  \textcolor{JungleGreen}{\cmark} & \textcolor{Salmon}{\xmark}  & \textcolor{Salmon}{\xmark} & \textcolor{JungleGreen}{\cmark} & \textcolor{Salmon}{\xmark} & \textcolor{JungleGreen}{\cmark} & \textcolor{JungleGreen}{\cmark} & \textcolor{Salmon}{\xmark} \\
\cellcolor{YellowOrange!40} & Insulating~(\citeauthor{driess2025knowledge}) & 2025-05 & \multicolumn{2}{c}{RGB, ROB, TX} & \multicolumn{4}{c}{FM + AR} & \textcolor{Salmon}{\xmark} & \textcolor{Salmon}{\xmark} & \textcolor{Salmon}{\xmark} & \textcolor{JungleGreen}{\cmark} & \textcolor{JungleGreen}{\cmark} & \textcolor{JungleGreen}{\cmark} & \textcolor{JungleGreen}{\cmark} & \textcolor{JungleGreen}{\cmark} \\
\cellcolor{YellowOrange!40} & InstructVLA~(\citeauthor{yang2025instructvla}) & 2025-07 & \multicolumn{2}{c}{RGB, ROB, TX} & \multicolumn{4}{c}{FM} & \textcolor{JungleGreen}{\cmark} & \textcolor{Salmon}{\xmark} & \textcolor{JungleGreen}{\cmark} & \textcolor{JungleGreen}{\cmark} & \textcolor{Salmon}{\xmark} & \textcolor{JungleGreen}{\cmark} & \textcolor{JungleGreen}{\cmark} & \textcolor{JungleGreen}{\cmark} \\
\cellcolor{YellowOrange!40} & GR-3~(\citeauthor{cheang2025gr}) & 2025-07 & \multicolumn{2}{c}{RGB, ROB, TX} & \multicolumn{4}{c}{FM} & \textcolor{Salmon}{\xmark} & \textcolor{Salmon}{\xmark} & \textcolor{JungleGreen}{\cmark} & \textcolor{JungleGreen}{\cmark} & \textcolor{Salmon}{\xmark} & \textcolor{JungleGreen}{\cmark} & \textcolor{JungleGreen}{\cmark} & \textcolor{Salmon}{\xmark} \\
 \midrule

\cellcolor{Orange!40}II: Reasoning \& & Seer~(\citeauthor{tian2025predictive}) & 2024-12 & \multicolumn{2}{c}{RGB, ROB, TX} & \multicolumn{4}{c}{DP} & \textcolor{Salmon}{\xmark} & \textcolor{JungleGreen}{\cmark} & \textcolor{Salmon}{\xmark} & \textcolor{JungleGreen}{\cmark} & \textcolor{JungleGreen}{\cmark} & \textcolor{JungleGreen}{\cmark} & \textcolor{JungleGreen}{\cmark} & \textcolor{JungleGreen}{\cmark} \\
\cellcolor{Orange!40} World Models & CoT-VLA~(\citeauthor{zhao2025cot}) & 2025-05 & \multicolumn{2}{c}{RGB, ROB, TX} & \multicolumn{4}{c}{AR} & \textcolor{JungleGreen}{\cmark} & \textcolor{JungleGreen}{\cmark} & \textcolor{Salmon}{\xmark} & \textcolor{JungleGreen}{\cmark} & \textcolor{JungleGreen}{\cmark} & \textcolor{JungleGreen}{\cmark} & \textcolor{JungleGreen}{\cmark} & \textcolor{JungleGreen}{\cmark} \\
\cellcolor{Orange!40} & WorldVLA~(\citeauthor{cen2025worldvla}) & 2025-06 & \multicolumn{2}{c}{RGB, ROB, TX} & \multicolumn{4}{c}{AR} & \textcolor{Salmon}{\xmark} & \textcolor{JungleGreen}{\cmark} & \textcolor{Salmon}{\xmark} & \textcolor{Salmon}{\xmark} & \textcolor{Salmon}{\xmark} & \textcolor{Salmon}{\xmark} & \textcolor{Salmon}{\xmark} & \textcolor{Salmon}{\xmark} \\
\cellcolor{Orange!40} & DreamVLA~(\citeauthor{zhang2025dreamvla}) & 2025-08 & \multicolumn{2}{c}{RGB, ROB, TX} & \multicolumn{4}{c}{DM} & \textcolor{Salmon}{\xmark} & \textcolor{JungleGreen}{\cmark} & \textcolor{Salmon}{\xmark} & \textcolor{Salmon}{\xmark} & \textcolor{Salmon}{\xmark} & \textcolor{JungleGreen}{\cmark} & \textcolor{JungleGreen}{\cmark} & \textcolor{Salmon}{\xmark} \\
\cellcolor{Orange!40} & ECoT VLA~(\citeauthor{zawalski2024roboticcontrolembodiedchainofthought}) & 2024-07 & \multicolumn{2}{c}{RGB, TX} & \multicolumn{4}{c}{AR} & \textcolor{JungleGreen}{\cmark} & \textcolor{JungleGreen}{\cmark} & \textcolor{Salmon}{\xmark} & \textcolor{JungleGreen}{\cmark} & \textcolor{Salmon}{\xmark} & \textcolor{JungleGreen}{\cmark} & \textcolor{JungleGreen}{\cmark} & \textcolor{Salmon}{\xmark} \\

\midrule
\cellcolor{Cerulean!40}III: Policy & PI-0~(\citeauthor{black2024pi_0}) & 2024-10 & \multicolumn{2}{c}{RGB, ROB, TX} & \multicolumn{4}{c}{FM} & \textcolor{Salmon}{\xmark} & \textcolor{Salmon}{\xmark} & \textcolor{Salmon}{\xmark} & \textcolor{JungleGreen}{\cmark} & \textcolor{JungleGreen}{\cmark} & \textcolor{JungleGreen}{\cmark} & \textcolor{JungleGreen}{\cmark} & \textcolor{JungleGreen}{\cmark} \\
\cellcolor{Cerulean!40} Execution & PI-Fast~(\citeauthor{pertsch2025fast}) & 2025-01 & \multicolumn{2}{c}{RGB, ROB, TX} & \multicolumn{4}{c}{FM} & \textcolor{Salmon}{\xmark} & \textcolor{Salmon}{\xmark} & \textcolor{Salmon}{\xmark} & \textcolor{JungleGreen}{\cmark} & \textcolor{JungleGreen}{\cmark} & \textcolor{JungleGreen}{\cmark} & \textcolor{JungleGreen}{\cmark} & \textcolor{JungleGreen}{\cmark} \\
\cellcolor{Cerulean!40} & PI-0.5~(\citeauthor{pmlr-v305-black25a}) & 2025-04 & \multicolumn{2}{c}{RGB, ROB, TX} & \multicolumn{4}{c}{FM} & \textcolor{JungleGreen}{\cmark} & \textcolor{Salmon}{\xmark} & \textcolor{Salmon}{\xmark} & \textcolor{JungleGreen}{\cmark} & \textcolor{JungleGreen}{\cmark} & \textcolor{JungleGreen}{\cmark} & \textcolor{JungleGreen}{\cmark} & \textcolor{JungleGreen}{\cmark} \\
\cellcolor{Cerulean!40} & DisDiffVLA~(\citeauthor{liang2025discrete}) & 2025-08 & \multicolumn{2}{c}{RGB, ROB, TX} & \multicolumn{4}{c}{DM} & \textcolor{Salmon}{\xmark} & \textcolor{Salmon}{\xmark} & \textcolor{Salmon}{\xmark} & \textcolor{JungleGreen}{\cmark} & \textcolor{JungleGreen}{\cmark} & \textcolor{JungleGreen}{\cmark} & \textcolor{Salmon}{\xmark} & \textcolor{JungleGreen}{\cmark} \\

 \midrule

\cellcolor{JungleGreen!40}IV: Adaptation \& & OpenVLA-OFT~(\citeauthor{kim2025fine}) & 2025-02 & \multicolumn{2}{c}{RGB, ROB, TX} & \multicolumn{4}{c}{DP} & \textcolor{Salmon}{\xmark} & \textcolor{Salmon}{\xmark} & \textcolor{Salmon}{\xmark} & \textcolor{JungleGreen}{\cmark} & \textcolor{JungleGreen}{\cmark} & \textcolor{JungleGreen}{\cmark} & \textcolor{JungleGreen}{\cmark} & \textcolor{JungleGreen}{\cmark} \\
\cellcolor{JungleGreen!40} Fine-Tuning & ConRFT~(\citeauthor{chen2025conrft}) & 2025-04 & \multicolumn{2}{c}{RGB, ROB, TX} & \multicolumn{4}{c}{DM} & \textcolor{Salmon}{\xmark} & \textcolor{Salmon}{\xmark} & \textcolor{Salmon}{\xmark} & \textcolor{Salmon}{\xmark} & \textcolor{Salmon}{\xmark} & \textcolor{Salmon}{\xmark} & \textcolor{JungleGreen}{\cmark} & \textcolor{Salmon}{\xmark} \\
\cellcolor{JungleGreen!40} & RIPT-VLA~(\citeauthor{tan2025interactive}) & 2025-05 & \multicolumn{2}{c}{RGB, ROB, TX} & \multicolumn{4}{c}{AR} & \textcolor{Salmon}{\xmark} & \textcolor{Salmon}{\xmark} & \textcolor{Salmon}{\xmark} & \textcolor{Salmon}{\xmark} & \textcolor{JungleGreen}{\cmark} & \textcolor{JungleGreen}{\cmark} & \textcolor{Salmon}{\xmark} & \textcolor{Salmon}{\xmark} \\
\cellcolor{JungleGreen!40} & ControlVLA~(\citeauthor{li2025controlvla}) & 2025-06 & \multicolumn{2}{c}{RGB, ROB, TX} & \multicolumn{4}{c}{DM} & \textcolor{Salmon}{\xmark} & \textcolor{Salmon}{\xmark} & \textcolor{Salmon}{\xmark} & \textcolor{Salmon}{\xmark} & \textcolor{Salmon}{\xmark} & \textcolor{Salmon}{\xmark} & \textcolor{JungleGreen}{\cmark} & \textcolor{Salmon}{\xmark} \\

\bottomrule
\end{tabular}}
    \begin{tablenotes}
        \footnotesize Observation: RGB images, D: depth information (e.g., point clouds), ROB: Robot Proprioception, TX: text (e.g., prompt, language goal). \\
        \footnotesize Action generation: (FM) Flow Matching, (DM) Diffusion Model, (AR) Autoregressive, (DP) Direct Prediction, (-) Not Applicable; \\
        \footnotesize (CoT) Chain-of-Thought, (FP) Future Prediction, (MEM) Memory Mechanisms, (MD) Multiple Datasets, \\ 
        \footnotesize (CE) Cross-embodiment Data, (MS) Multi-scenario, (RW) Real-world Deployment, (CE) Cross-embodiment Execution. \\
    \end{tablenotes}
\end{threeparttable}
\label{tab:vla_table}
\vspace{-1em}
\end{table*}

\begin{table*}[ht!]
    \centering
    \caption{Comparison of language-conditioned robot manipulation approaches I in Section \ref{sec:comparative_analysis}.}
    \begin{threeparttable}
    \begin{adjustbox}{width=\linewidth,center}
        \begin{tabular}{ l c c c c c c c c c c}
            \toprule
            Models & Years & Benchmark & Simulation Engine & \makecell{Language \\ Module} & \makecell{Perception \\ Module} & \makecell{Real world \\experiments} & FMs & \makecell{RL} & \makecell {IL} & \makecell {MP}\\
            \midrule
            IPRO (\citeauthor{hatori2018interactively}) & 2018 & \# & - & LSTM & CNN & \textcolor{JungleGreen}{\cmark} & \textcolor{Salmon}{\xmark} & \textcolor{Salmon}{\xmark} & \textcolor{Salmon}{\xmark} & \textcolor{JungleGreen}{\cmark} \\
            MaestROB (\citeauthor{munawar2018maestrob}) & 2018 & \# & - & IBM Watson & Artoolkit & \textcolor{JungleGreen}{\cmark} & \textcolor{Salmon}{\xmark} & \textcolor{Salmon}{\xmark} & \textcolor{Salmon}{\xmark} & \textcolor{JungleGreen}{\cmark}\\
            exePlan (\citeauthor{liu2018generating}) & 2018 & \# & - & coreNLP & * & \textcolor{JungleGreen}{\cmark} & \textcolor{Salmon}{\xmark} & \textcolor{Salmon}{\xmark} & \textcolor{Salmon}{\xmark} & \textcolor{JungleGreen}{\cmark}\\
            TLC (\citeauthor{wang2018facilitating}) & 2018 & \# & - & CCG & * & \textcolor{JungleGreen}{\cmark} & \textcolor{Salmon}{\xmark} & \textcolor{Salmon}{\xmark} & \textcolor{JungleGreen}{\cmark} & \textcolor{Salmon}{\xmark}\\
            IMNLU (\citeauthor{wang2018facilitating}) & 2018 & \# & - & WordNet, FrameNet & Stereo Vision & \textcolor{JungleGreen}{\cmark} & \textcolor{Salmon}{\xmark} & \textcolor{Salmon}{\xmark} & \textcolor{Salmon}{\xmark} & \textcolor{JungleGreen}{\cmark}\\
            \midrule
            Cut\&recombine (\citeauthor{tamosiunaite2019cut}) & 2019 & \# & - & Parser & * & \textcolor{JungleGreen}{\cmark} & \textcolor{Salmon}{\xmark} & \textcolor{Salmon}{\xmark} & \textcolor{Salmon}{\xmark} & \textcolor{JungleGreen}{\cmark}\\
            DREAMCELL (\citeauthor{paxton2019prospection}) & 2019 & \# & - & LSTM & * & \textcolor{Salmon}{\xmark} & \textcolor{Salmon}{\xmark} & \textcolor{Salmon}{\xmark} & \textcolor{JungleGreen}{\cmark} & \textcolor{Salmon}{\xmark} \\
            ICR (\citeauthor{8793667}) & 2019 & \# & - & Parser, DCG & YOLO9000 & \textcolor{JungleGreen}{\cmark} & \textcolor{Salmon}{\xmark} & \textcolor{Salmon}{\xmark} & \textcolor{Salmon}{\xmark} & \textcolor{JungleGreen}{\cmark}\\
            GroundedDA (\citeauthor{thomason2019improving}) & 2019 & \# & - & CCG & RANSAC & \textcolor{JungleGreen}{\cmark} & \textcolor{Salmon}{\xmark} & \textcolor{Salmon}{\xmark} & \textcolor{Salmon}{\xmark} & \textcolor{JungleGreen}{\cmark}\\
            \midrule
            MEC (\citeauthor{arkin2020multimodal}) & 2020 & \# & - & Parser, ADCG & Mask RCNN & \textcolor{JungleGreen}{\cmark} & \textcolor{Salmon}{\xmark} & \textcolor{Salmon}{\xmark} & \textcolor{Salmon}{\xmark} & \textcolor{JungleGreen}{\cmark}\\
            LMCR (\citeauthor{chen2020enabling}) & 2020 & \# & - &  RNN & Mask RCNN & \textcolor{JungleGreen}{\cmark} & \textcolor{Salmon}{\xmark} & \textcolor{Salmon}{\xmark} & \textcolor{Salmon}{\xmark} & \textcolor{JungleGreen}{\cmark}\\
            PixL2R (\citeauthor{goyal2021pixl2r})& 2020 & Meta-World & MuJoCo & LSTM & CNN & \textcolor{Salmon}{\xmark} & \textcolor{Salmon}{\xmark} & \textcolor{JungleGreen}{\cmark} & \textcolor{Salmon}{\xmark} & \textcolor{Salmon}{\xmark}\\
            Concept2Robot (\citeauthor{shao2021concept2robot}) &	2020 &	\# &	PyBullet & BERT &	ResNet-18 &	\textcolor{Salmon}{\xmark} & \textcolor{Salmon}{\xmark} & \textcolor{Salmon}{\xmark} & \textcolor{JungleGreen}{\cmark} & \textcolor{Salmon}{\xmark}\\
            LanguagePolicy(\citeauthor{stepputtis2020language}) & 2020 & \# & CoppeliaSim & GLoVe & Faster RCNN & \textcolor{Salmon}{\xmark} & \textcolor{Salmon}{\xmark} & \textcolor{Salmon}{\xmark} & \textcolor{JungleGreen}{\cmark} & \textcolor{Salmon}{\xmark}\\
            \midrule
            LOReL (\citeauthor{nair2022learning}) & 2021 & Meta-World & MuJoCo & distillBERT & CNN & \textcolor{JungleGreen}{\cmark} & \textcolor{Salmon}{\xmark} & \textcolor{JungleGreen}{\cmark} & \textcolor{Salmon}{\xmark} & \textcolor{Salmon}{\xmark}\\
            CARE (\citeauthor{pmlr-v139-sodhani21a}) & 2021 & Meta-World & MuJoCo & RoBERTa & * & \textcolor{Salmon}{\xmark} & \textcolor{JungleGreen}{\cmark} & \textcolor{JungleGreen}{\cmark} & \textcolor{Salmon}{\xmark} & \textcolor{Salmon}{\xmark}\\
            MCIL (\citeauthor{lynch2020language}) & 2021 & \# & MuJoCo & MUSE & CNN & \textcolor{Salmon}{\xmark} & \textcolor{Salmon}{\xmark} & \textcolor{Salmon}{\xmark} & \textcolor{JungleGreen}{\cmark} & \textcolor{Salmon}{\xmark}\\
            BC-Z (\citeauthor{jang2022bc}) & 2021 & \# & - & MUSE & ResNet18 & \textcolor{JungleGreen}{\cmark} & \textcolor{Salmon}{\xmark} & \textcolor{Salmon}{\xmark} & \textcolor{JungleGreen}{\cmark} & \textcolor{Salmon}{\xmark}\\
            CLIPort (\citeauthor{shridhar2022cliport}) & 2021 & \# & PyBullet & CLIP & CLIP & \textcolor{JungleGreen}{\cmark} & \textcolor{Salmon}{\xmark} & \textcolor{Salmon}{\xmark} & \textcolor{JungleGreen}{\cmark} & \textcolor{Salmon}{\xmark}\\
            \midrule
            LanCon-Learn (\citeauthor{silva2021lancon}) & 2022 & Meta-World & MuJoCo & GLoVe & * & \textcolor{Salmon}{\xmark} & \textcolor{Salmon}{\xmark} & \textcolor{JungleGreen}{\cmark} & \textcolor{JungleGreen}{\cmark} & \textcolor{Salmon}{\xmark}\\
            MILLION (\citeauthor{bing2023meta}) & 2022 & Meta-World & MuJoCo & GLoVe & * & \textcolor{JungleGreen}{\cmark} & \textcolor{Salmon}{\xmark} &\textcolor{JungleGreen}{\cmark} & \textcolor{Salmon}{\xmark} & \textcolor{Salmon}{\xmark}\\
            PaLM-SayCan (\citeauthor{ahn2022can}) & 2022 & \# & - & PaLM & ViLD & \textcolor{JungleGreen}{\cmark} & \textcolor{JungleGreen}{\cmark} & \textcolor{JungleGreen}{\cmark} & \textcolor{JungleGreen}{\cmark} & \textcolor{Salmon}{\xmark}\\
            ATLA (\citeauthor{ren2023leveraging}) & 2022 & \# & PyBullet & BERT-Tiny  & CNN & \textcolor{Salmon}{\xmark} & \textcolor{JungleGreen}{\cmark} & \textcolor{JungleGreen}{\cmark} & \textcolor{Salmon}{\xmark}& \textcolor{Salmon}{\xmark} \\
            HULC (\citeauthor{mees2022matters}) & 2022 & CALVIN & PyBullet & MiniLM-L3-v2 & CNN & \textcolor{Salmon}{\xmark} & \textcolor{Salmon}{\xmark} & \textcolor{Salmon}{\xmark} & \textcolor{JungleGreen}{\cmark} & \textcolor{Salmon}{\xmark}\\
            PerAct (\citeauthor{shridhar2023perceiver}) & 2022 & RLbench &CoppelaSim & CLIP & ViT & \textcolor{JungleGreen}{\cmark} & \textcolor{Salmon}{\xmark} & \textcolor{Salmon}{\xmark} & \textcolor{JungleGreen}{\cmark} & \textcolor{Salmon}{\xmark}\\
            RT-1 (\citeauthor{brohan2022rt})  & 2022 & \# & - & USE & EfficientNet-B3 & \textcolor{JungleGreen}{\cmark} & \textcolor{JungleGreen}{\cmark} & \textcolor{Salmon}{\xmark} & \textcolor{JungleGreen}{\cmark} & \textcolor{Salmon}{\xmark} \\ 
            LATTE (\citeauthor{bucker2023latte}) & 2023 & \# & CoppeliaSim & distillBERT, CLIP & CLIP & \textcolor{JungleGreen}{\cmark} & \textcolor{Salmon}{\xmark} &  \textcolor{Salmon}{\xmark} & \textcolor{Salmon}{\xmark}& \textcolor{JungleGreen}{\cmark}\\
            DIAL (\citeauthor{xiao2022robotic})  & 2022 & \# & - & CLIP & CLIP & \textcolor{JungleGreen}{\cmark} & \textcolor{JungleGreen}{\cmark} & \textcolor{Salmon}{\xmark} & \textcolor{JungleGreen}{\cmark} & \textcolor{Salmon}{\xmark}\\
            R3M (\citeauthor{nair2022r3m})  & 2022 & \# & - & distillBERT & ResNet18,34,50 & \textcolor{JungleGreen}{\cmark} & \textcolor{Salmon}{\xmark} & \textcolor{Salmon}{\xmark} & \textcolor{JungleGreen}{\cmark} & \textcolor{Salmon}{\xmark}\\
            Inner Monologue (\citeauthor{huang2022inner})  & 2022 & \# & - & CLIP & CLIP & \textcolor{JungleGreen}{\cmark} & \textcolor{JungleGreen}{\cmark} & \textcolor{Salmon}{\xmark} & \textcolor{Salmon}{\xmark} & \textcolor{JungleGreen}{\cmark}\\

            \midrule
            NLMap (\citeauthor{chen2023open})  & 2023 & \# & - & CLIP & ViLD & \textcolor{JungleGreen}{\cmark} & \textcolor{JungleGreen}{\cmark} & \textcolor{Salmon}{\xmark} & \textcolor{JungleGreen}{\cmark} & \textcolor{Salmon}{\xmark}\\ 
            Code as Policies (\citeauthor{liang2023code})  & 2023 & \# & - & GPT3, Codex & ViLD & \textcolor{JungleGreen}{\cmark} & \textcolor{JungleGreen}{\cmark} & \textcolor{Salmon}{\xmark} & \textcolor{Salmon}{\xmark} & \textcolor{JungleGreen}{\cmark}\\ 
            Progprompt (\citeauthor{singh:progprompt:ar})  & 2023 & Virtualhome & Unity3D & GPT-3 & * & \textcolor{JungleGreen}{\cmark} & \textcolor{JungleGreen}{\cmark} & \textcolor{Salmon}{\xmark} & \textcolor{Salmon}{\xmark} & \textcolor{JungleGreen}{\cmark}\\
            Language2Reward (\citeauthor{yu2023language}) & 2023 & \# & MuJoCo MPC & GPT-4 & * & \textcolor{JungleGreen}{\cmark} & \textcolor{JungleGreen}{\cmark} & \textcolor{JungleGreen}{\cmark} & \textcolor{Salmon}{\xmark} & \textcolor{Salmon}{\xmark}\\
            LfS (\citeauthor{yao2023learning})  & 2023 & Meta-World & MuJoCo & Cons. Parser  & * & \textcolor{JungleGreen}{\cmark} & \textcolor{Salmon}{\xmark} & \textcolor{JungleGreen}{\cmark} & \textcolor{Salmon}{\xmark} & \textcolor{Salmon}{\xmark}\\
            HULC++ (\citeauthor{mees23hulc2})& 2023 & CALVIN & PyBullet & MiniLM-L3-v2 & CNN & \textcolor{JungleGreen}{\cmark} & \textcolor{Salmon}{\xmark} & \textcolor{Salmon}{\xmark} & \textcolor{JungleGreen}{\cmark} & \textcolor{Salmon}{\xmark}\\
            ALOHA~(\citeauthor{Zhao-RSS-23}) & 2023 & \# & - & Transformer & CNN & \textcolor{JungleGreen}{\cmark} & \textcolor{Salmon}{\xmark} & \textcolor{Salmon}{\xmark} & \textcolor{JungleGreen}{\cmark} & \textcolor{Salmon}{\xmark}\\
            LEMMA (\citeauthor{gong2023lemma}) & 2023 & LEMMA & NVIDIA Omniverse & CLIP & CLIP & \textcolor{Salmon}{\xmark} & \textcolor{Salmon}{\xmark} & \textcolor{Salmon}{\xmark} & \textcolor{JungleGreen}{\cmark} & \textcolor{Salmon}{\xmark}\\
            SPIL (\citeauthor{zhou2023languageconditioned})& 2023 & CALVIN & PyBullet & MiniLM-L3-v2 & CNN & \textcolor{JungleGreen}{\cmark} & \textcolor{Salmon}{\xmark} & \textcolor{Salmon}{\xmark} & \textcolor{JungleGreen}{\cmark} & \textcolor{Salmon}{\xmark}\\
            PaLM-E (\citeauthor{driess2023palme}) & 2023 & \# & PyBullet & PaLM & ViT & \textcolor{JungleGreen}{\cmark} & \textcolor{JungleGreen}{\cmark} & \textcolor{Salmon}{\xmark} & \textcolor{JungleGreen}{\cmark} & \textcolor{Salmon}{\xmark}\\ 
            LAMP (\citeauthor{adeniji2023language}) & 2023 & RLbench & CoppelaSim  & ChatGPT & R3M & \textcolor{Salmon}{\xmark} & \textcolor{JungleGreen}{\cmark} & \textcolor{JungleGreen}{\cmark} & \textcolor{Salmon}{\xmark} & \textcolor{Salmon}{\xmark}\\
            MOO (\citeauthor{stone2023open}) & 2023 & \# & - & OWL-ViT & OWL-ViT & \textcolor{JungleGreen}{\cmark} & \textcolor{Salmon}{\xmark} & \textcolor{Salmon}{\xmark} & \textcolor{JungleGreen}{\cmark} & \textcolor{Salmon}{\xmark}\\
            Instruction2Act (\citeauthor{huang2023instruct2act}) & 2023 & VIMAbench & PyBullet & ChatGPT & CLIP & \textcolor{Salmon}{\xmark} & \textcolor{JungleGreen}{\cmark} &  \textcolor{Salmon}{\xmark} & \textcolor{Salmon}{\xmark} & \textcolor{JungleGreen}{\cmark}\\
            VoxPoser (\citeauthor{huang2023voxposer}) & 2023 & \# & SAPIEN & GPT-4 & OWL-ViT  & \textcolor{JungleGreen}{\cmark} & \textcolor{JungleGreen}{\cmark} & \textcolor{Salmon}{\xmark} & \textcolor{Salmon}{\xmark}& \textcolor{JungleGreen}{\cmark}\\ 
            SuccessVQA (\citeauthor{du2023visionlanguage}) & 2023 & \# & IA Playroom & Flamingo & Flamingo & \textcolor{JungleGreen}{\cmark} & \textcolor{JungleGreen}{\cmark} & \textcolor{Salmon}{\xmark} & \textcolor{JungleGreen}{\cmark}& \textcolor{Salmon}{\xmark}\\ 
            VIMA (\citeauthor{jiang2023vima}) & 2023 & VIMAbench & PyBullet & T5 & ViT & \textcolor{JungleGreen}{\cmark} & \textcolor{JungleGreen}{\cmark} & \textcolor{Salmon}{\xmark} & \textcolor{JungleGreen}{\cmark}& \textcolor{Salmon}{\xmark}\\ 
            TidyBot (\citeauthor{wu2023tidybot}) & 2023 & \# & - & GPT-3 & CLIP & \textcolor{JungleGreen}{\cmark} & \textcolor{JungleGreen}{\cmark} & \textcolor{Salmon}{\xmark} & \textcolor{Salmon}{\xmark}& \textcolor{JungleGreen}{\cmark}\\ 
            Text2Motion (\citeauthor{lin2023text2motion}) & 2023 & \# & - & GPT-3, Codex & * & \textcolor{JungleGreen}{\cmark} & \textcolor{JungleGreen}{\cmark} & \textcolor{JungleGreen}{\cmark} & \textcolor{Salmon}{\xmark}& \textcolor{Salmon}{\xmark}\\ 
            LLM-GROP (\citeauthor{ding2023task}) & 2023 & \# & Gazebo & GPT-3 & * & \textcolor{JungleGreen}{\cmark} & \textcolor{JungleGreen}{\cmark} & \textcolor{Salmon}{\xmark} & \textcolor{Salmon}{\xmark}& \textcolor{JungleGreen}{\cmark}\\ 
            Scaling Up (\citeauthor{ha2023scaling})  & 2023 & \# & MuJoCo & CLIP, GPT-3 & ResNet-18 & \textcolor{JungleGreen}{\cmark} & \textcolor{JungleGreen}{\cmark} & \textcolor{Salmon}{\xmark} & \textcolor{JungleGreen}{\cmark} & \textcolor{Salmon}{\xmark}\\ 
            Socratic Models (\citeauthor{zeng2022socratic}) & 2023 & \# & - & RoBERTa, GPT-3 & CLIP & \textcolor{JungleGreen}{\cmark} & \textcolor{JungleGreen}{\cmark} & \textcolor{Salmon}{\xmark} & \textcolor{Salmon}{\xmark}& \textcolor{JungleGreen}{\cmark}\\ 
            SayPlan (\citeauthor{rana2023sayplan})  & 2023 & \# & - & GPT-4 & * & \textcolor{JungleGreen}{\cmark} & \textcolor{JungleGreen}{\cmark} & \textcolor{Salmon}{\xmark} & \textcolor{Salmon}{\xmark} & \textcolor{JungleGreen}{\cmark}\\ 
             RT-2 (\citeauthor{brohan2023rt})  & 2023 & \# & - & PaLI-X, PaLM-E & PaLI-X, PaLM-E & \textcolor{JungleGreen}{\cmark} & \textcolor{JungleGreen}{\cmark} & \textcolor{Salmon}{\xmark} & \textcolor{JungleGreen}{\cmark} & \textcolor{Salmon}{\xmark}\\ 
              KNOWNO (\citeauthor{ren2023robots})  & 2023 & \# & PyBullet & PaLM-2L & * & \textcolor{JungleGreen}{\cmark} & \textcolor{JungleGreen}{\cmark} & \textcolor{Salmon}{\xmark} & \textcolor{Salmon}{\xmark} & \textcolor{JungleGreen}{\cmark}\\ 
              MDT(\citeauthor{reuss2024multimodal}) & 2023 & CALVIN & PyBullet &  CLIP & CLIP & \textcolor{Salmon}{\xmark} & \textcolor{Salmon}{\xmark} & \textcolor{Salmon}{\xmark} & \textcolor{JungleGreen}{\cmark}& \textcolor{Salmon}{\xmark}\\
              RT-Trajectory (\citeauthor{gu2023rttrajectory})  & 2023 & \# & - & PaLM-E & EfficientNet-B3 & \textcolor{JungleGreen}{\cmark} & \textcolor{JungleGreen}{\cmark} & \textcolor{Salmon}{\xmark} & \textcolor{JungleGreen}{\cmark} & \textcolor{Salmon}{\xmark} \\ 
              SuSIE(\citeauthor{black2023zero})&2023& CALVIN & PyBullet & InstructPix2Pix(GPT3) & InstructPix2Pix & \textcolor{JungleGreen}{\cmark} & \textcolor{JungleGreen}{\cmark} & \textcolor{Salmon}{\xmark} & \textcolor{JungleGreen}{\cmark}& \textcolor{Salmon}{\xmark}\\
              Playfusion(\citeauthor{chen2023playfusion})& 2023 & CALVIN & PyBullet & Sentence-bert & ResNet-18 & \textcolor{JungleGreen}{\cmark} & \textcolor{Salmon}{\xmark} & \textcolor{Salmon}{\xmark} & \textcolor{JungleGreen}{\cmark}& \textcolor{Salmon}{\xmark}\\
              ChainedDiffuser(\citeauthor{xian2023chaineddiffuser}) & 2023 & RLbench & CoppelaSim & CLIP &  CLIP & \textcolor{JungleGreen}{\cmark} & \textcolor{Salmon}{\xmark} & \textcolor{Salmon}{\xmark} & \textcolor{JungleGreen}{\cmark}& \textcolor{Salmon}{\xmark}\\
              GNFactor(\citeauthor{ze2023gnfactor})& 2023 & RLbench & CoppelaSim & CLIP & NeRF & \textcolor{JungleGreen}{\cmark} & \textcolor{Salmon}{\xmark} & \textcolor{Salmon}{\xmark} & \textcolor{JungleGreen}{\cmark}& \textcolor{Salmon}{\xmark}\\
              StructDiffusion(\citeauthor{liu2022structdiffusion}) & 2023 & \# & PyBullet & Sentence-bert & PCT & \textcolor{JungleGreen}{\cmark} & \textcolor{Salmon}{\xmark} & \textcolor{Salmon}{\xmark} & \textcolor{Salmon}{\xmark}& \textcolor{JungleGreen}{\cmark}\\
              \midrule
              PoCo(\citeauthor{wang2024poco})& 2024 & Fleet-Tools & Drake & T5 & ResNet-18 & \textcolor{JungleGreen}{\cmark} & \textcolor{Salmon}{\xmark} & \textcolor{Salmon}{\xmark} & \textcolor{JungleGreen}{\cmark} & \textcolor{Salmon}{\xmark}\\
              DNAct(\citeauthor{yan2024dnact})& 2024 & RLbench & CoppelaSim & CLIP & NeRF, PointNext & \textcolor{JungleGreen}{\cmark} & \textcolor{Salmon}{\xmark} & \textcolor{Salmon}{\xmark} & \textcolor{JungleGreen}{\cmark}& \textcolor{Salmon}{\xmark}\\
              3D Diffuser Actor(\citeauthor{ke20253d}) & 2024 & CALVIN & PyBullet & CLIP & CLIP & \textcolor{JungleGreen}{\cmark} & \textcolor{Salmon}{\xmark} & \textcolor{Salmon}{\xmark} & \textcolor{JungleGreen}{\cmark}& \textcolor{Salmon}{\xmark}\\
              RoboFlamingo~(\citeauthor{li2024visionlanguagefoundationmodelseffective}) & 2024 & CALVIN & Pybullet & OpenFlamingo & OpenFlamingo & \textcolor{Salmon}{\xmark} & \textcolor{JungleGreen}{\cmark} & \textcolor{Salmon}{\xmark} & \textcolor{JungleGreen}{\cmark}& \textcolor{Salmon}{\xmark}\\
              OpenVLA~(\citeauthor{kim2024openvlaopensourcevisionlanguageactionmodel}) & 2024 & Open X-Embodiment & - & Llama 2 7B & DINOv2 \& SigLIP & \textcolor{JungleGreen}{\cmark} & \textcolor{JungleGreen}{\cmark} & \textcolor{Salmon}{\xmark} & \textcolor{JungleGreen}{\cmark}& \textcolor{Salmon}{\xmark}\\
              RT-X~(\citeauthor{open_x_embodiment_rt_x_2023}) & 2024 & Open X-Embodiment & - & PaLI-X,PaLM-E & PaLI-X,PaLM-E & \textcolor{JungleGreen}{\cmark} & \textcolor{JungleGreen}{\cmark} & \textcolor{Salmon}{\xmark} & \textcolor{JungleGreen}{\cmark}& \textcolor{Salmon}{\xmark}\\
              PIVOT~(\citeauthor{nasiriany2024pivotiterativevisualprompting}) & 2024 & Open X-Embodiment & - & GPT-4, Gemini & GPT-4, Gemini & \textcolor{JungleGreen}{\cmark} & \textcolor{JungleGreen}{\cmark} & \textcolor{Salmon}{\xmark} & \textcolor{Salmon}{\xmark}& \textcolor{JungleGreen}{\cmark}\\
              RT-Hierarchy (\citeauthor{rth2024rss})  & 2024 & \# & - & PaLI-X & PaLI-X & \textcolor{JungleGreen}{\cmark} & \textcolor{JungleGreen}{\cmark} & \textcolor{Salmon}{\xmark} & \textcolor{JungleGreen}{\cmark} & \textcolor{Salmon}{\xmark} \\ 
              3D-VLA~(\citeauthor{zhen20243dvla3dvisionlanguageactiongenerative}) & 2024 & RL-Bench \& CALVIN & CoppeliaSim \& PyBullet & 3D-LLM & 3D-LLM & \textcolor{Salmon}{\xmark} & \textcolor{JungleGreen}{\cmark} & \textcolor{Salmon}{\xmark} & \textcolor{JungleGreen}{\cmark}& \textcolor{Salmon}{\xmark}\\
              Octo~(\citeauthor{octomodelteam2024octo}) & 2024 & Open X-Embodiment & - & T5 & CNN & \textcolor{JungleGreen}{\cmark} & \textcolor{JungleGreen}{\cmark} & \textcolor{Salmon}{\xmark} & \textcolor{JungleGreen}{\cmark}& \textcolor{Salmon}{\xmark}\\
              ECoT~(\citeauthor{zawalski2024roboticcontrolembodiedchainofthought}) & 2024& BridgeData V2 & - & Llama 2 7B & DinoV2 \& SigLIP & \textcolor{JungleGreen}{\cmark} & \textcolor{JungleGreen}{\cmark} & \textcolor{Salmon}{\xmark} & \textcolor{JungleGreen}{\cmark}& \textcolor{Salmon}{\xmark}\\
              LEGION(\citeauthor{meng2025preserving}) & 2024 & Meta-World & MuJoCo & RoBERTa& * & \textcolor{JungleGreen}{\cmark} & \textcolor{Salmon}{\xmark} &  \textcolor{JungleGreen}{\cmark} & \textcolor{Salmon}{\xmark} &\textcolor{Salmon}{\xmark} \\
              RACER~(\citeauthor{dai2024racer}) & 2024 & RLbench & CoppelaSim & Llama3-llava-next-8B & LLaVA & \textcolor{JungleGreen}{\cmark} & \textcolor{JungleGreen}{\cmark} & \textcolor{Salmon}{\xmark} &  \textcolor{JungleGreen}{\cmark} & \textcolor{Salmon}{\xmark} \\

            \bottomrule
        \end{tabular}
        \end{adjustbox}
        \begin{tablenotes}
        \small \# Their own benchmark; * Directly get environment states; FMs: use foundational models or not; \\
        \small RL: reinforcement learning; IL: imitation learning; MP: motion planning.
        \end{tablenotes}
    \end{threeparttable}
    
    \label{tab:approaches}
\end{table*}

\begin{table*}[ht!]
    \centering
    \caption{\textcolor{black}{Comparison of language-conditioned robot manipulation approaches II in Section \ref{sec:comparative_analysis}.}}
    \begin{threeparttable}
    \begin{adjustbox}{width=\linewidth,center}
        \begin{tabular}{ l c c c c c c c c c}
            \toprule
            Models & Years & \makecell{Benchmark or\\ \textit{Simulation Engine}} & \makecell{Language \\ Module} & \makecell{Perception \\ Module} & \makecell{Real world \\experiments} & FMs & \makecell{RL} & \makecell {IL} & \makecell {MP}\\
            
              \midrule
              Ground4Act~(\citeauthor{YANG2024105280}) & 2024 & Gazebo & Transformer & ResNet101, BERT & \textcolor{JungleGreen}{\cmark} & \textcolor{Salmon}{\xmark} & \textcolor{JungleGreen}{\cmark} & \textcolor{Salmon}{\xmark} & \textcolor{Salmon}{\xmark}\\
              LOVM~(\citeauthor{ye2024task}) & 2024 & \# & BiGRU & LOVM & \textcolor{Salmon}{\xmark} & \textcolor{Salmon}{\xmark} & \textcolor{JungleGreen}{\cmark} &  \textcolor{Salmon}{\xmark} & \textcolor{JungleGreen}{\cmark}\\
              ECLAIR~(\citeauthor{10803055}) & 2024 & \# & GPT-3-turbo & * & \textcolor{JungleGreen}{\cmark} & \textcolor{JungleGreen}{\cmark} & \textcolor{JungleGreen}{\cmark} & \textcolor{Salmon}{\xmark} & \textcolor{Salmon}{\xmark}\\
              PR2L~(\citeauthor{chen2024visionlanguagemodelsprovidepromptable}) & 2024 & MineDojo, HM3D & InstructBLIP & InstructBLIP & \textcolor{JungleGreen}{\cmark} & \textcolor{JungleGreen}{\cmark} & \textcolor{JungleGreen}{\cmark} & \textcolor{Salmon}{\xmark}  & \textcolor{Salmon}{\xmark}\\
              AHA~(\citeauthor{duan2024ahavisionlanguagemodeldetectingreasoning}) & 2024 & RLBench, ManiSkill, RoboFail & LLaMA-2-13B & CLIP & \textcolor{Salmon}{\xmark} & \textcolor{JungleGreen}{\cmark} & \textcolor{JungleGreen}{\cmark} & \textcolor{Salmon}{\xmark} & \textcolor{JungleGreen}{\cmark}\\
              KOI~(\citeauthor{lu2024koiacceleratingonlineimitation}) & 2024 & Meta-World, LIBERO & GPT-4v & KOI & \textcolor{JungleGreen}{\cmark} & \textcolor{JungleGreen}{\cmark} & \textcolor{Salmon}{\xmark} & \textcolor{JungleGreen}{\cmark} & \textcolor{JungleGreen}{\cmark}\\
              GPT-4V(ISION)~(\citeauthor{10711245}) & 2024 & \# & GPT-4 & GPT-4 & \textcolor{JungleGreen}{\cmark} & \textcolor{JungleGreen}{\cmark} & \textcolor{Salmon}{\xmark} & \textcolor{JungleGreen}{\cmark} & \textcolor{Salmon}{\xmark}\\
              HiRT~(\citeauthor{zhang2025hirtenhancingroboticcontrol}) & 2024 & Meta-World, Franka-Kitchen & InstructBLIP & CNN & \textcolor{JungleGreen}{\cmark} & \textcolor{JungleGreen}{\cmark} & \textcolor{Salmon}{\xmark} & \textcolor{JungleGreen}{\cmark} & \textcolor{Salmon}{\xmark}\\
              Sentinel~(\citeauthor{agia2024unpackingfailuremodesgenerative}) & 2024 & \# & GPT-4o & PointNet++ & \textcolor{JungleGreen}{\cmark} & \textcolor{JungleGreen}{\cmark} & \textcolor{Salmon}{\xmark} & \textcolor{Salmon}{\xmark} & \textcolor{JungleGreen}{\cmark}\\
              
              RoLD~(\citeauthor{tan2024roldrobotlatentdiffusion}) & 2024 & Open X-E, Robomimic, Meta-World & DistilBERT & DistilBERT & \textcolor{Salmon}{\xmark} & \textcolor{Salmon}{\xmark} & \textcolor{Salmon}{\xmark} & \textcolor{Salmon}{\xmark} & \textcolor{JungleGreen}{\cmark}\\
              
              \midrule
              
              ITS~(\citeauthor{Jin2025}) & 2025 & * & LLaMA & A2C & \textcolor{Salmon}{\xmark} & \textcolor{JungleGreen}{\cmark} & \textcolor{JungleGreen}{\cmark} & \textcolor{Salmon}{\xmark} & \textcolor{JungleGreen}{\cmark}\\
              SIAMS~(\citeauthor{hatanaka2025reinforcement}) & 2025 & Miniworld & LTL & CNN & \textcolor{Salmon}{\xmark} & \textcolor{JungleGreen}{\cmark} & \textcolor{JungleGreen}{\cmark} & \textcolor{Salmon}{\xmark} & \textcolor{Salmon}{\xmark}\\
              CRTO~(\citeauthor{10955186}) & 2025 & Continual World & ChatGPT & * & \textcolor{Salmon}{\xmark} & \textcolor{JungleGreen}{\cmark} & \textcolor{JungleGreen}{\cmark} & \textcolor{Salmon}{\xmark} & \textcolor{JungleGreen}{\cmark}\\
              LAMARL~(\citeauthor{ZhuLAMARL}) & 2025 & \textcolor{Salmon}{\xmark} & OpenAI & MADDPG & \textcolor{JungleGreen}{\cmark} & \textcolor{JungleGreen}{\cmark} & \textcolor{JungleGreen}{\cmark} & \textcolor{Salmon}{\xmark} & \textcolor{JungleGreen}{\cmark}\\
              ARCHIE~(\citeauthor{TurcatoARCHIE}) & 2025 & \# & GPT-4 & * & \textcolor{JungleGreen}{\cmark} & \textcolor{JungleGreen}{\cmark} & \textcolor{JungleGreen}{\cmark} & \textcolor{Salmon}{\xmark} & \textcolor{JungleGreen}{\cmark}\\
              RealBEF~(\citeauthor{wang2025guiding}) & 2025 & Meta-World & ALBEF & CNN & \textcolor{Salmon}{\xmark} & \textcolor{JungleGreen}{\cmark} & \textcolor{JungleGreen}{\cmark} & \textcolor{Salmon}{\xmark} & \textcolor{Salmon}{\xmark}\\
              LLMRewardShaping~(\citeauthor{guo2024utilizing}) & 2025 & Meta-World & GPT-4 & * & \textcolor{JungleGreen}{\cmark} & \textcolor{JungleGreen}{\cmark} & \textcolor{JungleGreen}{\cmark} & \textcolor{Salmon}{\xmark} & \textcolor{Salmon}{\xmark}\\
              BOSS~(\citeauthor{YangBOSS}) & 2025 & LIBERO & OpenVLA & ResNet & \textcolor{Salmon}{\xmark} & \textcolor{JungleGreen}{\cmark} & \textcolor{Salmon}{\xmark} & \textcolor{JungleGreen}{\cmark} & \textcolor{JungleGreen}{\cmark}\\
              LAV-ACT~(\citeauthor{TripathiLAV-ACT}) & 2025 & MuJuCo & Voltron & Voltron & \textcolor{JungleGreen}{\cmark} & \textcolor{Salmon}{\xmark} & \textcolor{Salmon}{\xmark} & \textcolor{JungleGreen}{\cmark} & \textcolor{Salmon}{\xmark}\\
              TPM~(\citeauthor{YangTPM}) & 2025 & MuJuCo & GPT-4 & ResNet & \textcolor{JungleGreen}{\cmark} & \textcolor{JungleGreen}{\cmark} & \textcolor{Salmon}{\xmark} & \textcolor{JungleGreen}{\cmark} & \textcolor{JungleGreen}{\cmark}\\
              Mamba~(\citeauthor{10966860}) & 2025 & \# & Mamba & Mamba & \textcolor{JungleGreen}{\cmark} & \textcolor{JungleGreen}{\cmark} & \textcolor{Salmon}{\xmark} & \textcolor{JungleGreen}{\cmark} & \textcolor{JungleGreen}{\cmark}\\
              TransformerPolicy~(\citeauthor{10934975}) & 2025 & CALVIN & Transformer & Sentence-BERT & \textcolor{JungleGreen}{\cmark} & \textcolor{Salmon}{\xmark} & \textcolor{Salmon}{\xmark} & \textcolor{JungleGreen}{\cmark} & \textcolor{JungleGreen}{\cmark}\\
              HierarchicalLCL~(\citeauthor{10.1007/978-981-96-1614-5_11}) & 2025 & CALVIN & OpenFlamingoM-3B & ViT & \textcolor{Salmon}{\xmark} & \textcolor{JungleGreen}{\cmark} & \textcolor{Salmon}{\xmark} & \textcolor{JungleGreen}{\cmark} & \textcolor{JungleGreen}{\cmark}\\
              BLADE~(\citeauthor{liu2024learning}) & 2025 & CALVIN & GPT-4 & PCT & \textcolor{JungleGreen}{\cmark} & \textcolor{JungleGreen}{\cmark} & \textcolor{Salmon}{\xmark} & \textcolor{JungleGreen}{\cmark} & \textcolor{JungleGreen}{\cmark}\\
              LES6DPose~(\citeauthor{11075556}) & 2025 & Isaac Gym & GPT-4 & PointNet++ & \textcolor{JungleGreen}{\cmark} & \textcolor{JungleGreen}{\cmark} & \textcolor{Salmon}{\xmark} & \textcolor{Salmon}{\xmark} & \textcolor{JungleGreen}{\cmark}\\
              SafetyFilter~(\citeauthor{10933541}) & 2025 & \# & GPT-4o & CLIP & \textcolor{JungleGreen}{\cmark} & \textcolor{JungleGreen}{\cmark} & \textcolor{Salmon}{\xmark} & \textcolor{Salmon}{\xmark} & \textcolor{JungleGreen}{\cmark}\\
              TARAD~(\citeauthor{11124589}) & 2025 & RLBench & GPT-4o & CLIP & \textcolor{JungleGreen}{\cmark} & \textcolor{JungleGreen}{\cmark} & \textcolor{Salmon}{\xmark} & \textcolor{JungleGreen}{\cmark} & \textcolor{JungleGreen}{\cmark}\\
              DISCO~(\citeauthor{hao2025language}) & 2025 & CALVIN & GPT-4o & * & \textcolor{JungleGreen}{\cmark} & \textcolor{JungleGreen}{\cmark} & \textcolor{Salmon}{\xmark} & \textcolor{Salmon}{\xmark} & \textcolor{JungleGreen}{\cmark}\\
              TinyVLA~(\citeauthor{10900471}) & 2025 & Meta-World & Pythia & MLP & \textcolor{JungleGreen}{\cmark} & \textcolor{JungleGreen}{\cmark} & \textcolor{Salmon}{\xmark} & \textcolor{Salmon}{\xmark} & \textcolor{JungleGreen}{\cmark}\\
              ASD-QR~(\citeauthor{wang2025integrating}) & 2025 & ScalingUp & GPT3 & CLIP & \textcolor{Salmon}{\xmark} & \textcolor{JungleGreen}{\cmark} & \textcolor{JungleGreen}{\cmark} & \textcolor{Salmon}{\xmark} & \textcolor{JungleGreen}{\cmark}\\
              RDT-1B~(\citeauthor{liu2025rdt1bdiffusionfoundationmodel}) & 2025 & \# & GPT-4-Turbo & T5-XXL & \textcolor{JungleGreen}{\cmark} & \textcolor{JungleGreen}{\cmark} & \textcolor{Salmon}{\xmark} & \textcolor{JungleGreen}{\cmark} & \textcolor{Salmon}{\xmark}\\
              GRAVMAD~(\citeauthor{chen2025gravmad}) & 2025 & RLBench & GPT-4o & CLIP & \textcolor{JungleGreen}{\cmark} & \textcolor{JungleGreen}{\cmark} & \textcolor{Salmon}{\xmark} & \textcolor{JungleGreen}{\cmark} & \textcolor{JungleGreen}{\cmark}\\
              GR-MG~(\citeauthor{li2025gr}) & 2025 & CALVIN & Transformer & T5-Base & \textcolor{JungleGreen}{\cmark} & \textcolor{Salmon}{\xmark} & \textcolor{Salmon}{\xmark} & \textcolor{Salmon}{\xmark} & \textcolor{JungleGreen}{\cmark}\\  LEMMo-Plan~(\citeauthor{chen2025lemmo}) & 2025 & \textcolor{Salmon}{\xmark} & GPT-4o & * & \textcolor{JungleGreen}{\cmark} & \textcolor{JungleGreen}{\cmark} & \textcolor{Salmon}{\xmark} & \textcolor{Salmon}{\xmark} & \textcolor{JungleGreen}{\cmark}\\   
              LGGD~(\citeauthor{jiang2025language}) & 2025 & \textcolor{Salmon}{\xmark} & CLIP & CLIP, ResNet50 & \textcolor{JungleGreen}{\cmark} & \textcolor{Salmon}{\xmark} & \textcolor{Salmon}{\xmark} & \textcolor{Salmon}{\xmark} & \textcolor{JungleGreen}{\cmark} \\
              AffordanceGrasp-R1~(\citeauthor{zhou2026affordancegrasp}) & 2026 & HANDAL & Qwen2.5-VL-72B-Instruct & SAM2 & \textcolor{JungleGreen}{\cmark} & \textcolor{JungleGreen}{\cmark} & \textcolor{JungleGreen}{\cmark} & \textcolor{Salmon}{\xmark} & \textcolor{JungleGreen}{\cmark}\\ 
            \bottomrule
        \end{tabular}
        \end{adjustbox}
        \begin{tablenotes}
        \small \# Their own benchmark; * Directly get environment states; FMs: use foundational models or not; \\
        \small RL: Reinforcement Learning; IL: Imitation Learning; MP: Motion Planning.
        \end{tablenotes}
    \end{threeparttable}
    
    \label{tab:approaches-v2}
\end{table*}

Table \ref{tab:vla_table} provides a comprehensive comparison and classification of various Vision-Language-Action (VLA) models discussed in Section \ref{sec:vla}. The table organizes these models based on their primary \textit{Optimization Direction}, which includes improvements in Group I for Perception (Data Source and Augmentation, Spatial Understanding, Multimodal Fusion), Group II for Reasoning (Long-horizon Task Solving, Knowledge Preservation, Reasoning with World Models), Group III for Action Generation, and Group IV for Learning \& Adaptation. For each model, the table details several key design dimensions:
\begin{itemize}[nosep, noitemsep]
    \item \textbf{Observation}: The input modalities used by the model, such as RGB images (RGB), depth information (D), including point clouds or other types, robot proprioception (ROB), and text instructions (TX).
    \item \textbf{Action generation}: The methodology for producing actions, categorized into Flow Matching (FM), Diffusion Models (DM), Autoregressive (AR) decoding, or Direct Prediction (DP).
    \item \textbf{VLM policy}: The internal mechanisms of the policy, highlighting the use of advanced reasoning like Chain-of-Thought (CoT), Future Prediction (FP), or explicit Memory (MEM).
    \item \textbf{Pretraining}: The type of the training data, specifying whether it involves Multiple Datasets (MD) or Cross-embodiment (CE) data.
    \item \textbf{Scenarios}: The evaluation context, indicating if the model was tested across Multi-scenario (MS) setups, deployed in the Real World (RW), or demonstrated Cross-embodiment Execution (CE).
\end{itemize}
This classification highlights the key features, capabilities, and architectural choices of recent VLA models, offering a structured overview of the current research landscape.

\subsection*{B. Tabular comparison of language-conditioned robotic manipulations}
\label{sec:apxb}
Table \ref{tab:approaches} and Table \ref{tab:approaches-v2} present a tabular comparison of various language-conditioned robot manipulation approaches discussed in Sections \ref{sec:language-for-state-evaluation} -  \ref{sec:comparative_analysis}. The tables summarize key aspects such as the year of publication, benchmarks or simulation engines used, language and perception modules, real-world experiment status, and the utilization of foundational models (FMs), reinforcement learning (RL), imitation learning (IL), and motion planning (MP).

\bibliographystyle{SageH}
\bibliography{reference}

\end{document}